\documentclass[1p]{elsarticle}

\usepackage{lineno,hyperref} 
\modulolinenumbers[1] 

\journal{Neural Networks}%

\usepackage{bm}
\usepackage{amsmath}

\usepackage{amsfonts}

\usepackage[subrefformat=parens]{subcaption}

\usepackage{multirow}
\usepackage{graphicx}
\usepackage{color}

\usepackage{here}

\begin{document}

\begin{frontmatter}

\title{Chaos-based reinforcement learning with TD3}



\author[NDA]{Toshitaka Matsuki\corref{CORRESPONDING}}
\cortext[CORRESPONDING]{Corresponding author}
\ead{t_matsuki@nda.ac.jp}
\author[CHIBA,TOKYO]{Yusuke Sakemi}
\author[CHIBA,TOKYO]{Kazuyuki Aihara}

\address[NDA]{National Defense Academy of Japan, Kanagawa, Japan}
\address[CHIBA]{Research Center for Mathematical Engineering, Chiba Institute of Technology, Narashino, Japan}
\address[TOKYO]{International Research Center for Neuro intelligence (WPI-IRCN), The University of Tokyo, Tokyo, Japan}

\begin{abstract}
Chaos-based reinforcement learning (CBRL) is a method in which the agent's internal chaotic dynamics drives exploration.
However, the learning algorithms in CBRL have not been thoroughly developed in previous studies, nor have they incorporated recent advances in reinforcement learning.
This study introduced Twin Delayed Deep Deterministic Policy Gradients (TD3), which is one of the state-of-the-art deep reinforcement learning algorithms that can treat deterministic and continuous action spaces, to CBRL.
The validation results provide several insights.
First, TD3 works as a learning algorithm for CBRL in a simple goal-reaching task.
Second, CBRL agents with TD3 can autonomously suppress their exploratory behavior as learning progresses and resume exploration when the environment changes.
Finally, examining the effect of the agent's chaoticity on learning shows that there exists a suitable range of chaos strength in the agent's model to flexibly switch between exploration and exploitation and adapt to environmental changes.
\end{abstract}


\end{frontmatter}


\section{Introduction}
Neural networks have been studied for many years, partially inspired by findings in neuroscience research \cite{hassabis2017neuroscience}.
In recent years, the development of deep learning techniques used to successfully train deep neural networks has produced remarkable results in various fields, e.g. image recognition \cite{krizhevsky2012imagenet, 2016resnet}, speech recognition \cite{2014ETE_Speech, 2016deepspeech}, natural language processing \cite{cho2014learning, sutskever2014sequence}, and practical applications including next generation wireless communications, economic forecasting, and other areas \cite{vaigandla2025comprehensive, adamopoulos2025enhancing}.
The transformer \cite{2017transformer}, a groundbreaking model for natural language processing, has provided a breakthrough in the capabilities of artificial intelligence (AI) as the underlying technology for Large Language Models (LLMs) \cite{2018bert}. It has also demonstrated high performance in other areas such as image processing and speech recognition \cite{2020ViT, 2023whisper}.
AI capabilities based on deep learning techniques have developed quickly over the past decade, leading to innovations in diverse fields.

In recent years, AI has demonstrated even high-quality creative abilities \cite{Sora2024}, and the creativity of AI has been attracting much attention \cite{franceschelli2023creativity, guzik2023originality}.
Historically, J. McCarthy et al. discussed the difference between creative thinking and unimaginative competent thinking in a proposal for the Dartmouth Workshop in 1956 \cite{dartmouth1955}.
They raised the importance of randomness in AI and the need to pursue how randomness can be effectively injected into machines, just as the brain does.
Reinforcement learning is a machine learning approach using random exploration inspired by behavioral insights, where animals learn from their actions and their consequences.
In reinforcement learning, 
a learning agent performs exploratory action driven by random numbers to an environment and improves its policy based on feedback from the environment.
Various studies on reinforcement learning have been conducted over long years \cite{Sutton}.
In recent years, research on deep reinforcement learning, which incorporates deep learning techniques, has become popular, and this approach has made it possible to learn even more difficult tasks \cite{Mnih2013, hessel2018rainbow}
Deep reinforcement learning can also provide effective learning performance in more complex tasks such as Go
\cite{silver2017mastering, silver2018general}.
Randomness enables AI systems to 
perform exploratory and interactively learning without explicit teacher.

Organisms can act spontaneously and autonomously in environments with diverse dynamics, and can adapt to the environment based on the experience gained through these behaviors \cite{Sutton}.
Reinforcement learning agents stochastically explore the environment by introducing random numbers to select actions that deviate from the optimal action determined by the current policy.
On the other hand, it remains a fascinating unsolved problem to understand how the biological brain behaves in various ways and realizes exploratory learning.
One hypothesis for the essential property of the source of exploration is to utilize fluctuations within the brain.
Various studies have shown that fluctuations caused by spontaneous background activity in the neural populations vary their responses \cite{Fontanini2008Behavioral}.
Experiments measuring neural activity in the leeches \cite{briggman2005optical} and human motor cortex \cite{fox2007intrinsic} suggest that fluctuations in neural activity influence decision-making and behavioral variability.

Freeman pointed out the possibility that the brain uses chaotic dynamics for exploratory learning \cite{Freeman2}.
Skarda and Freeman 
showed that there are many chaotic attractors in the dynamics of the olfactory bulb that are attracted when the rabbit is exposed to the known olfactory conditioned stimulus \cite{Freeman1}. 
This study also suggests that chaotic dynamics is used for the reorganization of new attractors corresponding to new stimuli.
Freeman argues that the chaotic dynamics of the brain continually generate new patterns of activity necessary to generate new structures and that this underlies the ability of trial and error problem solving \cite{Freeman2}.
Aihara et al. discovered the chaotic dynamics in the squid giant axon and constructed chaotic neural network models based \cite{aihara1986, aihara1990chaotic, adachi1997associative}, and proposed Chaotic Simulated Annealing effective for combinatorial optimization problems \cite{chen1995chaotic}.
Hoerzer et al. showed that a reservoir network, which fixes the recurrent and input weights, can acquire various functions using an exploratory learning algorithm based on random noise \cite{hoerzer2012}. 
Additionally, this study suggests that stochasticity (trial-to-trial variability) of neuronal response plays a crucial role in the learning process.
It has also been shown that the system's own chaotic dynamics can drive exploration in this learning process \cite{matsuki_iconip2016, matsuki2020adaptive}.
These studies have implications for understanding how the brain achieves exploratory learning and utilizes chaotic dynamics in the process.

Shibata et al. have proposed chaos-based reinforcement learning (CBRL), which exploits the internal chaotic dynamics of the system for exploration  \cite{shibata2015}. This method uses a recurrent neural network as an agent system and its internal chaotic dynamics as a source of exploration components.
The algorithms used in CBRL require treating deterministic and continuous action to eliminate stochastic exploration and human-designed action selection. 
Due to the necessity of a reinforcement learning algorithm that can handle deterministic and continuous actions without requiring random noise, Shibata et al. proposed causality trace learning and used it to train CBRL agents \cite{2017goto, 2019sato}.
However, this method has limited performance, and the CBRL algorithm is not yet well established.
To address this limitation, this study proposes a novel CBRL framework that introduces the Twin Delayed Deep Deterministic Policy Gradient (TD3) algorithm, which is one of the state-of-the-art deep reinforcement learning algorithms designed for handling deterministic and continuous actions.
The ability of the agent is evaluated by learning a simple goal task.
We also examine how CBRL agents trained using TD3 respond to environmental changes.
Through these experiments, we demonstrate that internal-dynamics-based exploration enables more flexible switching between exploration and exploitation, in contrast to standard deterministic approaches that rely on external random noise.
Furthermore, we investigate how the behavior of the agents changes depending on the levels of chaoticity in the model's dynamics and how the dynamics affects their learning performance and ability to adapt to environmental changes.
Finally, we evaluate the learning performance of the proposed method with several environments, including a nonlinear control task and a partially observable task.

This paper is organized as follows.
Section 2 summarizes chaos-based reinforcement learning.
Section 3 describes the experimental method.
Section 4 presents the results of the experiments.
Section 5 discusses the experimental results.
Section 6 summarizes the conclusions of this study and future research directions.

\begin{figure}[t]
  \begin{minipage}[t]{0.49\linewidth}
    \centering
    \includegraphics[bb=0 0 1800 1200, scale=0.1]{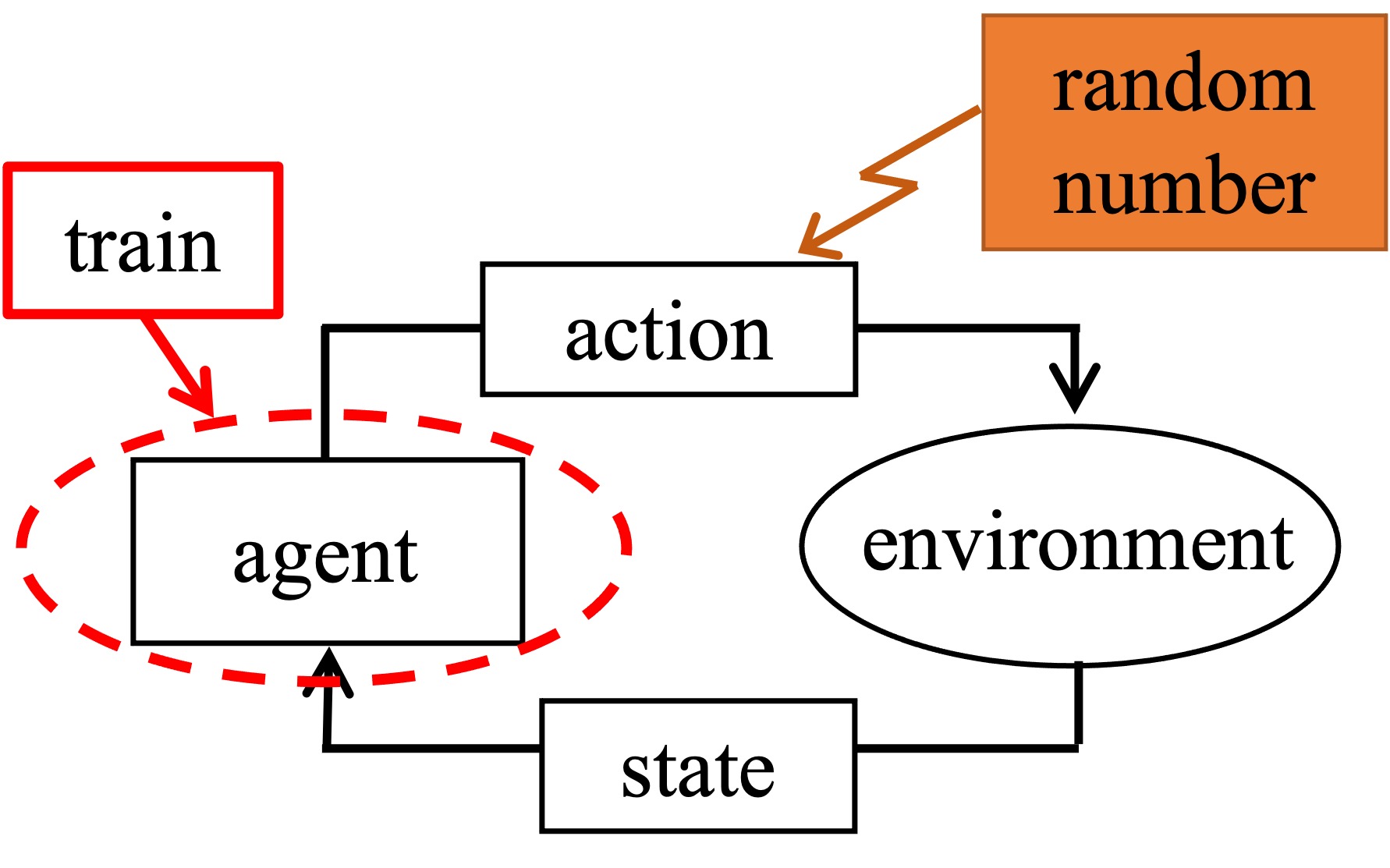}
    \subcaption{Regular reinforcement learning.}
  \end{minipage}
  \begin{minipage}[t]{0.49\linewidth}
    \centering
    \includegraphics[bb=0 0 1800 1200, scale=0.1]{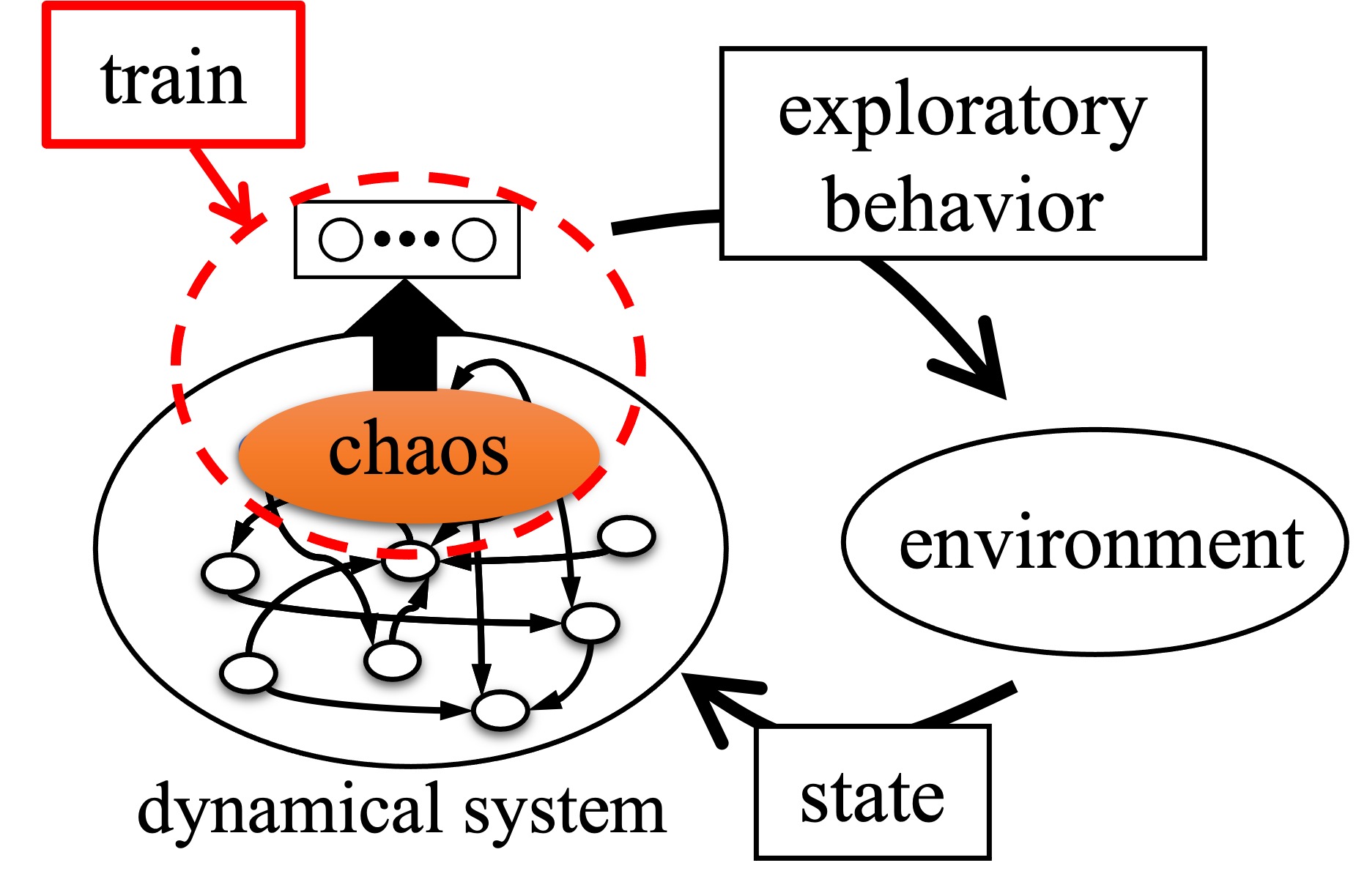}
    \subcaption{Chaos-based reinforcement learning.}
  \end{minipage}
  \caption{Chaos-Based Reinforcement Learning (CBRL).
(a) Overview of Regular Reinforcement Learning:
The agent decides an action, and then stochastic choices or noise based on random numbers affect the action and drive exploration.
The action changes the agent's state through the environment.
The agent improves its policy based on experience gained from the interactions.
Since exploration noise is externally provided with random numbers, Regular RL agents cannot improve their exploration through learning.
(b) Overview of CBRL:
A dynamical system that exhibits chaotic behavior is used as the agent's system.
In CBRL, the agent behaves exploratively due to fluctuations caused by internal dynamics rather than external random numbers.
Since internal dynamics drives exploration, CBRL agents have the potential to be able to optimize exploration by purposefully adapting their internal dynamics through learning.
}
\label{fig:rl_and_cbrl}
\end{figure}

\section{Chaos-based reinforcement learning}
\subsection{Exploration driven by internal chaotic dynamics}
In the Chaos-based reinforcement learning method, the agent explores using internal chaotic dynamics.
Figure \ref{fig:rl_and_cbrl} shows an overview of regular reinforcement learning and chaos-based reinforcement learning (CBRL). In general, exploration in reinforcement learning is performed probabilistically based on external random numbers. The $\epsilon$-greedy method selects a random action with a probability of $\epsilon$. The softmax policy selects an action stochastically using random numbers from a Boltzmann distribution of the state action values. Some algorithms for continuous action spaces explore by adding random noise to the action outputs for exploration purpose \cite{ddpg}.
On the other hand, in CBRL, exploration is driven by chaotic dynamics within the agent system rather than relying on external random numbers.

A structure in which internal dynamics serves as the basis for exploration resources is plausible as a model of the biological brain and offers several learning benefits.
As shown in Fig. \ref{fig:rl_and_cbrl}(b), the source of exploration (i.e., chaotic dynamics) is a property of the agent system itself in CBRL. The properties of the system dynamics can be changed by learning. Therefore, the agent has the potential to optimize its exploration strategy through training.
Previous studies have shown that CBRL agents can autonomously switch from an exploration mode to an exploitation mode as learning progresses \cite{shibata2015, matsuki2020q}.  In the initial training episodes, when learning has not progressed sufficiently, the agent behaves chaotically and exploratively in the environment. As learning progresses, the system's dynamics becomes more ordered, and the agent's exploratory behavior subsides autonomously.
Additionally, the agent can autonomously resume exploration when the environment changes its rules and the previously learned behavior is no longer rewarding.

\subsection{Expectation for CBRL}
CBRL has been studied with the expectation of clarifying the role of chaotic dynamics in biological brains and realizing intelligent systems based on transient dynamics.
Shibata et al. hypothesized that the key to creating systems with higher exploratory behavior and human-like intelligence is having a dynamical system with rich and spontaneous activity \cite{shibata2015}.
The dynamics of brain activity can be considered as a chaotic itinerancy or as a process that transits through a series of saddle points \cite{kaneko2003chaotic, rabinovich2008transient, kanamaru2023maximal}.
The hypothesis expects the system to reconstruct its exploratory chaotic dynamics into transient dynamics that is purposeful to maximize returns.

Since internal dynamics drives exploration, CBRL agents are also expected to be able to optimize exploration itself by reconstructing their internal dynamics into purposeful transitions through learning. Goto et al. demonstrated that CBRL agents can change motor noise-like exploration into more sophisticated exploration that selects routes to avoid obstacles using an obstacle avoidance task \cite{2016goto, 2017goto}.
It is expected that CBRL agents can acquire more advanced exploration capabilities that effectively utilize previously acquired behaviors by constructing the agent's internal state and decision-making processes as transitive dynamics such as chaotic itinerancy.

\subsection{Issue of learning algorithm for CBRL}
To ensure freedom and autonomy in learning, CBRL agents have used algorithms that deal with deterministic and continuous action spaces rather than stochastic and discrete ones in which action selections are defined heteronomously.
In previous studies, the learning algorithm for CBRL agents has been an Actor-Critic, which can handle deterministic and continuous action spaces \cite{shibata2015}.
However, a classic Actor-Critic method that trains an actor network using a correlation between the external exploration noise and the resulting change in value cannot be employed for CBRL, which does not use random number exploration.
Therefore, causality trace learning, which is similar to Hebbian learning, has been proposed and employed as the training method for the actor network \cite{shibata2015, 2016goto, 2017goto, 2019sato}.
This method generates the teacher signal for the actor network by multiplying the TD error by the input values stored in a tracing mechanism that changes the input tracing rate according to changes in the neuron's output.
However, this method has many problems, such as the inability to introduce the backpropagation method and the difficulty of learning complex tasks. Therefore, the learning algorithm for CBRL has not yet been sufficiently well-established.

\section{Method}

\subsection{TD3}
This study introduces Twin Delayed Deep Deterministic Policy Gradients (TD3) \cite{td3}, a deep reinforcement learning algorithm that can handle deterministic policy and continuous action spaces, to CBRL.
TD3 is an improved algorithm based on the model-free deep reinforcement learning method called ``Deep Deterministic Policy Gradients" (DDPG) \cite{ddpg}, and is one of the state-of-the-art reinforcement learning algorithms\cite{td3}.

In the following part of this subsection, we first describe the DDPG and then the improvements introduced in TD3.
In the DDPG, the agent model consists of an actor network $\mu(s|\theta^{\mu})$ and a critic network $Q(s, a|\theta^Q)$.
Here, $\theta^{\mu}$ and $\theta^Q$ are the weight value parameters of each network.
$\mu(s|\theta^{\mu})$ determines the action output $a$ based on the agent's state $s$ and $Q(s,a|\theta^Q)$ estimates the state action value from $s$ and $a$.
Target networks $\mu'(s|\theta^{\mu'})$ and $Q'(s, a|\theta^{Q'})$ are generated for each network with their weights copied and are used to generate the teacher signals to stabilize learning.

The agent acts in state $s_t$ at each time $t$ according to the following action output, to which the external exploration noise $\epsilon^{\mathrm{a}}_t$ is added as follows:
\begin{eqnarray}
  a_t  = \mu(s_t|\theta_t^{\mu}) + \epsilon^{\mathrm{a}}_t.
\label{eq:exploration_action}
\end{eqnarray}
As a result of the action, the agent receives the reward $r_t$, and the state transitions to $s'_{t}(=s_{t+1})$.
The experience $e_t = (s_t, a_t, r_t, s'_{t})$ is stored in the replay buffer $\mathcal{B}$.

Training is performed using $N$ minibatch data of $e_i=(s_i, a_i, r_i, s'_{i})$ randomly sampled from $\mathcal{B}$ every step.
Note that $i$ indicates the index number of the samples in the $N$ minibatch data.
The teacher signal for the critic network $Q$ for the input data $s_i$ and $a_i$ is estimated as follows:
\begin{eqnarray}
 T^\mathrm{c}_i = r_i + \gamma Q'(s'_{i}, \mu'(s'_{i}|\theta^{\mu'}) | \theta^{Q'}),
\label{eq:Q_teacher}
\end{eqnarray}
where the discount factor $0 \leq \gamma \leq 1$ is a hyperparameter that determines the present value of future rewards.
We then update $ \theta^{Q}$ to minimize the loss function such that
\begin{eqnarray}
 L_Q = \frac{1}{N} \sum_i^N (T^{\mathrm{c}}_i - Q(s_i, a_i | \theta^{Q}))^2.
\label{eq:Q_train}
\end{eqnarray}
The actor network $\mu$ learns with deterministic policy gradient \cite{silver2014deterministic} estimated based on the sampled data:
\begin{eqnarray}
\nabla_{\theta^{\mu}} J(\theta^{\mu}) \approx  
   \frac{1}{N}\sum^N_i \nabla_{\mu(s_i)} Q(s_i,\mu(s_i|\theta^{\mu})|\theta^{Q}) \nabla_{\theta^{\mu}} \mu(s_i|\theta^{\mu}),
\label{eq:mu_train}
\end{eqnarray}
where $J = \mathbb{E}[Q(s_i,\mu(s_i))]$.
Note that $\theta^{Q}$ is fixed and only $\theta^{\mu}$ is updated for training based on Equation (\ref{eq:mu_train}).
The target network weights $\theta^{Q'}$ and $\theta^{\mu'}$ are updated as follows:
\begin{eqnarray}
 \theta^{Q'} & \leftarrow & \tau \theta^{Q} + (1 - \tau) \theta^{Q'},\\
 \theta^{\mu'} & \leftarrow & \tau \theta^{\mu} + (1 - \tau) \theta^{\mu'},
\label{eq:train_target_net}
\end{eqnarray}
where $0<\tau \leq 1$ is the constant parameter that determines the update speed of the target network.

TD3 introduces three methods to the DDPG algorithm: Clipped Double Q-learning, Delayed Policy Updates, and Target Policy Smoothing.
Clipped Double Q-learning is a method to suppress overestimation of the value by preparing two critic networks $Q_1$ and $Q_2$ and adopting the smaller output value when generating teacher signals.
Target Policy Smoothing is a technique to suppress the overfitting of $Q$ to inaccurate value estimations by adding noise limited between $-C$ and $C$ to the output of $\mu'$ during the generation of the teacher signal.
With these techniques, the teacher signal for $Q$ is given by
\begin{eqnarray}
 T^\mathrm{c}_i  = r_i + \gamma \min_{j=1,2} Q_j'(s'_{i}, \mu'(s'_{i}|\theta^{\mu'}) + \epsilon^{\mathrm{t}}_i | \theta^{Q_j'}) \nonumber, \\
 \epsilon^{\mathrm{t}}_i \sim \mathrm{clip}(\mathcal{N}(0, \sigma), -C, C).
\label{eq:TD3_Q_teacher}
\end{eqnarray}
Note that the learning of $\mu$ by equation (\ref{eq:mu_train}) always uses $Q_1$.
Therefore, the training is based on the following gradient 
\begin{eqnarray}
 \nabla_{\theta^{\mu}} J (\theta^{\mu}) \approx \frac{1}{N}\sum_i^N \nabla_{\mu(s_i)}  Q_1(s_i,\mu(s_i|\theta^{\mu})|\theta^{Q_1})\nabla_{\theta^{\mu}} \mu(s_i|\theta^{\mu}).
\label{eq:loss_mu}
\end{eqnarray}
Delayed Policy Updates stabilizes learning by limiting updates of $\mu$, $\mu'$ and $Q'$ to once every $d$ steps.

\subsection{Reservoir network}
CBRL requires the agent system to have chaotic dynamics.
Recurrent neural networks are dynamical systems and appropriate models for CBRL agents. However, training recurrent neural networks with CBRL presents a challenge in balancing the formation of convergent dynamics beneficial for completing the task while maintaining the chaotic dynamics required for exploration.
To avoid this problem, we use an Echo State Network (ESN) that can be trained without modifying the parameters that determine the dynamical properties of the recurrent layer.
The ESN is a type of model known as a Reservoir Network \cite{maass2002real, jaeger2001echo}. It consists of a recurrent layer called the ``reservoir" and an output layer called the ``readout."
The connection weights in the reservoir are randomly initialized and fixed, and only the weights from the reservoir to the readout are trained.
The reservoir receives the time-series input and generates an output that nonlinearly reflects the spatio-temporal context of the input series. 
The readout generates the network output by performing a linear combination of the reservoir state and input.
The dynamical properties of the ESN can be modified with a single fixed parameter. Therefore, it is easy to tune the chaoticity of the system during learning with chaos-based exploration \cite{matsuki2020adaptive}.
Note that, in this study, the ESN is not used to process time series data but rather to generate chaotic dynamics in the agent system.

\begin{figure}[t]
    \centering
    \includegraphics[bb=0 0 944 1324, scale=0.13]{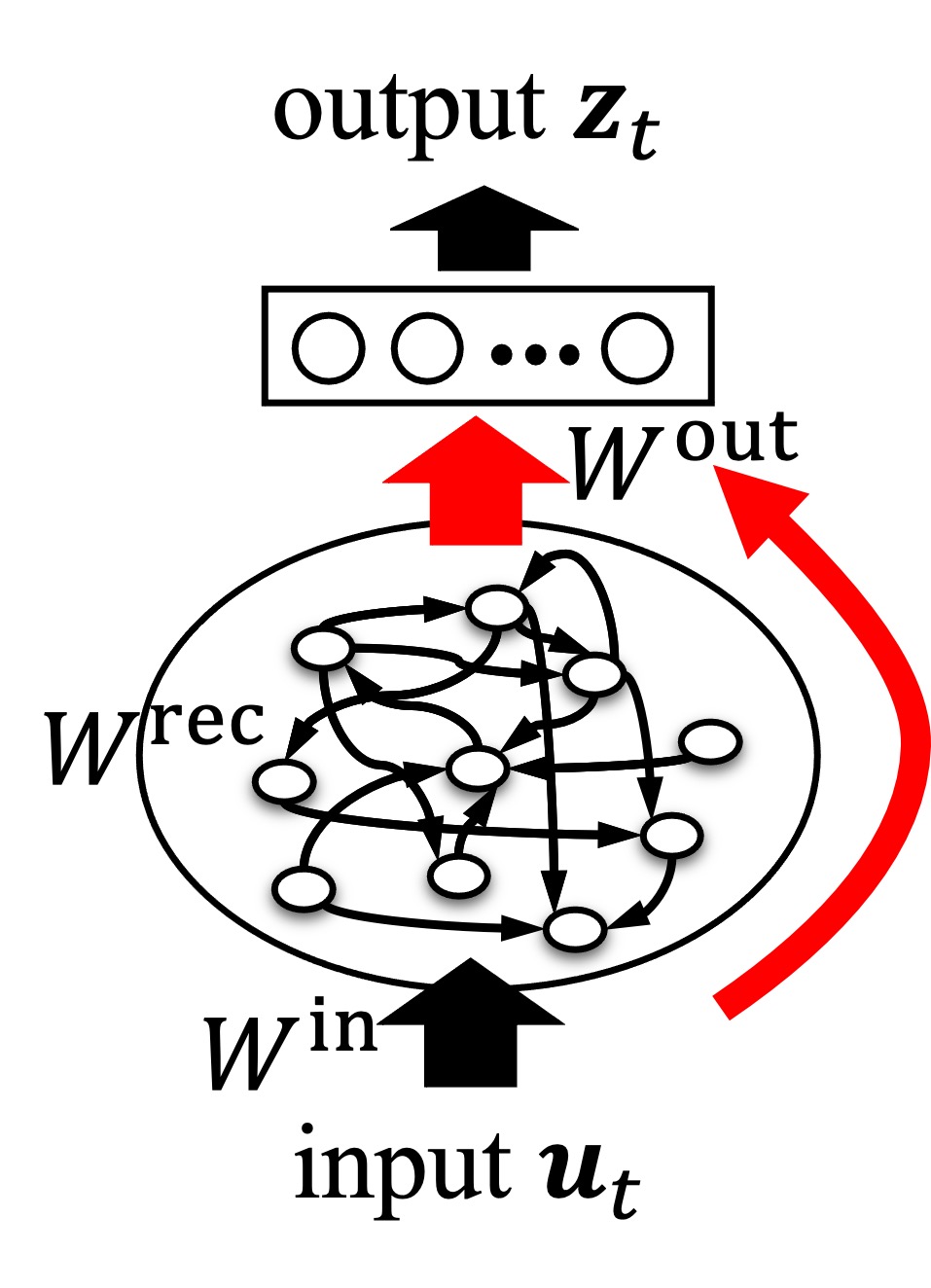}
  \caption{Echo state network (ESN). An ESN has a reservoir that is a special recurrent layer whose recurrent and input weights are randomly and sparsely connected and fixed. Only the weights of the output layer (indicated by the red arrows) are trained.
}
\label{fig:RN}
\end{figure}

The structure of the ESN is shown in Fig. \ref{fig:RN}. 
The reservoir has $N_{\rm{x}}$ neurons that are recurrently 
connected with $\bm{W}^{\rm{rec}}  \in \mathbb{R}^{N_{\rm{x}} \times N_{\rm{x}}}$ with probability of $p$. 
The reservoir receives $N_{\rm{i}}$-dimensional inputs through the weight matrix $\bm{W}^{\rm{in}} \in \mathbb{R}^{N_x \times N_i}$.
Then, the reservoir state is computed by the following equation:
\begin{equation}
\label{eq:reservoir_output}
    \bm{x}_t = f(g \bm{W}^{\rm{rec}} \bm{x}_{t-1} + \bm{W}^{\rm{in}}\bm{u}_t),
\end{equation}
where $g$ is a scaling parameter that scales the strength of the recurrent connections and $\bm{u}_t \in \mathbb{R}^{N_{\rm{i}}}$ is the input vector.
$f(\cdot)=\tanh(\cdot)$ is the activation function applied element-wise.
Typically, $\bm{W}^{\rm{rec}}$ is a sparse and fixed weight matrix initialized with a spectral radius of 1 by the following procedure. A random $N_{\rm{x}} \times N_{\rm{x}}$ matrix $\bm{W}$ is generated (from a uniform distribution in this study), and then the elements of $\bm{W}$ are set to $0$ with probability of $(1-p)$. The matrix is normalized by its spectral radius $\rho$. Thus, the $\bm{W}^{\rm{rec}}$ is initialized as follows:
\begin{equation}
\label{eq:set_spectrum_radius_1}
    \bm{W}^{\rm{rec}} = \frac{1}{\rho}\bm{W}.
\end{equation}
The arbitrary constant $g$ can rescale the spectral radius of $\bm{W}^{\rm{rec}}$. In general, $g$ is usually set to $g<1$ to fulfill the Echo State Property that requires the reservoir state to depend on past input series, but the influence of these inputs fades over finite time. A small $g$ tends to cause rapid decay of input context, while a large $g$ tends to cause a slow decay of it. However, $g>1$ sometimes achieves better performance, and it is argued that the best value for learning is realized when the reservoir dynamics is around the edge of chaos \cite{bertschinger2004real, Asada}.
When $g$ is further increased beyond $1$, and the reservoir dynamics crosses the edge of chaos, the reservoir state exhibits chaotic and spontaneous behavior.
In this study, $g$ is set larger than $1$ to induce chaotic dynamics, which allows the CBRL agent to explore.

The network output $\bm{z}_t \in \mathbb{R}^{N_{\rm{o}}}$ is calculated as
\begin{equation}
\label{eq:readout_output}
    \bm{z}_t =  \tanh \left( \bm{W}^{\rm{out}} [\bm{x}_t; \bm{u}_t] \right),
\end{equation}
where $\bm{W}^{\rm{out}} \in  \mathbb{R}^{N_{\rm{o}} \times (N_{\rm{x}} +  N_{\rm{i}})}$ 
is the output weight matrix, $N_{\rm{o}}$ is the output dimension, and $[\cdot ; \cdot]$ denotes the concatenation of two column vectors.
$\bm{W}^{\rm{out}}$ is often fitted using algorithms such as ridge regression in reservoir computing.
This study uses a reservoir network as an actor network for CBRL agents.
Since $\bm{W}^{\rm{out}}$ is trained by a slightly modified version of the TD3 algorithm as described in the following subsection, $\bm{W}^{\rm{out}}$ is updated by using gradient descent with the Adam optimizer \cite{td3}.

\begin{figure}[t]
  \begin{minipage}[t]{0.32\linewidth}
    \centering
    \includegraphics[bb=0 0 1000 2000, scale=0.1]{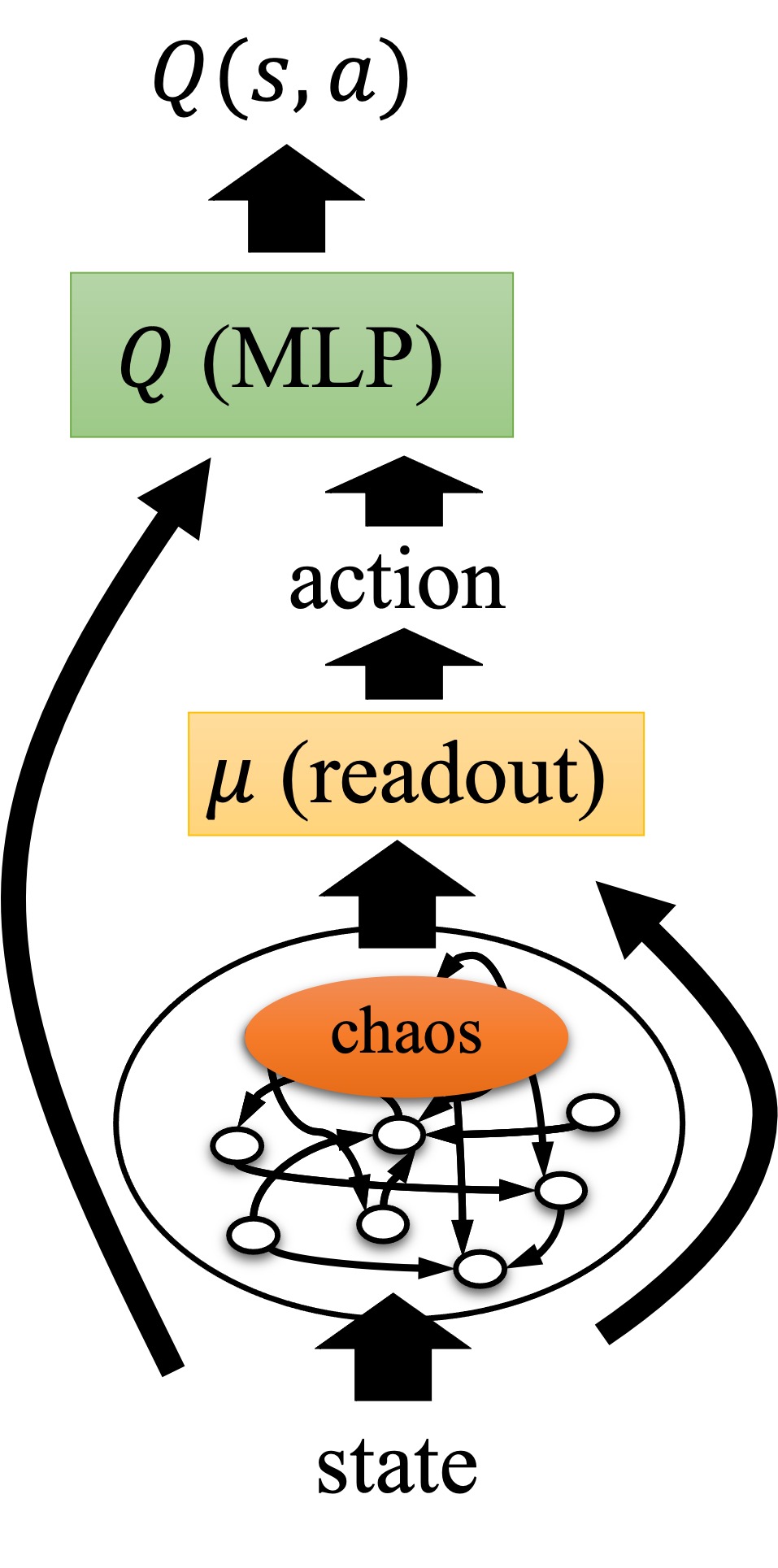}
    \subcaption{TD3-CBRL}
  \end{minipage}
  \begin{minipage}[t]{0.32\linewidth}
    \centering
    \includegraphics[bb=0 0 700 1400, scale=0.1]{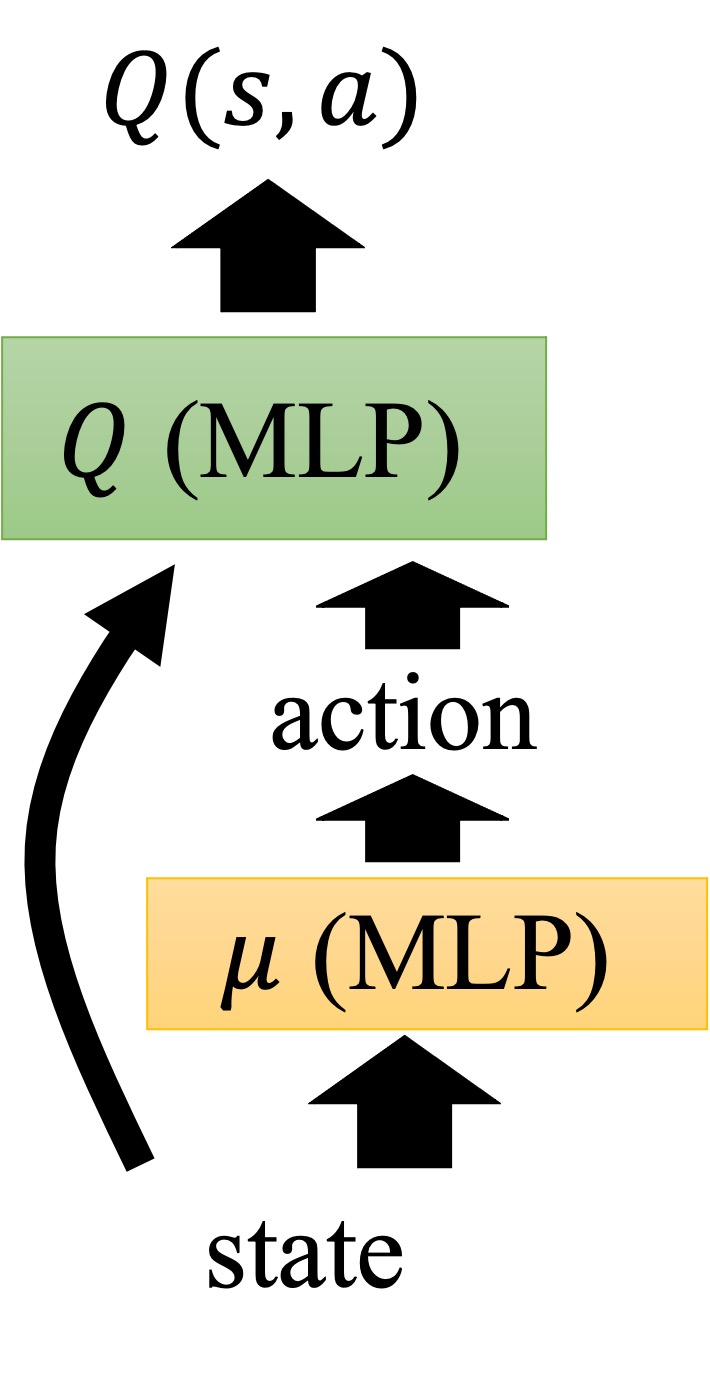}
    \subcaption{TD3 without external exploration noise}
  \end{minipage}
  \begin{minipage}[t]{0.32\linewidth}
    \centering
    \includegraphics[bb=0 0 1200 1400, scale=0.1]{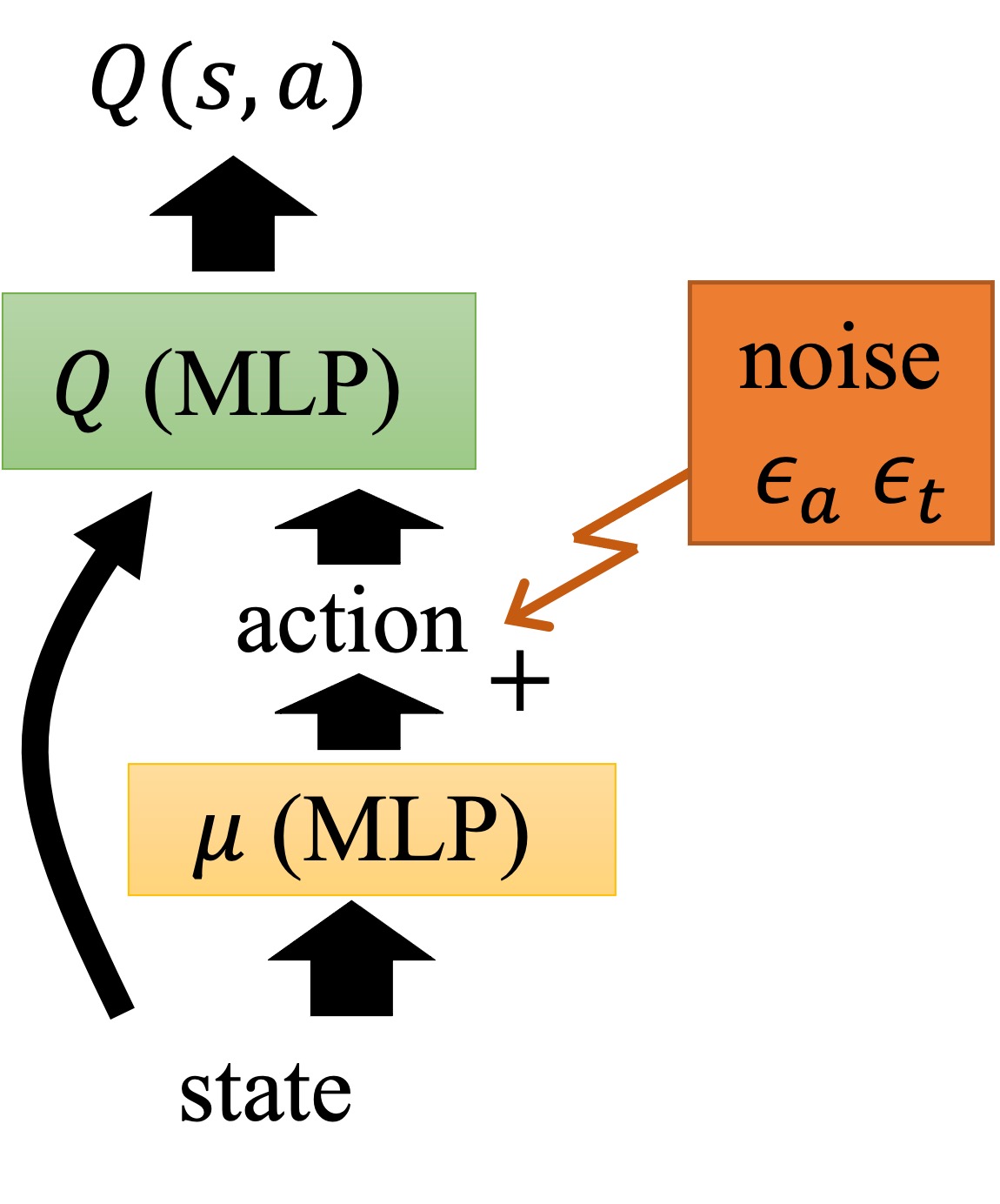}
    \subcaption{Regular TD3}
  \end{minipage}
  \caption{Network Configurations.
This study compares three network configurations of agents.
(a) A TD3-CBRL agent, which uses a chaotic ESN in the actor network $\mu$.
(b) An agent whose actor network is an MLP with a hidden layer consisting of the same number of neurons as the reservoir in (a). In this case, the agent learns without the external random noises $\epsilon^{\mathrm{t}}_a, \epsilon^{\mathrm{t}}_i$ as added in the regular TD3.
(c) An agent that has the same structure as (b) but explores using external random noise as in the regular TD3.
}
\label{fig:td3_cbrls}
\end{figure}

\subsection{TD3-CBRL}
The following modifications are made to the TD3 algorithm to adapt to the CBRL approach.
We eliminated the random number exploration noise $\epsilon^a$ and $\epsilon^t$. Instead of exploration by random numbers, we rely on exploration driven spontaneously by chaotic dynamics and use an ESN with a larger spectral radius, as shown in Fig. \ref{fig:td3_cbrls}(a), for the $\mu$ network.
Here, $g=2.2$ unless otherwise mentioned.
We also add the reservoir state to the experience stored in the replay buffer.
That is, $e_t = (\bm{u}_t, \bm{x}_t, a_t, r_t, \bm{u}_{t+1}, \bm{x}_{t+1})$ is stored into the replay buffer $\mathcal{B}$.
The agent learns without the Back Propagation Through Time method using the stored $\bm{u}$ and $\bm{x}$ as state $s_t$.
This method has been employed in several studies and efficiently trains deep reinforcement learning agents using ESN \cite{2020deqn, matsuki2022deep}.
The variant of CBRL that introduces the above modified TD3 algorithm is called TD3-CBRL in this study.
In the experiment, we compare the three cases shown in Fig. \ref{fig:td3_cbrls} (a-c) to confirm that the ESN dynamics contributes to the exploration during learning.

\begin{figure}[t]
    \centering
    \includegraphics[bb=0 0 1700 1100, scale=0.1]{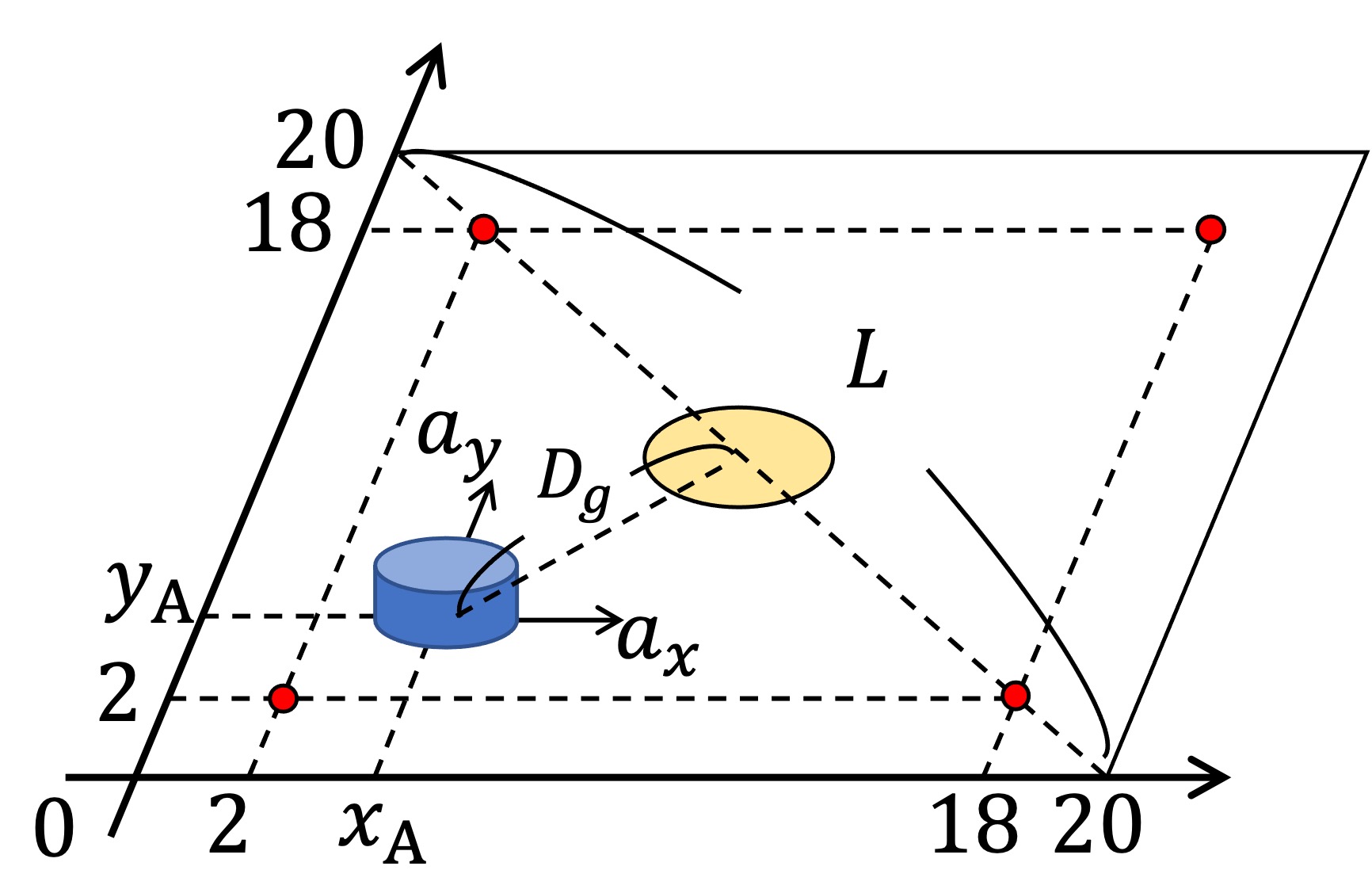}
  \caption{Goal task.
The blue cylinder represents the agent, whose position is denoted by $x^{\mathrm{A}}$ and $y^{\mathrm{A}}$.
The field is surrounded by walls, and $x^{\mathrm{A}}$ and $y^{\mathrm{A}}$ cannot be out of the range from $0$ to $20$.
The agent can move horizontally and vertically within the environment, with a maximum distance of $1$ each.
The yellow circle indicates the goal, and if the agent's center is within this circle, the agent is rewarded and considered to have accomplished the task.
The $4$ red dots indicate the initial positions. At the beginning of each episode, the agent starts at one of these coordinates, with a slight noise added to the position.
}
\label{fig:goal_task}
\end{figure}

\subsection{Goal task}
This study uses a goal task to estimate the agent's learning ability.
Figure \ref{fig:goal_task} shows the task outline.
The task field is inside the range $0$ to $20$ on the x-y plane.
There are $4$ initial positions $(x, y)=(2, 2),(2, 18),(18, 2),(18, 18)$ in the environment.
At the beginning of an episode, the agent is randomly placed at one of the initial positions, with Gaussian noise $\mathcal{N}(0,1)$ added to its position in the field.
The agent obtains an input $\bm{u}$ based on its position $(x^{\mathrm{A}}, y^{\mathrm{A}})$ and uses it to determine its action outputs $-1 \leq a^x \leq 1$ and $-1 \leq a^y \leq 1$.
As a result of the action, the agent's position is updated according to the following equation
\begin{eqnarray}
\label{eq:state_update}
    x^{\mathrm{A}}_{t+1} &=& x^{\mathrm{A}}_{t} + a^x_t, \\
    y^{\mathrm{A}}_{t+1} &=& y^{\mathrm{A}}_{t} + a^y_t,
\end{eqnarray}
where the agent cannot leave the task field and its range of movement is limited to $0 \leq x^{\mathrm{A}} \leq 20$ and $0 \leq x^{\mathrm{A}} \leq 20$.
When the agent is at $(x^{\mathrm{A}}, y^{\mathrm{A}})$, the input $\bm{u}$ from the environment is given by

\begin{eqnarray}
\label{eq:state}
    \bm{u} = \left[ \frac{x^{\mathrm{A}}}{20} \ \ 
    1 - \frac{x^{\mathrm{A}}}{20} \ \ 
    \frac{y^{\mathrm{A}}}{20} \ \ 
    1 - \frac{y^{\mathrm{A}}}{20} \ \ 
    1 - \frac{D_{\mathrm{G}}}{2L} \right]^\top,
\end{eqnarray}
where $D_{\mathrm{G}}$ is the Euclidean distance between the center of the agent and the goal, and $L$ is the length of the diagonal of the task field.
This task is episodic, and an episode ends when the agent either enters the goal circle of radius $2$ (i.e., $D_{\mathrm{G}} < 2$) or fails to reach the goal within $200$ steps.
The agent receives a reward of $r=1$ for reaching the goal, $r=-0.01$ for colliding with the wall surrounding the field, and $r=0$ for any other step.

We also examine a goal change task to estimate the ability to resume exploration and re-learn when the environment changes.
This task initially places the goal at the initial position $G_p^{\mathrm{1}}=(15, 10)$, but changes the position to $G_p^{\mathrm{2}}=(5, 10)$ when the number of steps reaches $N_\mathrm{c}$.
After the goal change, the agent can no longer receive rewards for the behavior it has learned previously, and it needs to resume exploration and learn to reach the new goal to receive rewards again.

\section{Experiment}
\subsection{Conditions}
The reservoir size was set to $N_x=256$ and the recurrent connection probability was set to $p=0.1$.
The input weight matrix $W_\mathrm{in}$ was sampled from a uniform distribution over $[-0.5, 0.5]$.
The critic network $Q$ is a fully connected Multilayer Perceptron (MLP) with two hidden layers consisting of $32$ ReLU nodes, and the output neuron is a linear node.
The initial weights were set using the default settings in PyTorch (version 1.13.1 in this study.).
Both the critic network $Q$ and the actor network $\mu$ are trained using the Adam optimizer, with a learning rate of $5 \times 10^{-4}$.
We paused training every $N_v$ steps and verified the agent's behavior starting from 8 initial positions $(2, 2)$, $(2, 10)$, $(2, 18)$, $(10, 2)$, $(10, 18)$, $(18, 2)$, $(18, 10)$, $(18, 18)$ and slightly different positions (shifted $0.002$ in the $x$ and $y$ axes, respectively). 
The replay buffer size 
was set to $10^6$ and the batch size was set to 64.
The discount factor $\gamma$ was set to 0.95.
The time constant $\tau$ of the target network was set to 0.05.
We set $\epsilon^a=0, \epsilon^t=0$, and the Delayed Policy Updates step to $d=2$.

\begin{figure}[t]
  \begin{minipage}[t]{0.49\linewidth}
    \centering
    \includegraphics[bb=0 0 450 225, scale=0.35]{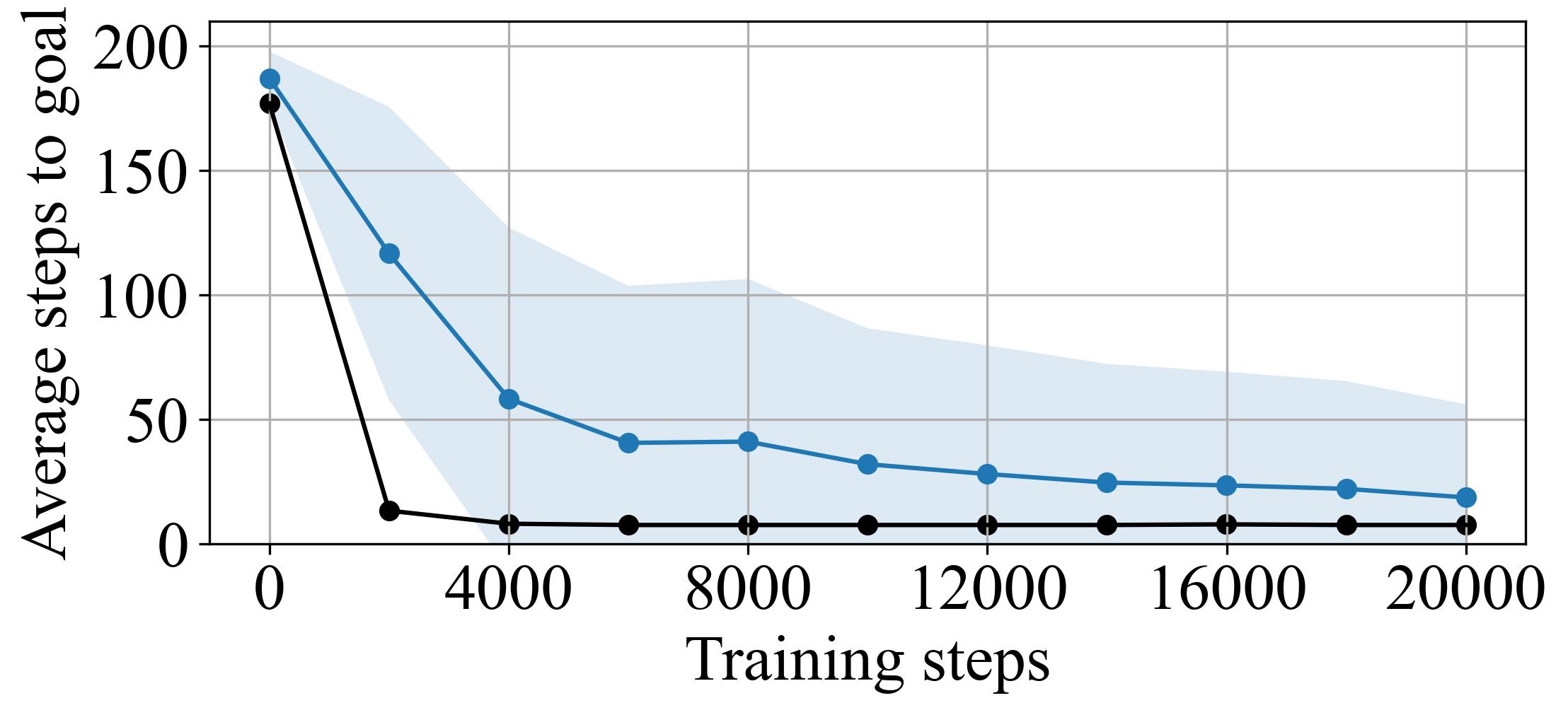}
    \subcaption{TD3-CBRL}
  \end{minipage}
  \begin{minipage}[t]{0.49\linewidth}
    \centering
    \includegraphics[bb=0 0 450 225, scale=0.35]{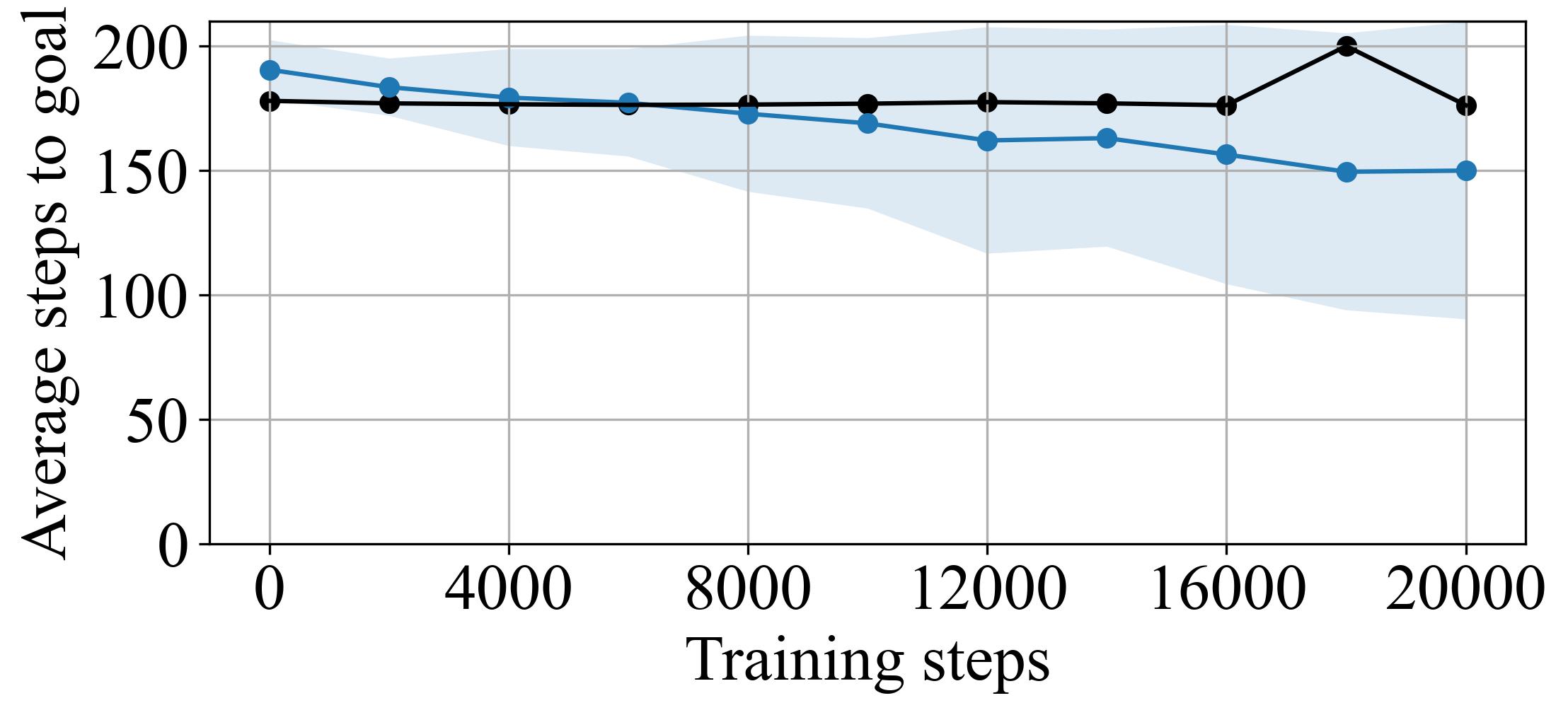}
    \subcaption{TD3 without external exploration noises}
  \end{minipage}
\\ \\ \\
  \begin{minipage}[t]{0.99\linewidth}
    \centering
    \includegraphics[bb=0 0 450 225, scale=0.35]{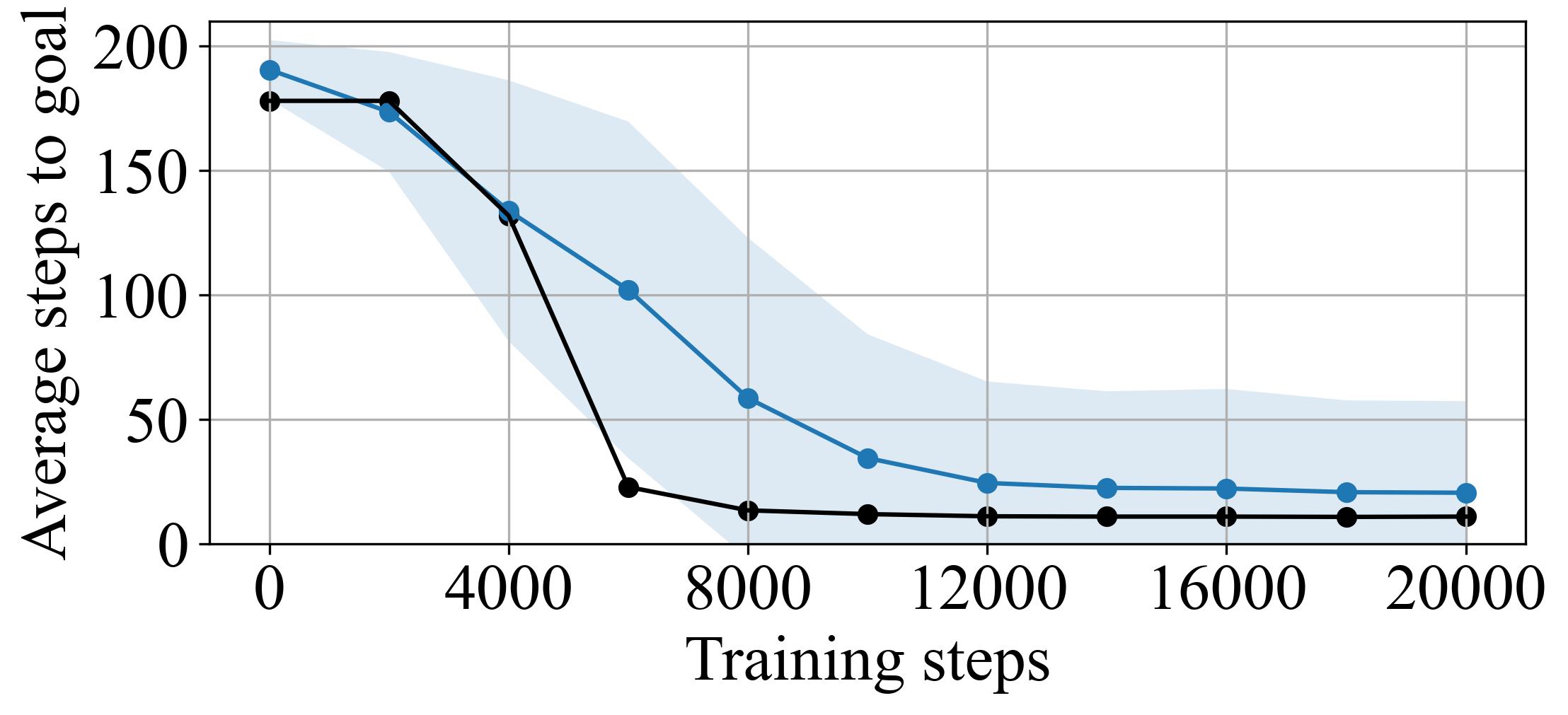}
    \subcaption{Regular TD3}
  \end{minipage}

  \caption{Learning curves.
The vertical axis shows the average number of steps required to reach the goal starting from the 16 initial positions.
The horizontal axis shows the training steps.
The black line shows the representative results with a specific random seed, and the blue line and shaded area show the mean and standard deviation of the steps from the results of experiments with 100 different random number seeds.
(a) shows the learning results with TD3-CBRL.
(b) shows the learning results with TD3 without external exploration noises.
(c) shows the learning results with regular TD3.
}
\label{fig:LC}
\end{figure}

\subsection{Learning result}
\begin{figure}[t]
  \begin{minipage}[t]{0.99\linewidth}
    \centering
    \includegraphics[bb=0 0 1500 300, scale=0.21]{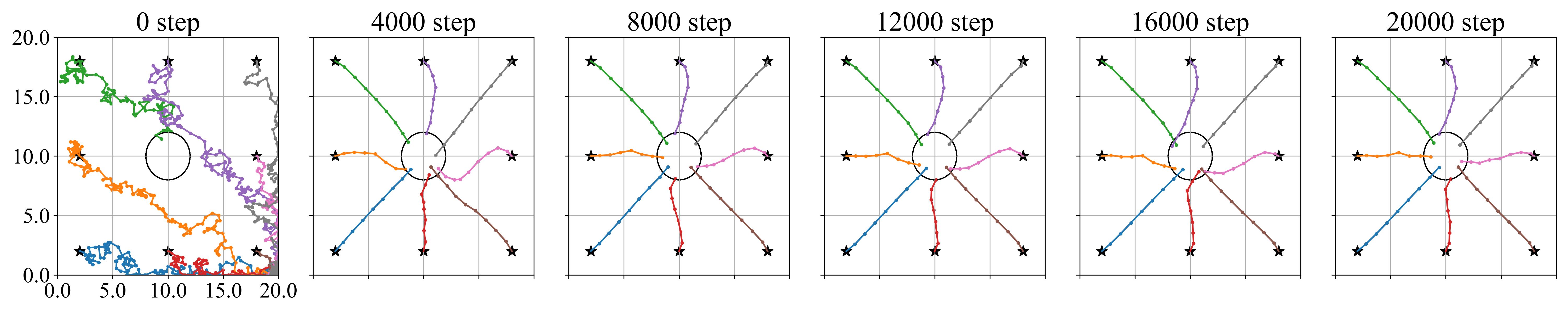}
    \subcaption{TD3-CBRL }
  \end{minipage}
\\
  \begin{minipage}[t]{0.99\linewidth}
    \centering
    \includegraphics[bb=0 0 1500 300, scale=0.21]{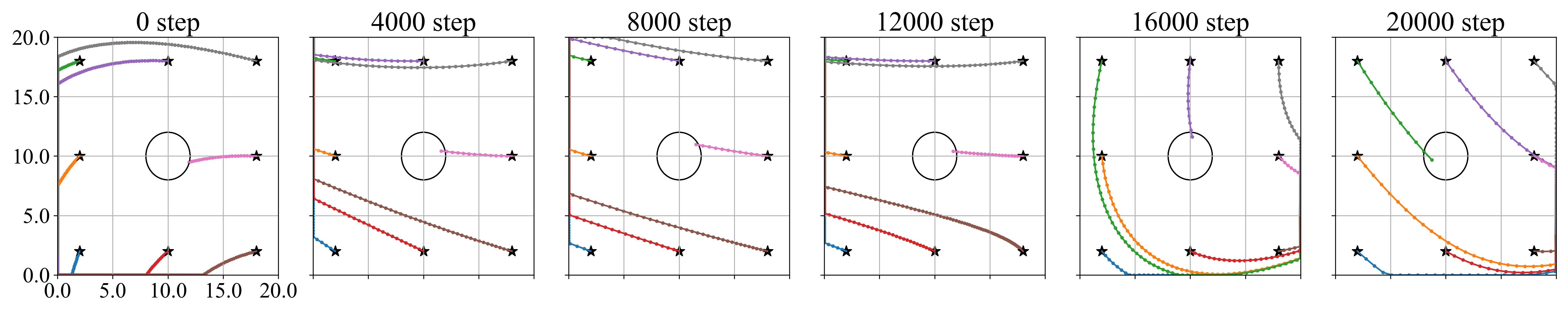}
    \subcaption{TD3 without external exploration noises}
  \end{minipage}
\\ 
  \begin{minipage}[t]{0.99\linewidth}
    \centering
    \includegraphics[bb=0 0 1500 300, scale=0.21]{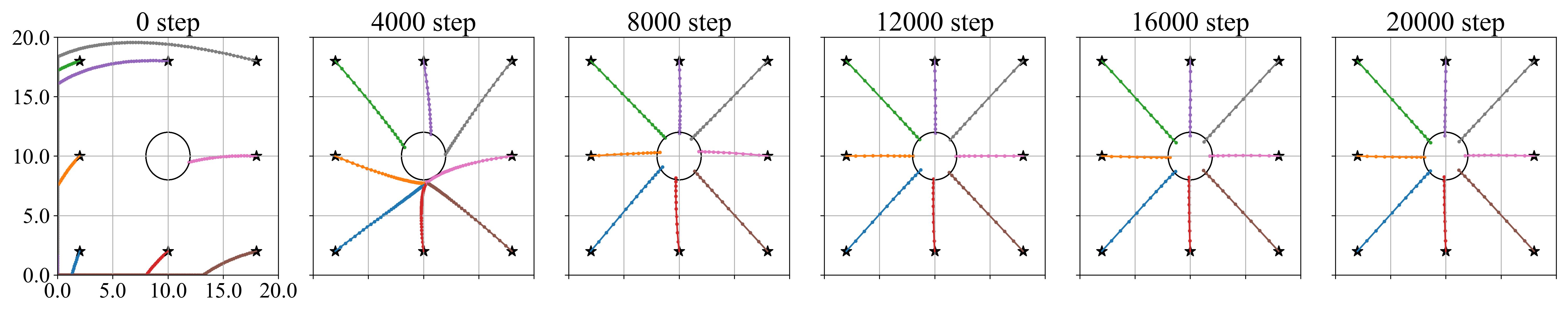}
    \subcaption{Regular TD3}
  \end{minipage}
\\ 
  \begin{minipage}[t]{0.99\linewidth}
    \centering
    \includegraphics[bb=0 0 1500 300, scale=0.21]{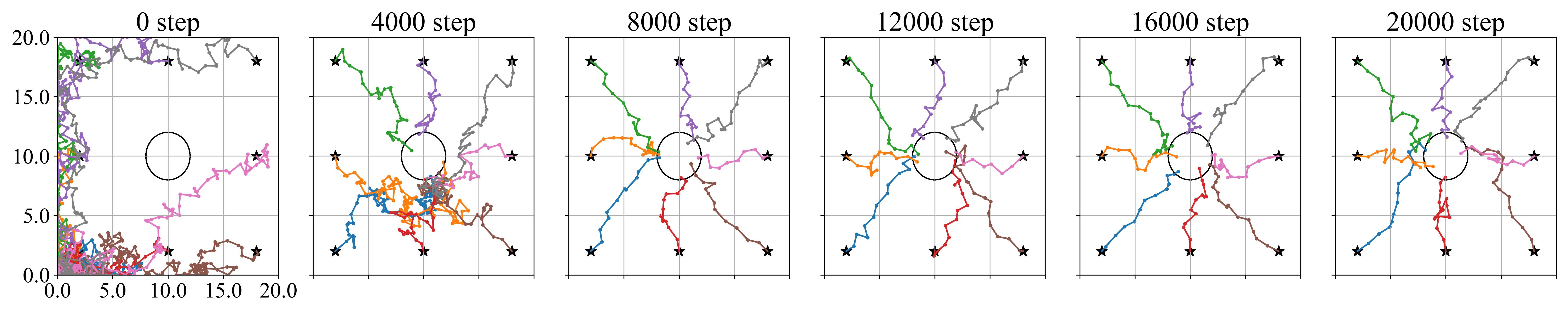}
    \subcaption{Regular TD3 (with external exploration noise in the test phase)}
  \end{minipage}
  \caption{Agent trajectories during the test. Each colored trajectory represents the agent's behavior from the $8$ initial test positions. Each graph in the subfigures shows the test results at $0$, $4000$, $8000$, $12000$, $16000$, and $20000$ training steps. The random seed is the same as the one used for the representative results in Fig. \ref{fig:LC}. (a) shows the training results for TD3-CBRL. (b) shows the results of TD3 without exploration by random noises. (c) shows the results with regular TD3. (d) shows the behavior of the same agent as in (c) when the exploration by random noises is not stopped during the test.
}
\label{fig:POS}
\end{figure}

The TD3-CBRL agent learned the goal task in $20000$ steps.
Figure \ref{fig:LC}(a) shows the learning curve resulting from the test conducted every $N_v=2000$ step.
This figure shows that the average number of steps required by the agent to reach the goal decreases as the learning progresses.
This result indicates that TD3-CBRL can successfully learn the goal task.
Figure \ref{fig:POS}(a) shows the trajectories of the agent's movement in the environment for each test trial.
This figure shows that in the initial stage of learning ($0$ step), the agent acts exploratively in the environment.
On the other hand, in the trajectories after $4000$ steps, the agent is moving toward the goal from each initial position.
We also see that the exploratory behavior subsides as the learning progresses.
It is important to note that no external random noises for exploration were added to the agent's action output during the learning process.
This result indicates that the internal dynamics of the reservoir caused spontaneous exploratory behavior, and as learning progressed, such variability in the output was autonomously suppressed.

\begin{figure}[t]
  \begin{minipage}[t]{0.49\linewidth}
    \centering
    \includegraphics[bb=0 0 500 300, scale=0.35]{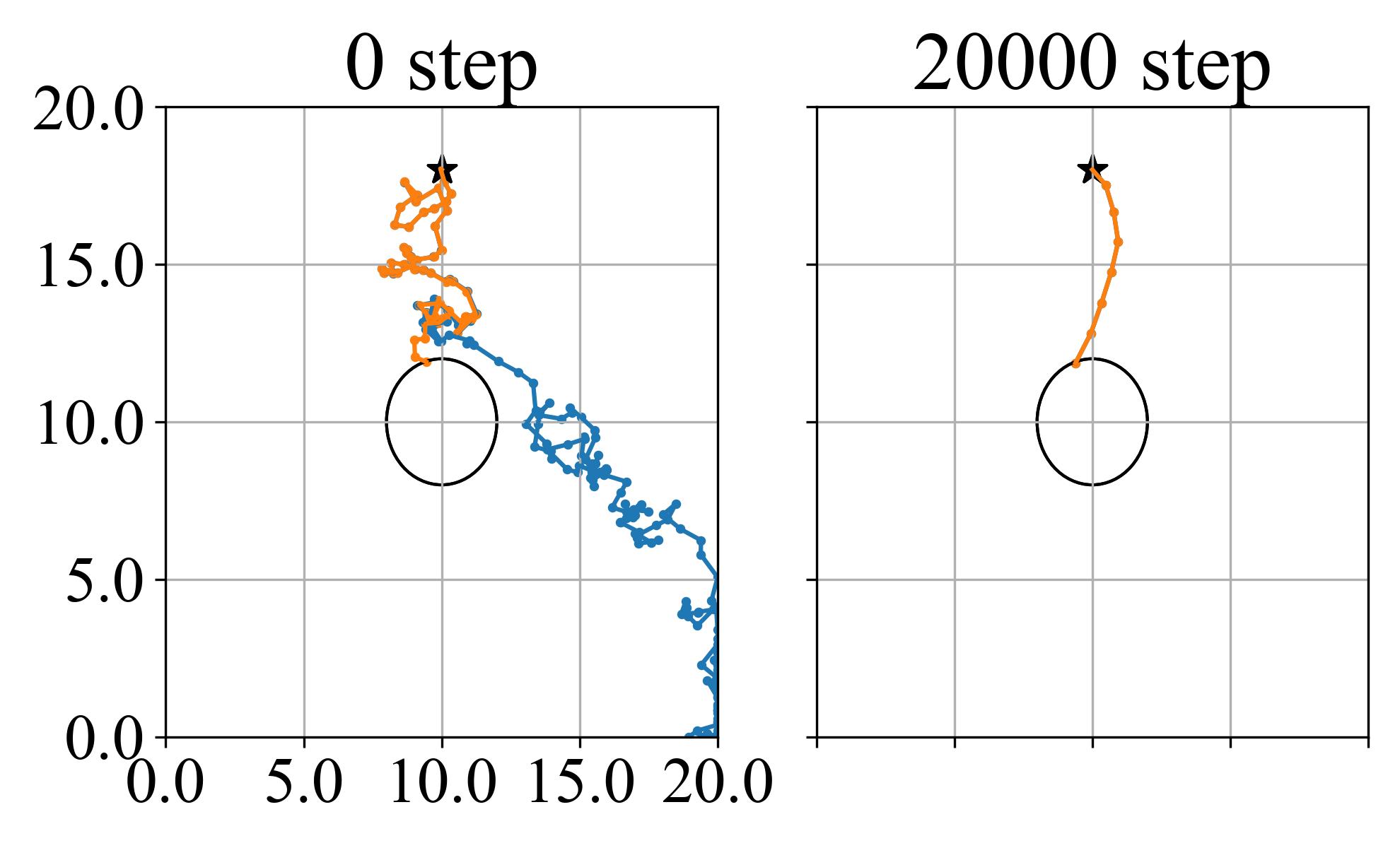}
  \subcaption{TD3-CBRL
}
  \end{minipage}
  \begin{minipage}[t]{0.49\linewidth}
    \centering
    \includegraphics[bb=0 0 500 300, scale=0.35]{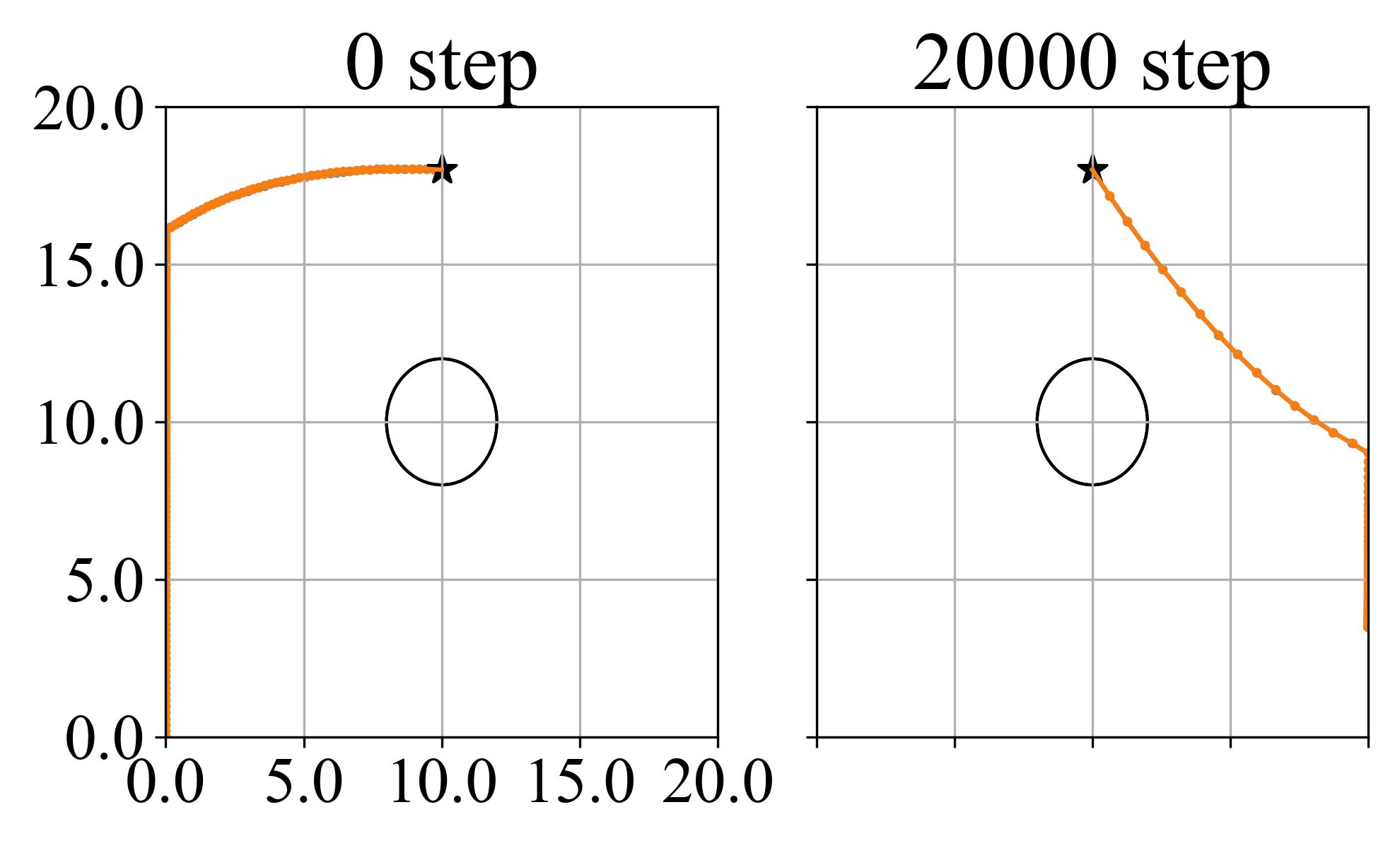}
  \subcaption{TD3 without external exploration noise
}
      \end{minipage}
  \caption{Sensitivity of agent trajectories for initial position. The blue and orange lines show the agent's trajectory when the agent started from (10, 18) and from a slightly shifted initial position (10.002, 18.002), respectively. Each graph in the subfigure shows the test results at $0$
  and $20000$ training steps.
}
\label{fig:sensitivity}
\end{figure}

Figure \ref{fig:sensitivity} (a) shows the results of investigating the sensitivity of the agent's behavior to variations of the initial positions in this case.
This figure shows the trajectories of agents starting from slightly different initial positions.
At the $0$ step, the agent, starting from its original initial position, continues toward the wall and ends the episode without reaching the goal. On the other hand, although the agent starting from the slightly shifted position behaves like the agent starting from the original initial position initially, it leaves the original trajectory after a while and eventually reaches the goal.
This result indicates that in the early phases of learning, a slight difference in the initial position can significantly change the agent's behavior and even determine the task's success or failure.
At the $20000$ steps, the trajectories starting from two different initial positions reach the goal before they diverge.

\subsection{Effects by presence or absence of exploration component}
To confirm that the chaoticity of the model contributes to the agent's exploration, we evaluated the learning performance when the model does not have chaotic dynamics.
Specifically, instead of an actor network with a chaotic reservoir, we used a Multilayer Perceptron (MLP) with one hidden layer consisting of 256 $\tanh$ neurons, as shown in Fig. \ref{fig:td3_cbrls} (b).
In this case, the learning rate of the actor network was readjusted to $1.6 \times 10^{-5}$.
Figure \ref{fig:LC}(b) shows the learning curve for the MLP case without exploration random noise.
This figure shows that the number of steps to the goal scarcely decrease and that the agent failed to learn the goal task.
Figure \ref{fig:POS}(b) and Fig. \ref{fig:sensitivity} (b) show the agent trajectory during the test phase and the sensitivity test under this condition.
These figures show that the agent lacked exploratory behavior and failed to learn.
This result appears to be due to the fact that the learning model did not have the dynamics to generate spontaneous activity, and thus, the exploration did not occur without random noises.

To clarify that the absence of exploration random noise is the direct cause of the learning failure, we verified the case where the MLP is trained with random noise for exploration in the same way as in the regular TD3, as shown in Fig. \ref{fig:td3_cbrls} (c).
Here, $\epsilon^a$ is sampled from $\mathcal{N}(0, 0.5^2)$, and $\epsilon^t$ is sampled from $\mathcal{N}(0, 0.2^2)$ clipped to the range from -1 to 1.
Figure \ref{fig:LC}(c) shows the learning curve under this condition.
This figure shows that the number of steps to reach the goal decreased and that the agent succeeded in learning the task.
Figure \ref{fig:POS}(c) shows the trajectories of the agent's behavior in this validation.
This figure shows that the agent can learn the behavior of moving toward the goal due to exploration with random numbers during the learning process. Note that during the test phase, adding random numbers to the action outputs is stopped.
These results indicate that the presence or absence of exploration by random numbers has a significant influence on the success or failure of learning and that the TD3-CBRL agent successfully learns the goal task through exploration driven by the chaotic dynamics of the reservoir.

Comparing (a) and (c) in Fig. \ref{fig:POS}, the regular TD3 agent can go to the goal in a straighter path than the TD3-CBRL agent.
However, this is due to an external intervention that removes the random noise during the test phase.
Figure \ref{fig:POS}(d) shows the result when the agent in (c) acts with the exploration random noise during the test phase.
If the exploration random noise is not eliminated, the agent cannot reach the goal in a straight path.
On the other hand, the TD3-CBRL agent can autonomously reduce the variability driven by its chaoticity as its learning progresses.

\begin{figure}[t]
  \begin{minipage}[t]{0.49\linewidth}
    \centering
    \includegraphics[bb=0 0 450 225, scale=0.35]{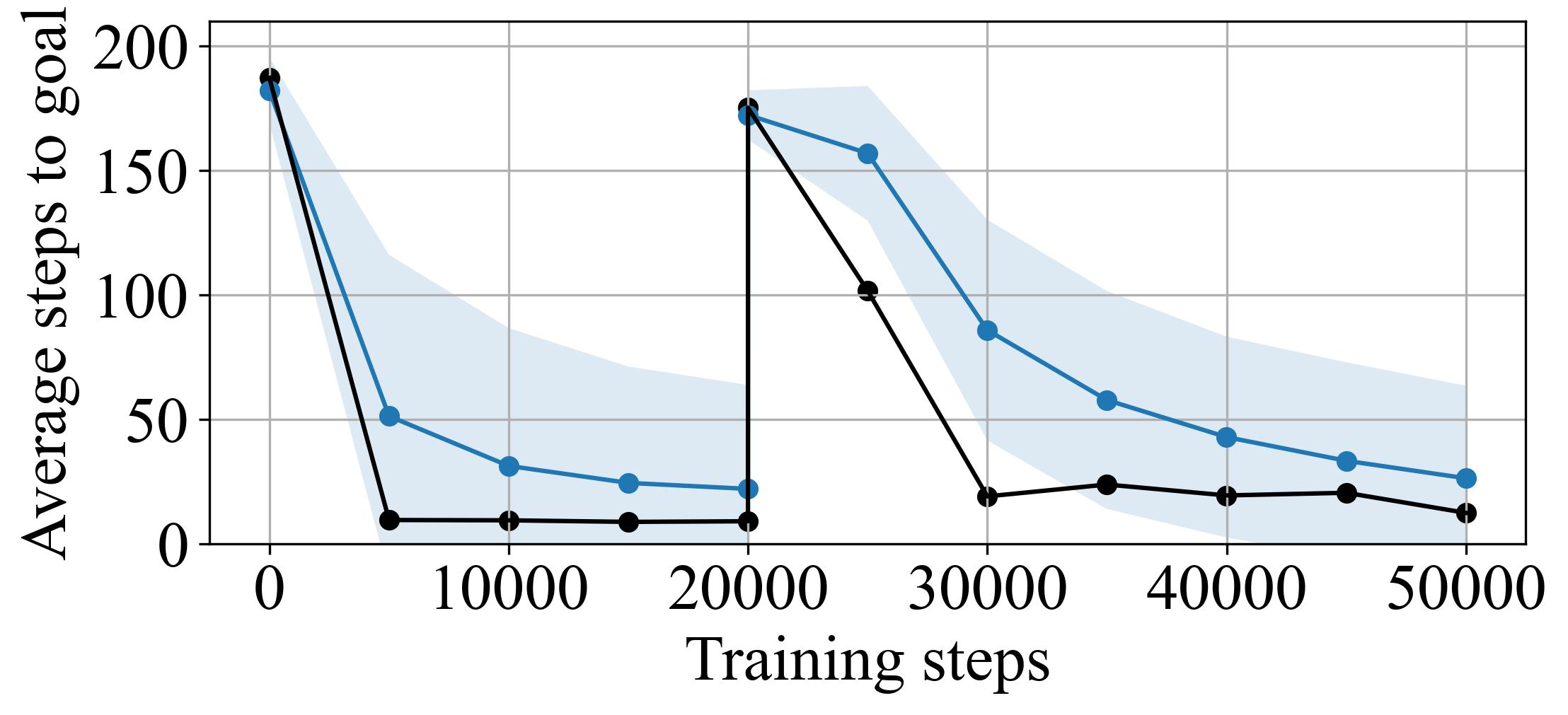}
  \subcaption{Regular case $(g=2.2)$.
}
  \end{minipage}
  \begin{minipage}[t]{0.49\linewidth}
    \centering
    \includegraphics[bb=0 0 450 225, scale=0.35]{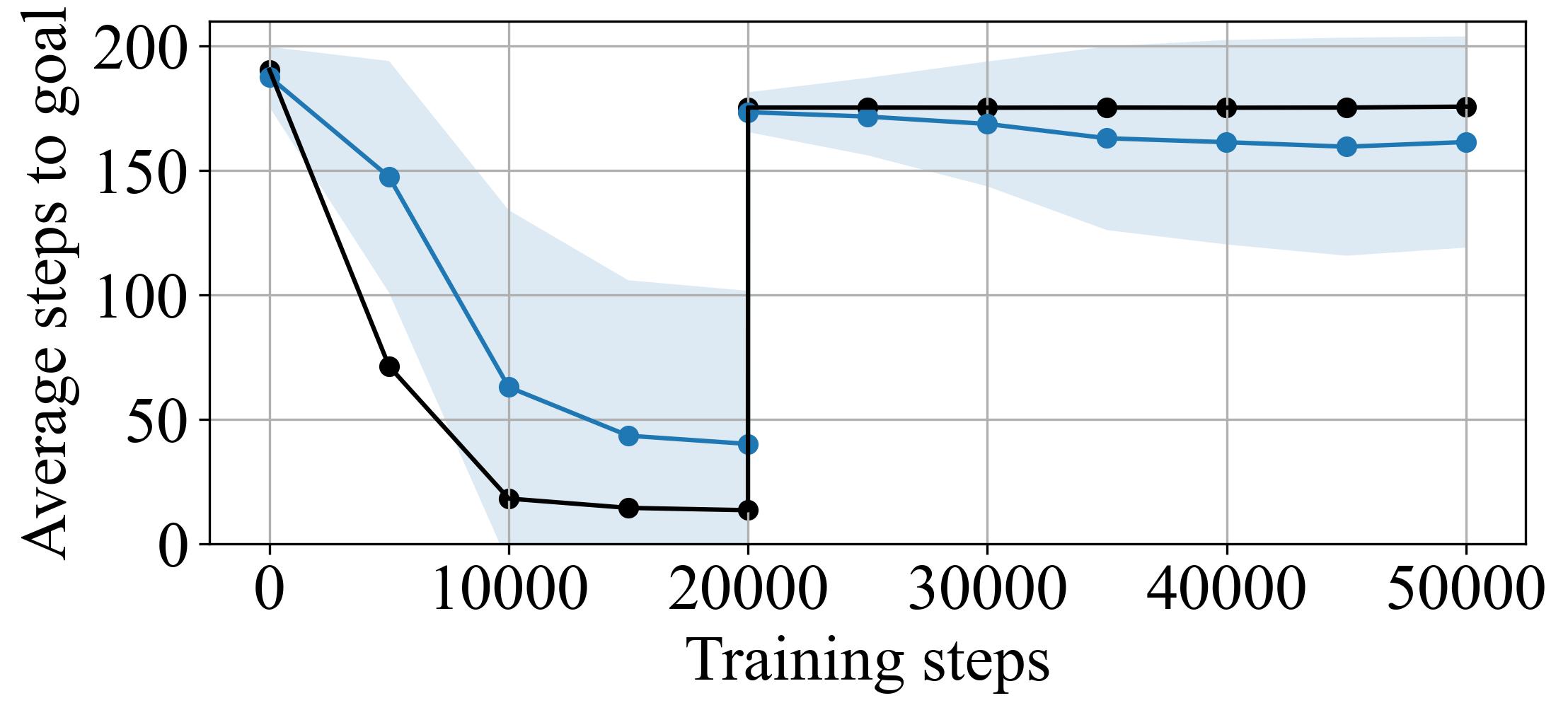}
  \subcaption{Regular TD3.
}
      \end{minipage}
\\ \\ \\
  \begin{minipage}[t]{0.99\linewidth}
    \centering
    \includegraphics[bb=0 0 450 225, scale=0.35]{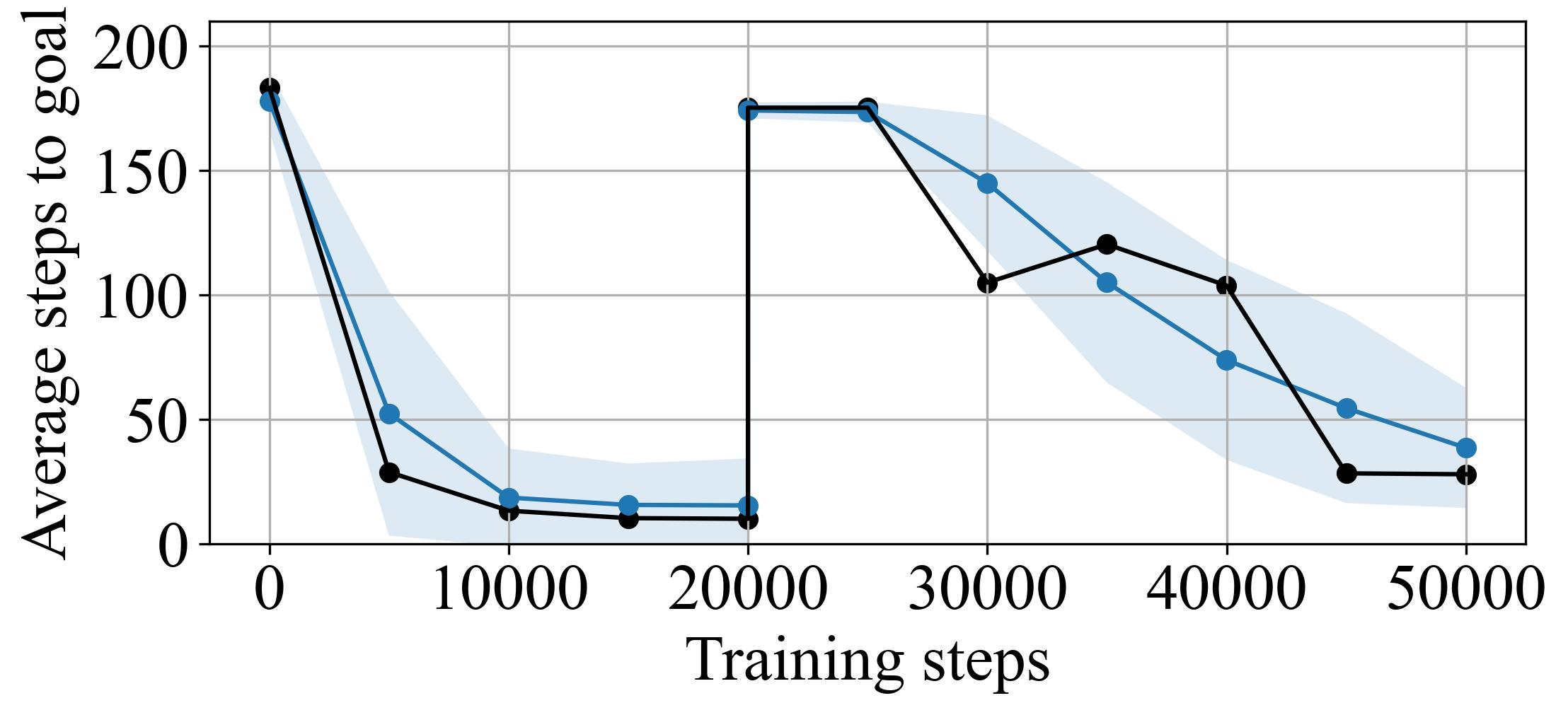}
  \subcaption{Larger spectral radius case $(g=5)$.
}
      \end{minipage}
  \caption{Learning curves of goal change task.
  The definitions of line colors are the same as in Fig. \ref{fig:LC}.
}
\label{fig:LC_goal_change}
\end{figure}

\subsection{Goal change task.}
\label{sec:goal_change_task}
Previous studies have shown that when the environment changes and the agent cannot be rewarded with previously learned behaviors, CBRL agents can autonomously resume their exploration and learn to adapt to the new environment \cite{shibata2015}.
These results suggest that CBRL agents have the flexibility to adapt to dynamic and uncertain environments.
Here, we observed the agent's response to changing the goal position to test whether TD3-CBRL agents can re-learn when the task rule changes.

In the goal change task, the goal was initially placed at $G_p^{\mathrm{1}}=(15, 10)$, and learning was performed for $20000$ steps. Then, at the $N_\mathrm{c}=20001$ steps, the goal position was changed to $G_p^{\mathrm{2}}=(5, 10)$, and learning continued in the new environment for another $30000$ steps. The test is conducted every $N_v=5000$ steps. The learning curve under these conditions is shown in Fig. \ref{fig:LC_goal_change}(a).
This figure shows that when the goal position is changed, the number of steps required to reach the goal, which had decreased during learning in the initial environment, temporarily increases. However, as learning in the new environment progresses, the number of steps decreases again. This result indicates that the agent is adapting to the new environment.

\begin{figure}[t]
  \begin{minipage}[t]{0.99\linewidth}
    \centering
    \includegraphics[bb=0 0 1500 600, scale=0.25]{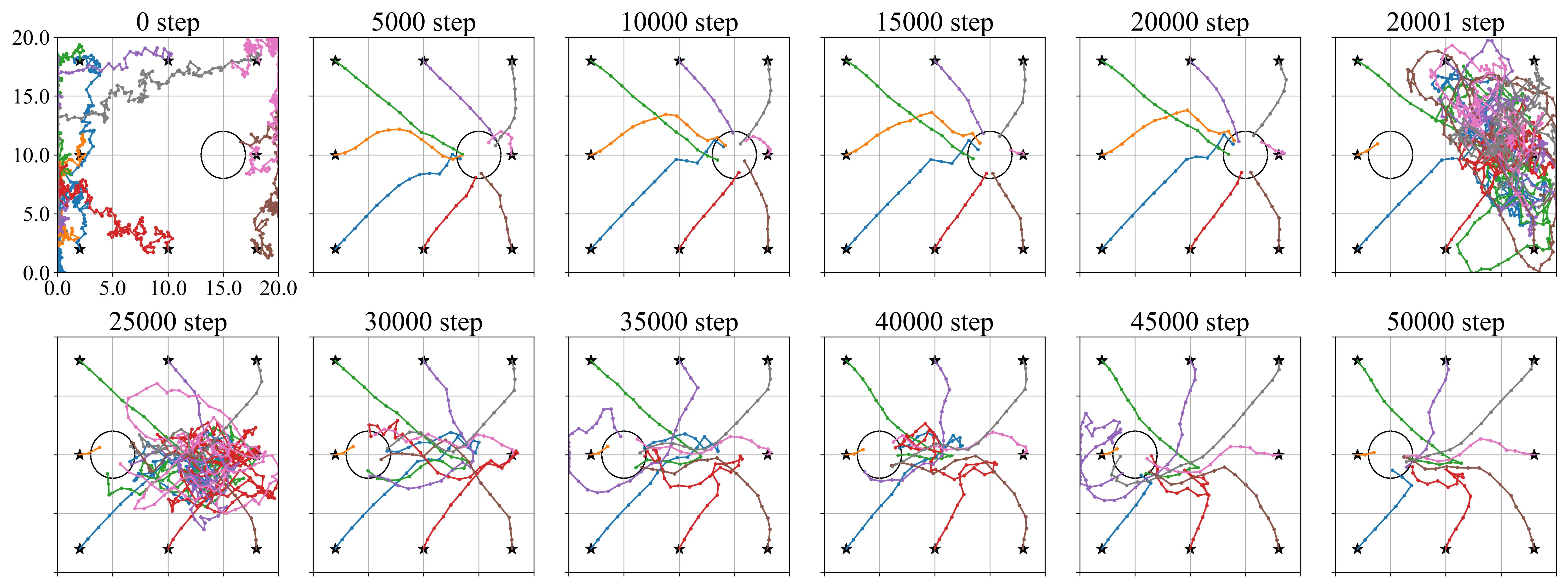}
    \subcaption{Regular case $(g=2.2)$}
  \end{minipage}
\\
  \begin{minipage}[t]{0.99\linewidth}
    \centering
    \includegraphics[bb=0 0 1500 600, scale=0.25]{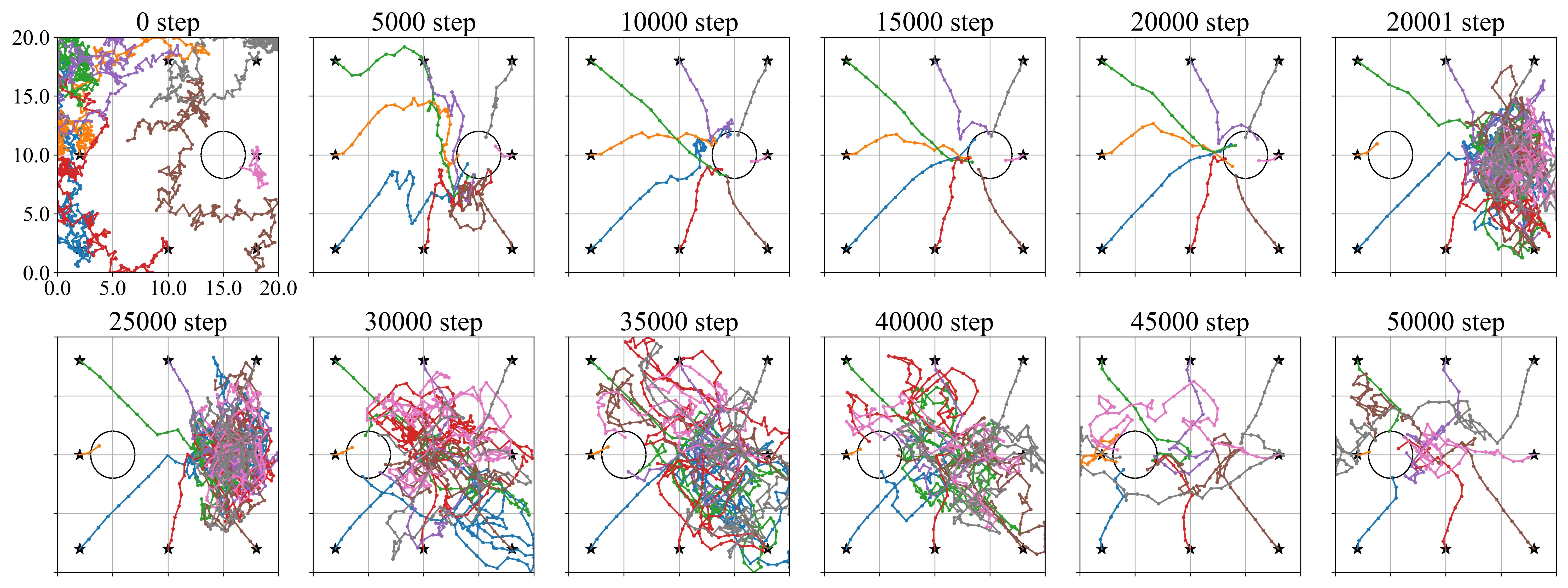}
    \subcaption{Larger spectral radius case $(g=5)$}
  \end{minipage}
  \caption{Agent trajectories during the test of the goal change task. Each graph in the subfigures shows the test results conducted every 5000 training steps. The definitions of the line colors are the same as in Fig. \ref{fig:POS}.
}
\label{fig:POS_goal_change}
\end{figure}

The trajectories of the agent in the environment under these conditions are shown in Fig. \ref{fig:POS_goal_change}(a).
This figure shows that at the $25000$ steps, the agent autonomously resumes its exploration with internal chaotic dynamics due to the change in the goal position. After that, the agent gradually adapts its behavior to move toward the new goal.

Figure \ref{fig:LC_goal_change} (b) shows the results of applying the goal change task to the regular TD3 agent. This figure demonstrates that the regular TD3 agent fails to re-learn the new goal. While there is a possibility of task dependence, it suggests that TD3-CBRL agents exhibit greater flexibility in adapting to new environments compared to regular TD3 agents in goal change task. Two hypotheses can be proposed to explain this advantage of the CBRL agents. First, the short-term memory capability of the reservoir allows the agent to distinguish between the state after and before reaching the initial goal area. Second, the exploratory behavior driven by the chaotic dynamics of the reservoir is more effective than exploration based on simply adding random noise to the actions.

\begin{figure}[t]
  \begin{minipage}[t]{0.49\linewidth}
    \centering
    \includegraphics[bb=0 0 600 300, scale=0.28]{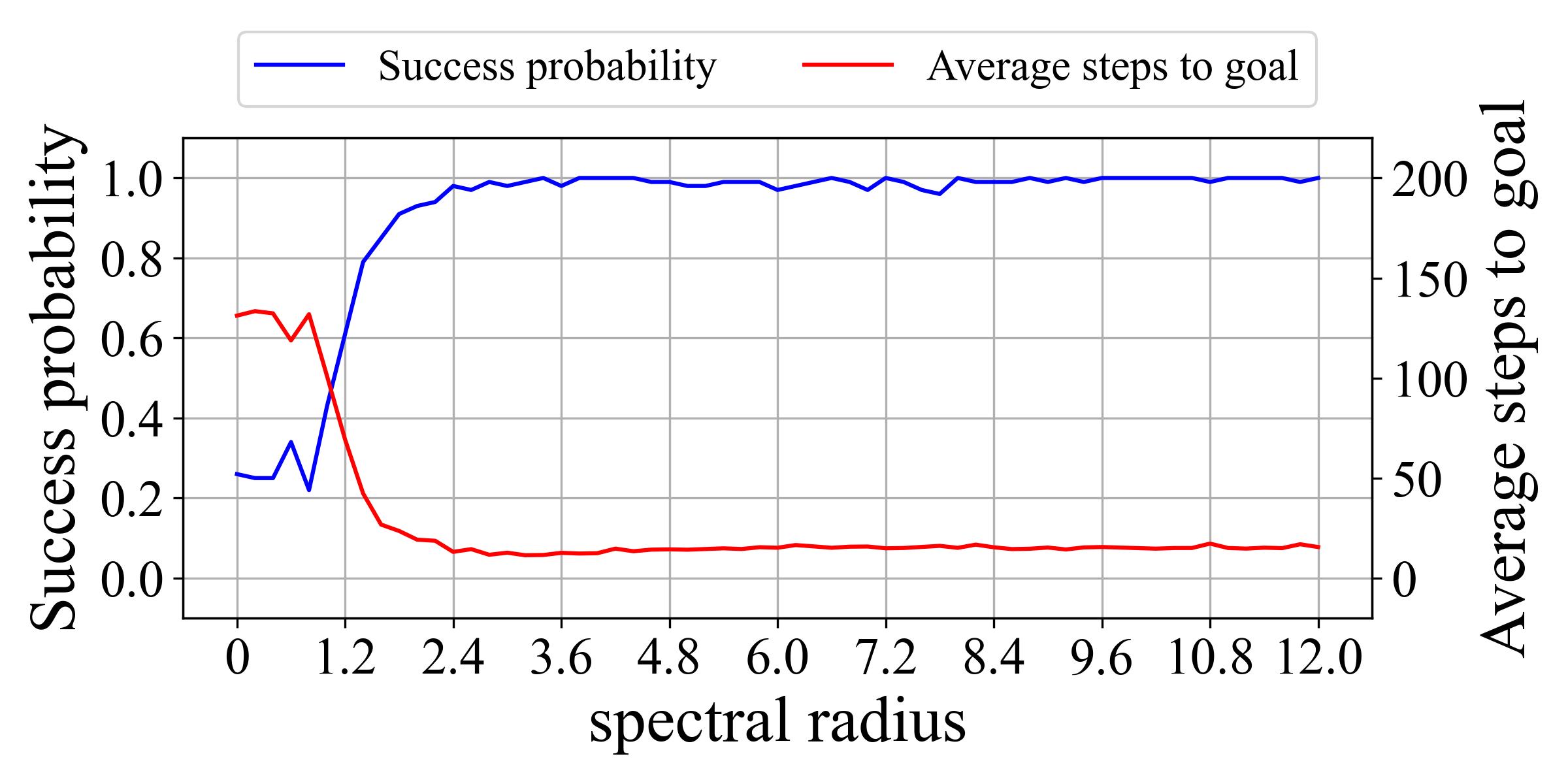}
    \subcaption{Goal task.}
  \end{minipage}
  \begin{minipage}[t]{0.49\linewidth}
    \centering
    \includegraphics[bb=0 0 600 300, scale=0.28]{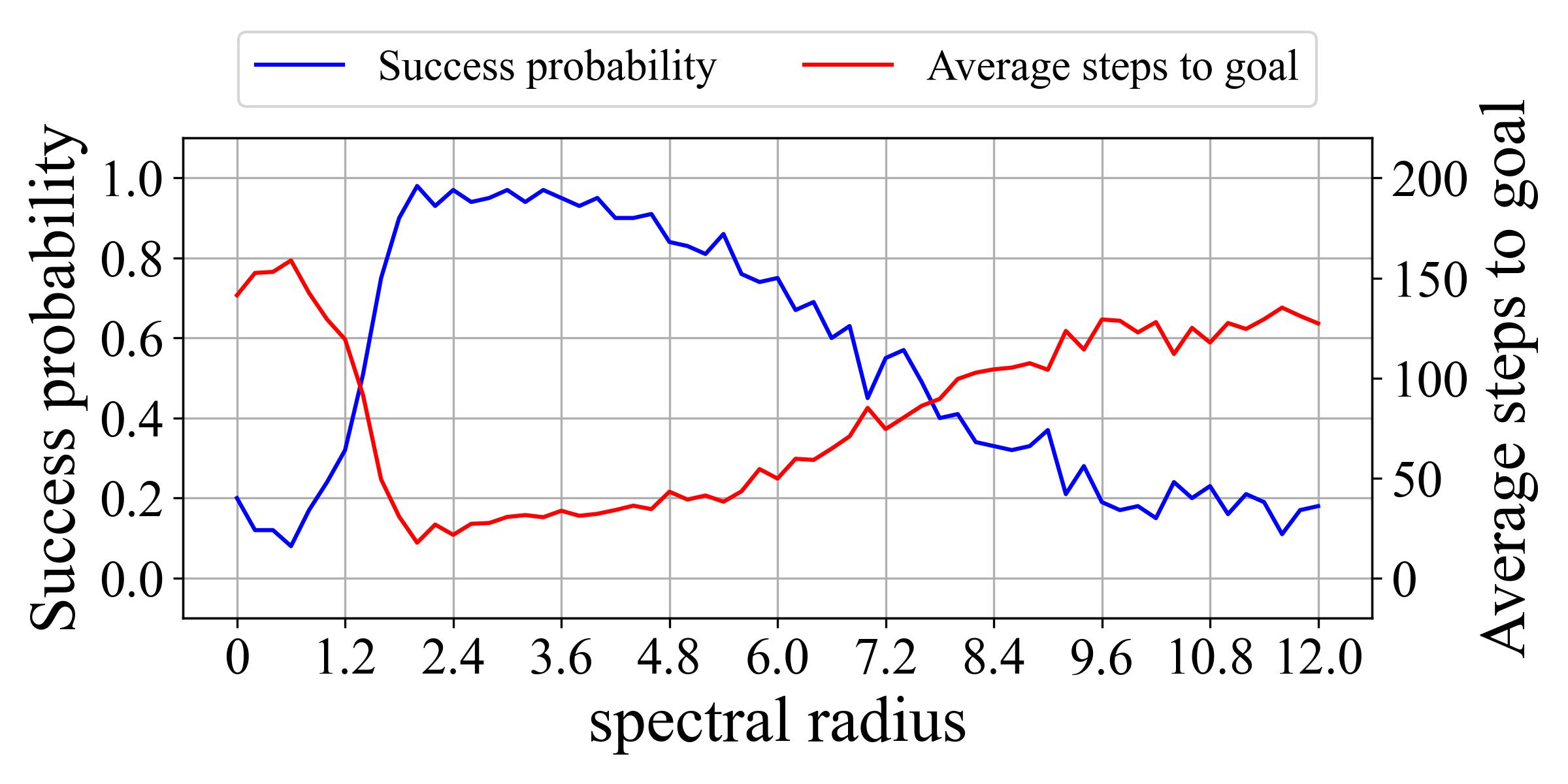}
    \subcaption{Goal change task.}
  \end{minipage}
\\ \\ 
  \begin{minipage}[t]{0.49\linewidth}
    \centering
    \includegraphics[bb=0 0 600 300, scale=0.28]{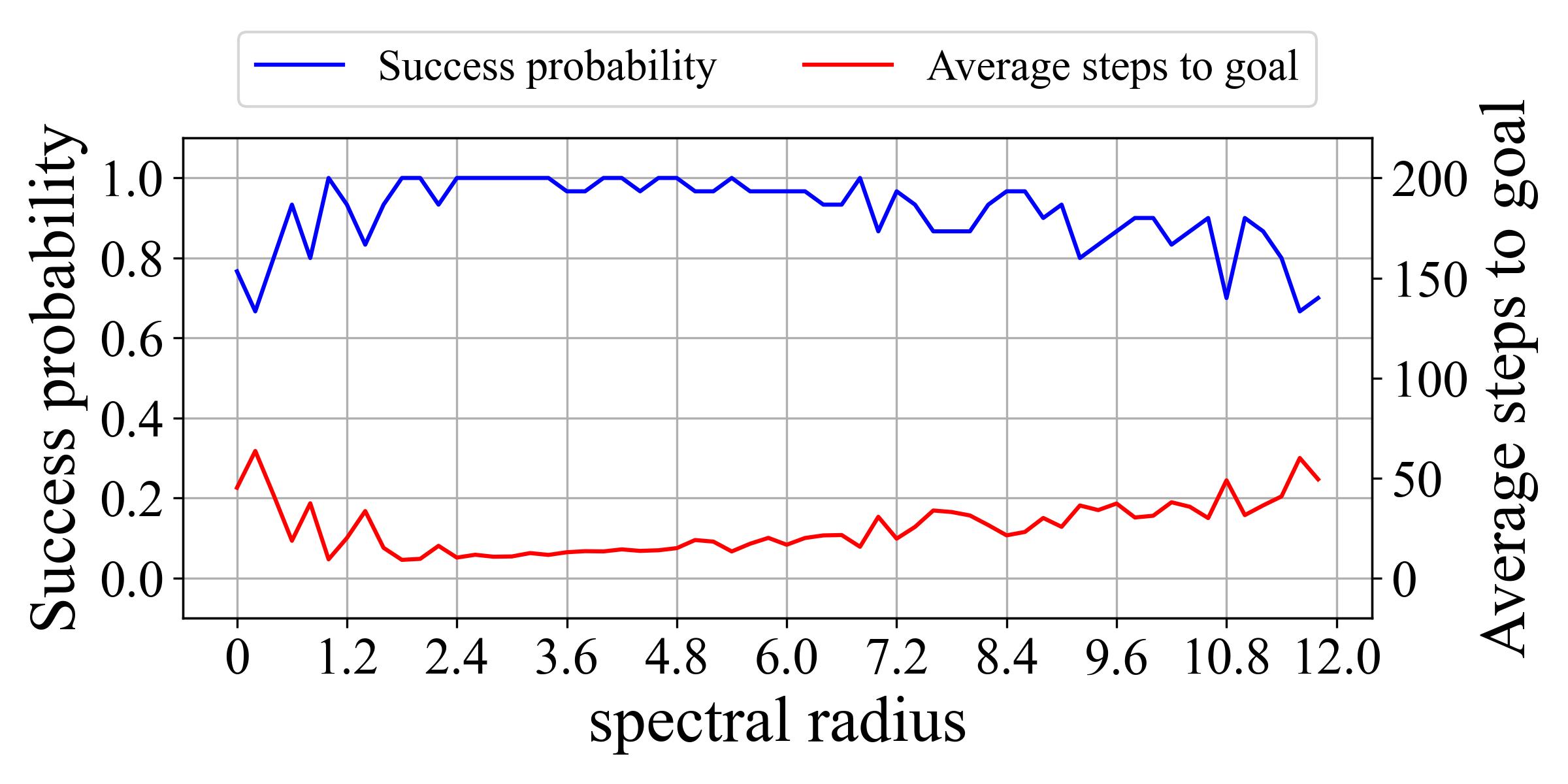}
    \subcaption{Goal change task with long-term learning.}
  \end{minipage}
  \begin{minipage}[t]{0.49\linewidth}
    \centering
    \includegraphics[bb=0 0 600 300, scale=0.28]{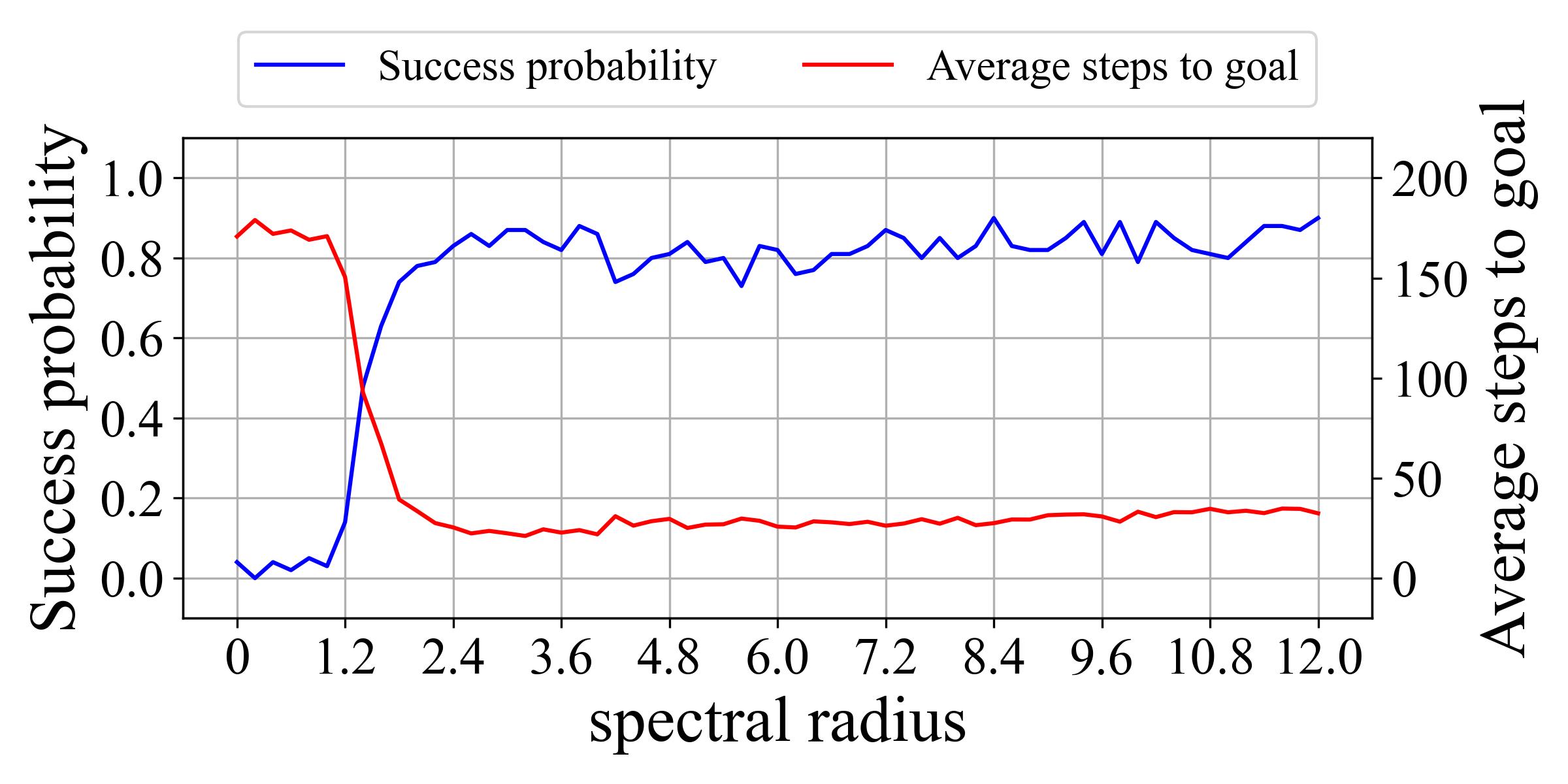}
    \subcaption{Goal change task with a replay buffer size of 64.}
  \end{minipage}
\\ \\ 
  \begin{minipage}[t]{0.49\linewidth}
    \centering
    \includegraphics[bb=0 0 600 300, scale=0.28]{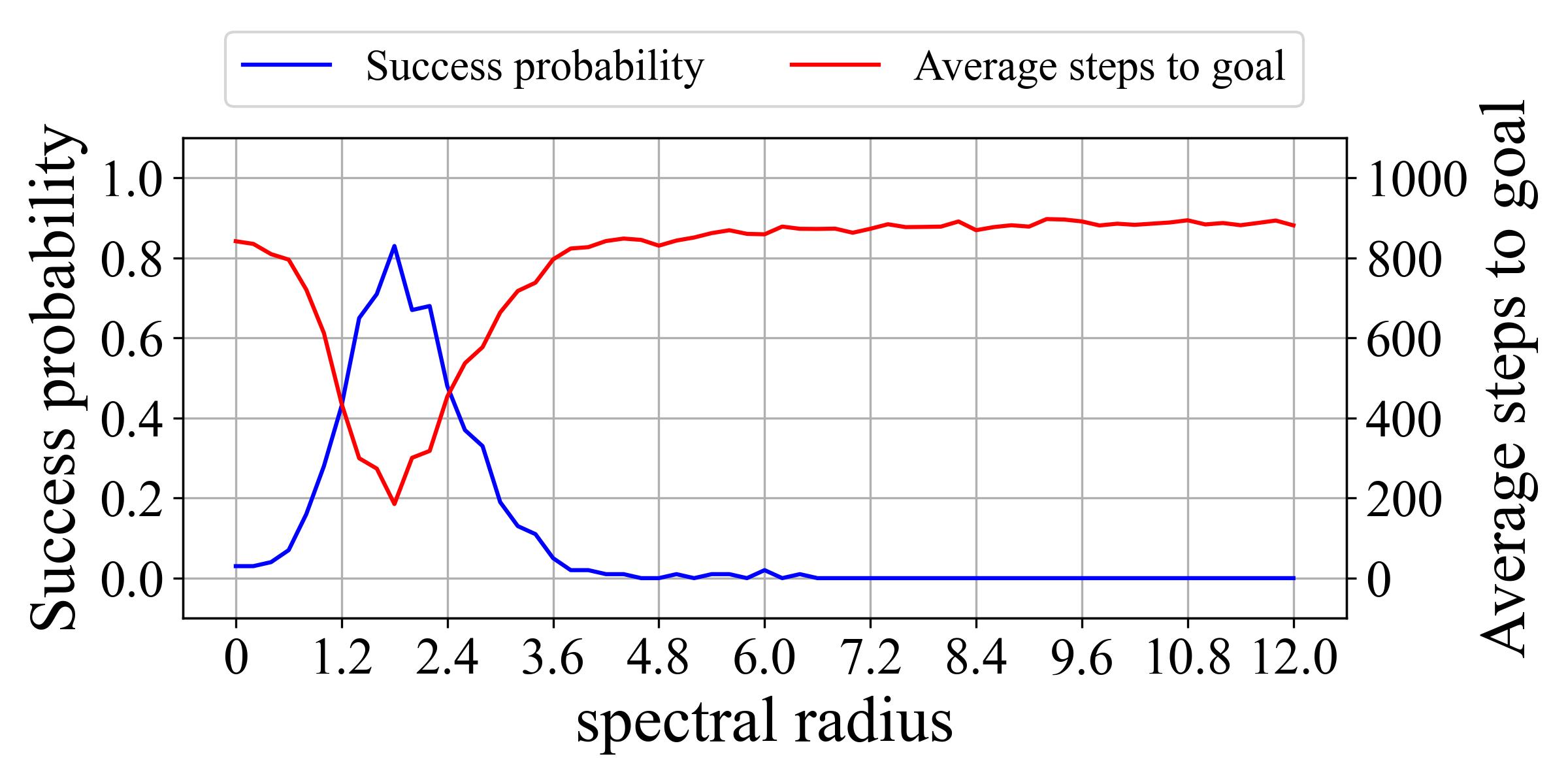}
    \subcaption{Goal task with expanded task field.}
  \end{minipage}
  \begin{minipage}[t]{0.49\linewidth}
    \centering
    \includegraphics[bb=0 0 600 300, scale=0.28]{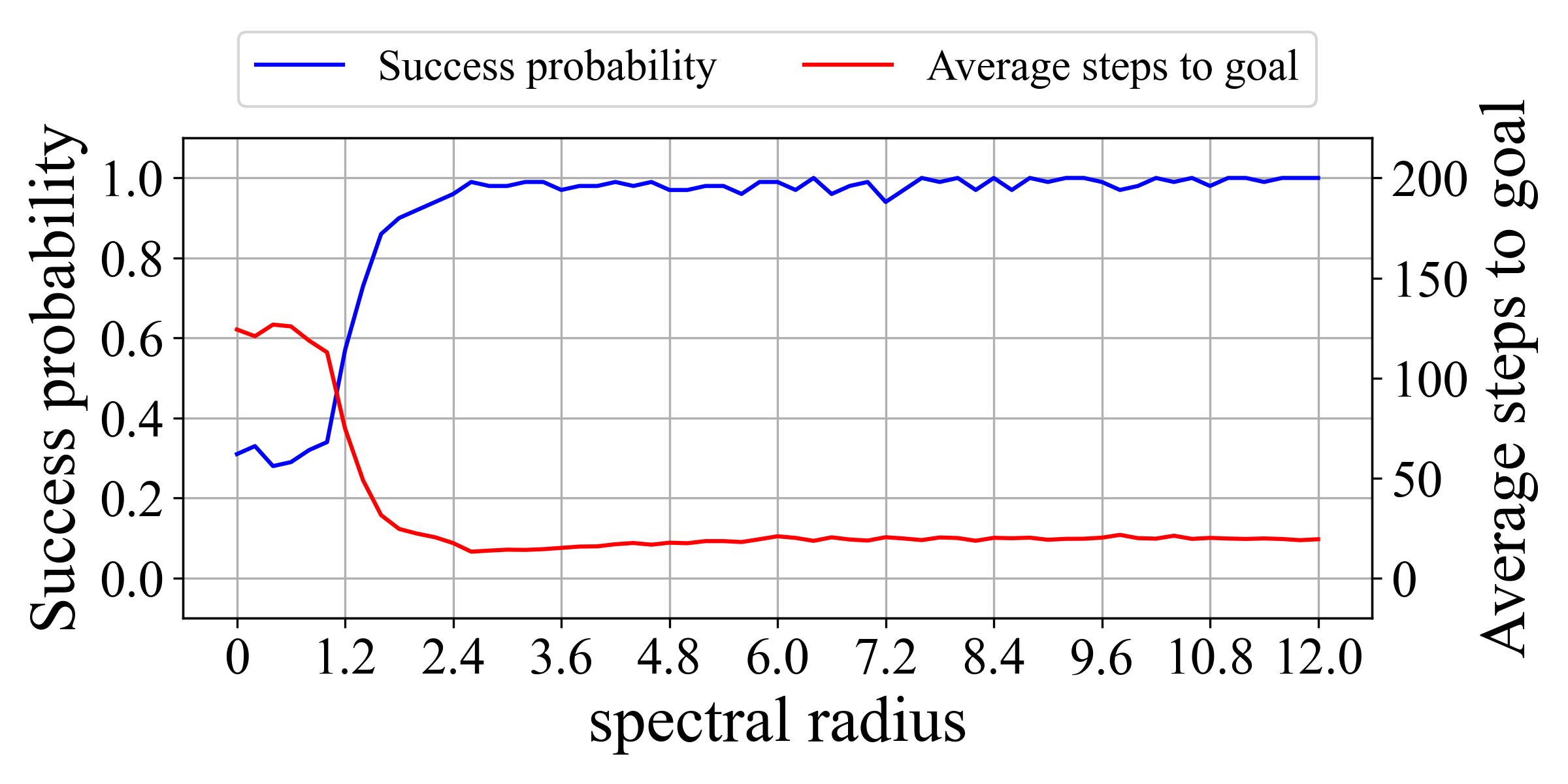}
    \subcaption{Goal task with observation noises.}
  \end{minipage}
  \caption{Learning performance with varying spectral radius. The blue line shows the successful learning probability, indicated on the left vertical axis. The red line shows the average number of steps required to reach the goal, indicated on the right vertical axis. The horizontal axis shows the value of the spectral radius. (a) shows the results for the goal task. (b) shows the results of the goal change task. (c) shows the results of learning over a long period of 180000 steps after the goal change. (d) shows the results with a replay buffer size of $64$. (e) shows the results for the goal task whose task field was expanded from the range $[0, 20]$ to $[0, 100]$ in the x-y plane. (f) shows the results for the goal task with observation noises.
}
\label{fig:sr_score}
\end{figure}

\subsection{Learning performance and chaoticity}
\label{sec:performance_chaoticity}
We investigated the effect of the chaoticity of the system on learning performance.
The chaoticity of the reservoir network can be tuned by changing the spectral radius of the reservoir's recurrent weight matrix using the parameter $g$.
We changed the value of $g$ from $0$ to $12$ in $0.2$ increments and conducted trials with 100 different random number seeds under each condition to obtain the successful learning probability and the average number of steps to reach the goal. 
Here, success is defined as an event when the agent reaches the goal from all $16$ initial positions during the final test.
The results are shown in Fig. \ref{fig:sr_score}(a).
This figure shows that learning performance begins to improve as $g$ exceeds 
around $1$, and learning becomes successful in most cases when $g$ exceeds approximately $2$. This result indicates that the chaoticity of the system is essential for successful exploration and learning. The results also indicate that the probability of success remains high even when the value of $g$ is further increased.
In general, there is an appropriate range for the parameter $g$ for time series processing in reservoir computing. It can be considered that since this study does not target tasks that require memory and the task is simple, the success probability of learning by TD3-CBRL remains stable even when $g$ is larger than the typical critical value.

We conducted the same experiment with a goal change task to examine how the flexibility of switching between exploration and exploitation was affected by the value of $g$.
The task settings are the same as in Section \ref{sec:goal_change_task} except for $g$. The results are shown in Fig. \ref{fig:sr_score}(b).
This figure shows that performance on the goal change task decreases with an extremely large $g$. 
These results indicate that choosing an appropriate $g$ value is still essential in CBRL using reservoir networks.

We experimented with long-term re-learning to investigate whether a larger $g$ makes the agent unable to re-learn or increases the required steps. Specifically, after changing the goal position at the $N_\mathrm{c} = 20001$ steps, the agent learned for $180000$ steps in the new environment. The results are shown in Fig. \ref{fig:sr_score}(c).
This figure shows that long-term learning mitigated the decrease in learning performance when $g$ is large. This result indicates that a model with a larger $g$ and stronger chaoticity requires more re-learning steps, consequently making re-learning more difficult.

\begin{figure}[t]
  \begin{minipage}[t]{0.49\linewidth}
    \centering
    \includegraphics[bb=0 0 600 300, scale=0.35]{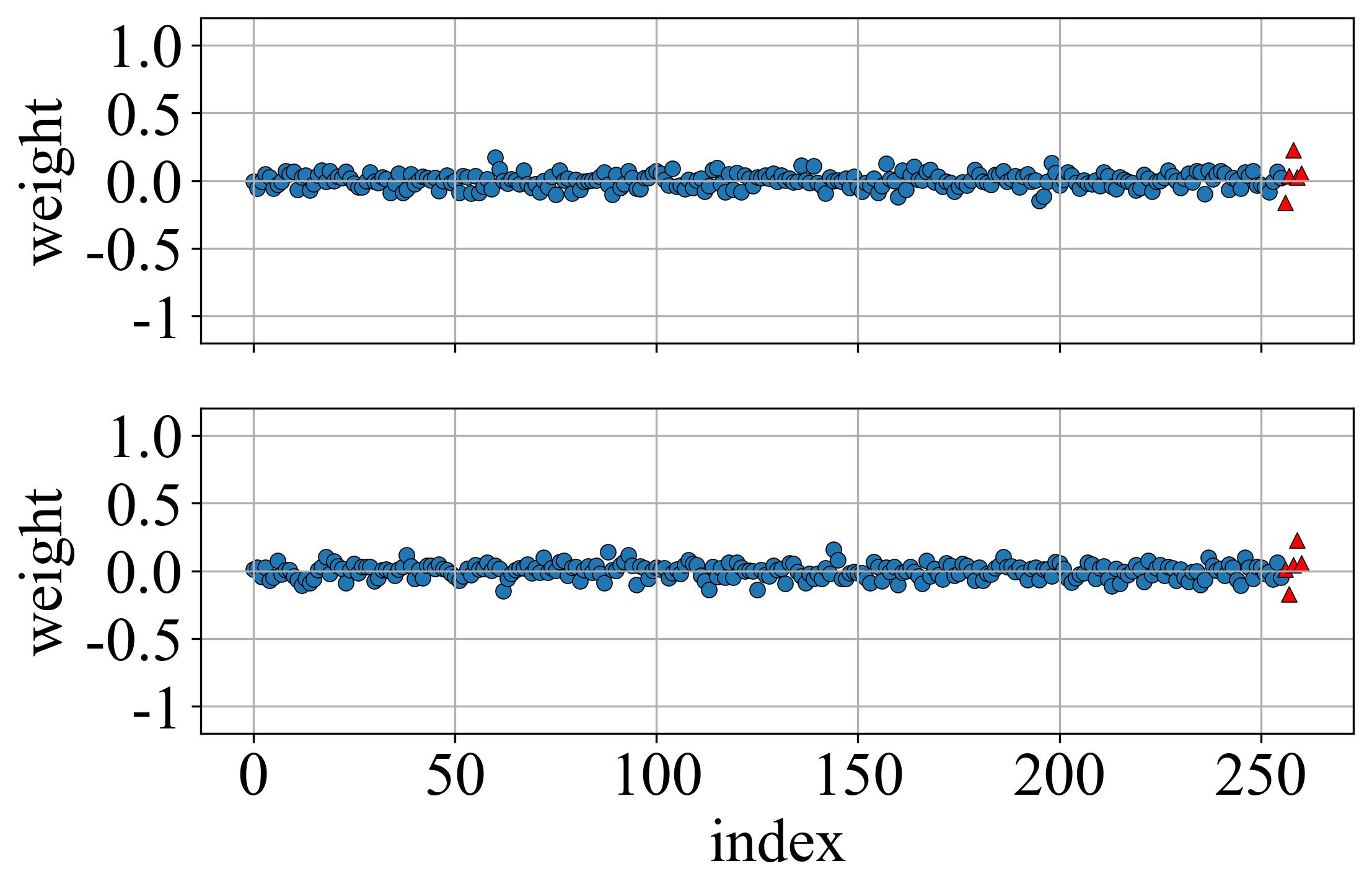}
    \subcaption{$g = 2.2$}
  \end{minipage}
  \begin{minipage}[t]{0.49\linewidth}
    \centering
    \includegraphics[bb=0 0 600 300, scale=0.35]{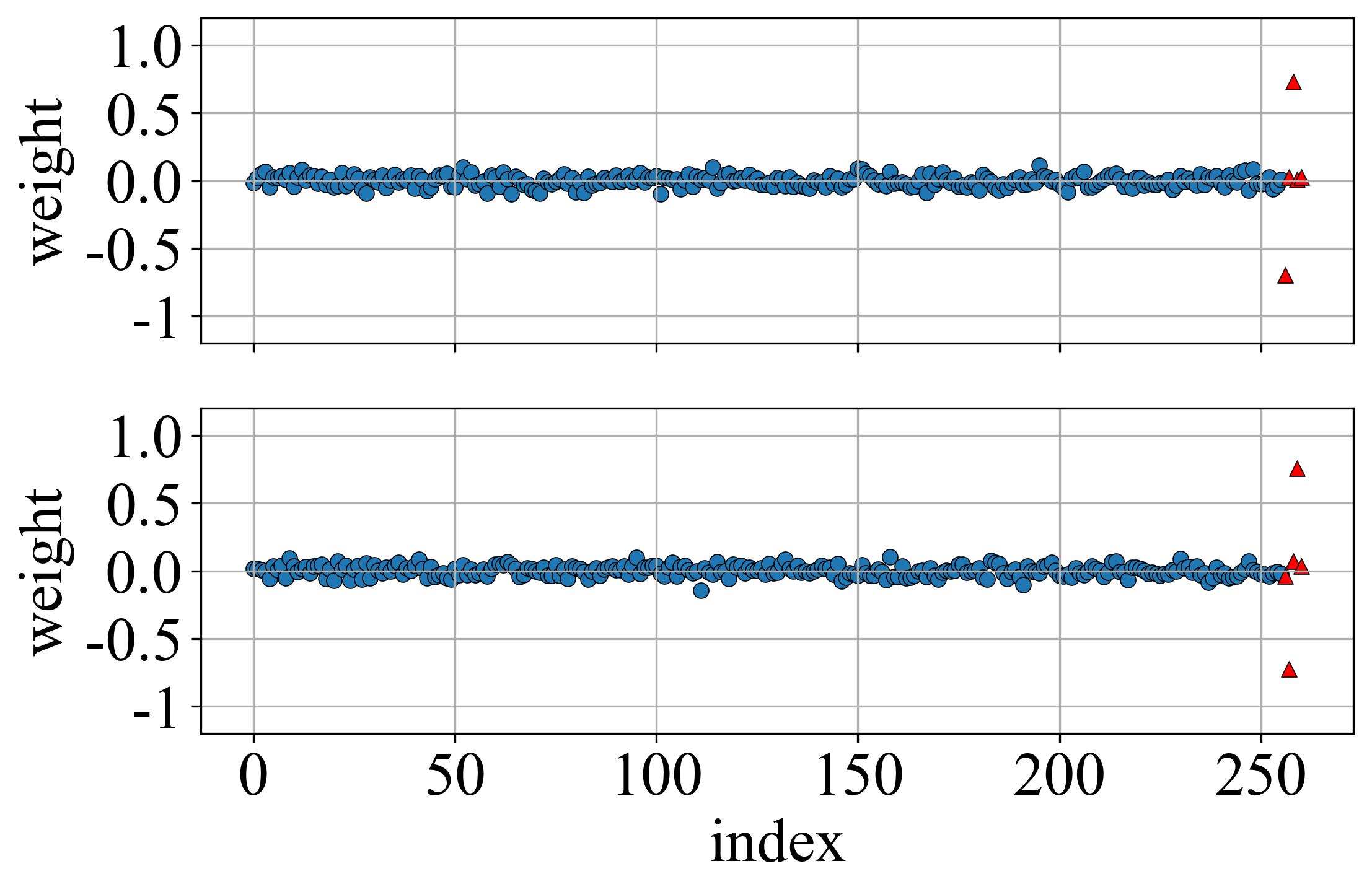}
    \subcaption{$g = 5$}
  \end{minipage}
  \caption{Readout weights. Blue dots indicate $256$ weights from the reservoir to the readout. The red triangles show the bypass weights through which the state of the environment is given directly to the readout. (a) shows the result when $g=2.2$. (b) shows the result when $g=5$. The upper and lower figures show the weights given to the action output decision unit in the $x$-axis and $y$-axis directions, respectively.
  }
\label{fig:compare_weights}
\end{figure}

\begin{figure}[t]
  \begin{minipage}[t]{0.49\linewidth}
    \centering
    \includegraphics[bb=0 0 600 300, scale=0.32]{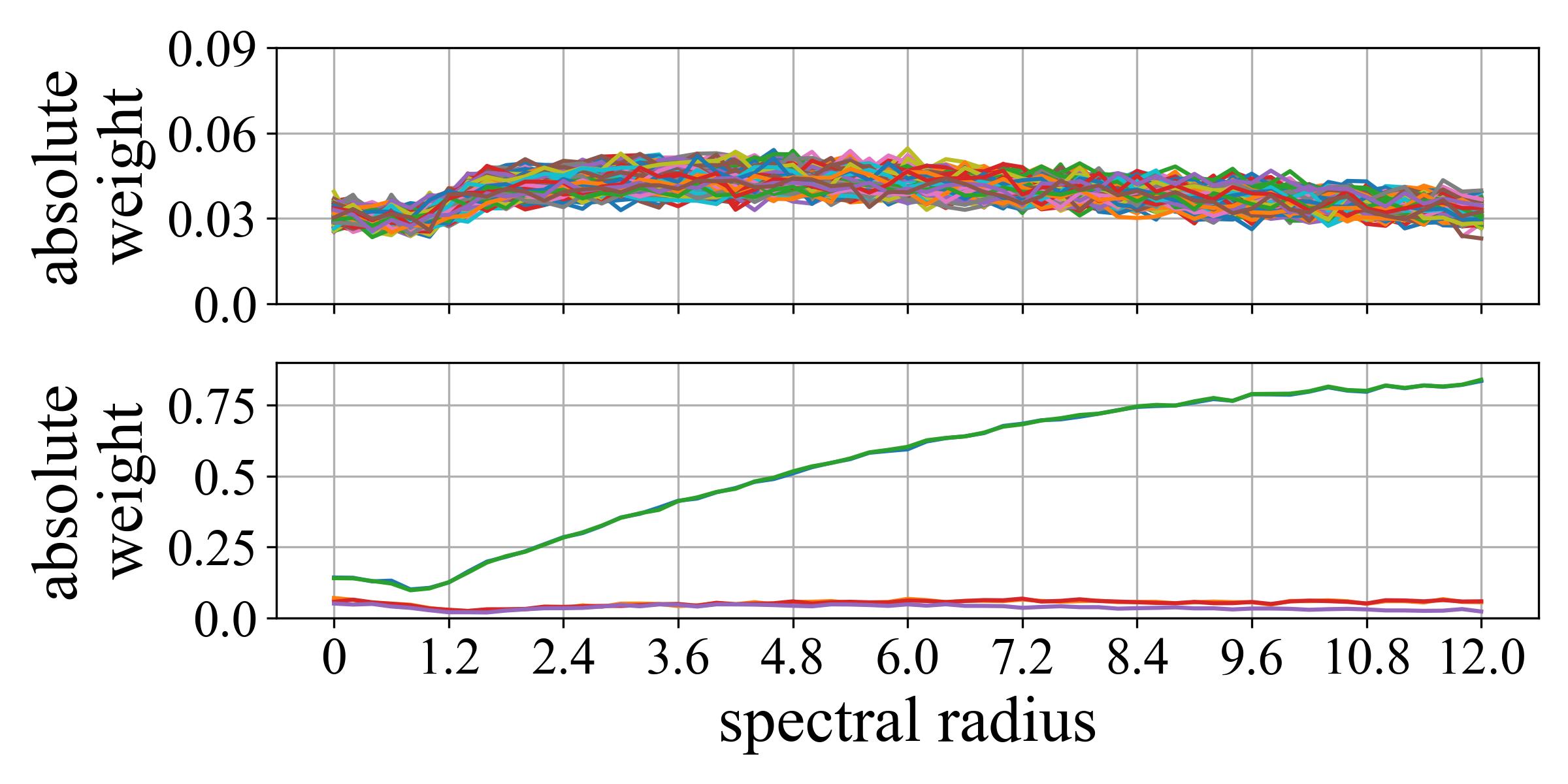}
    \subcaption{$x$-axis direction unit}
  \end{minipage}
  \begin{minipage}[t]{0.49\linewidth}
    \centering
    \includegraphics[bb=0 0 600 300, scale=0.32]{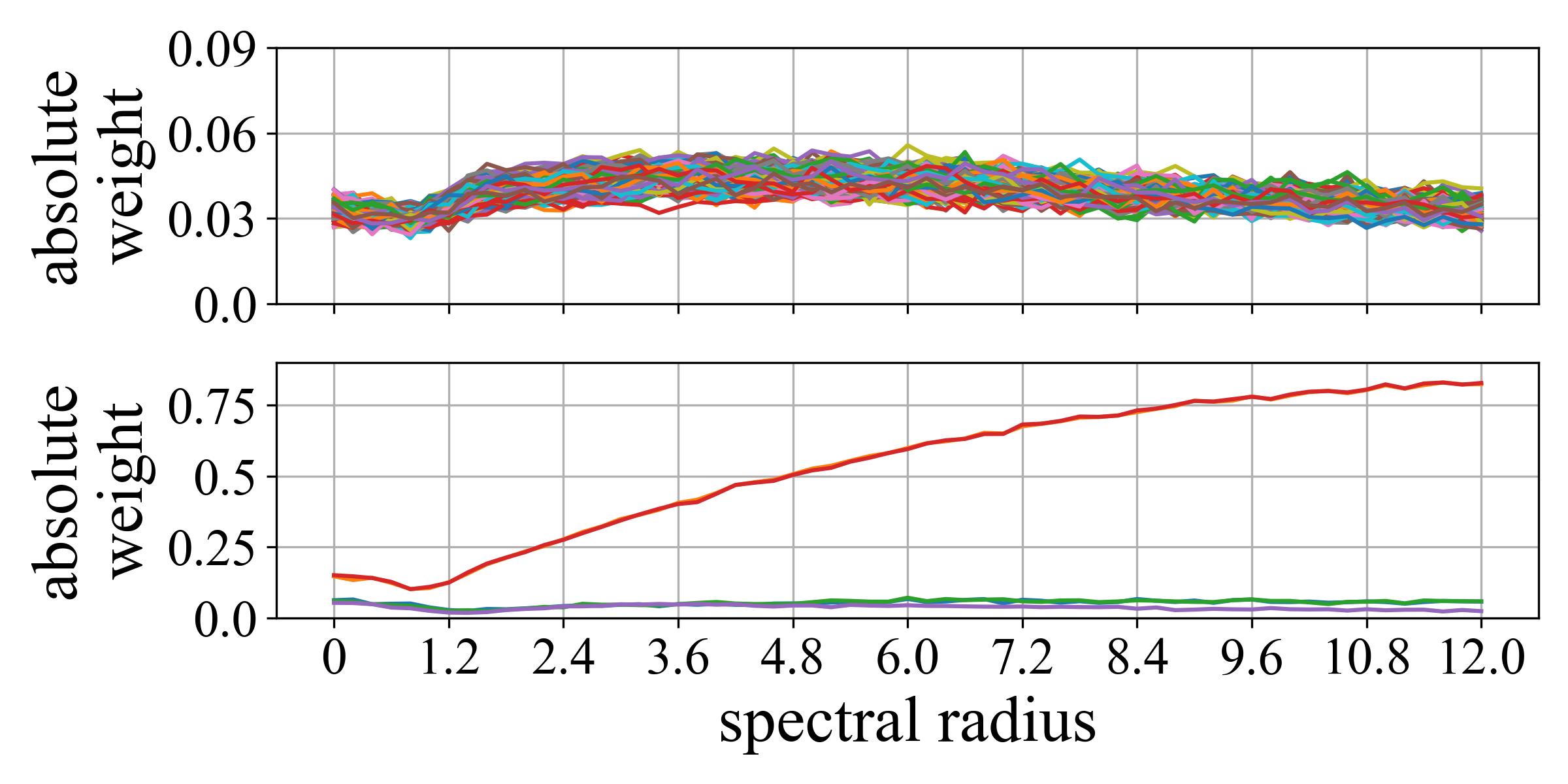}
    \subcaption{$y$-axis direction unit}
  \end{minipage}
  \caption{Spectral radius and readout weights. The vertical axis shows the absolute average of the learned weights across 100 different random seeds. The horizontal axis shows the spectral radius. The upper graph shows the $256$ weights from the reservoir to the readout, and the lower graph shows the $5$ bypass weights. (a) and (b) show the weights of the readout units that determine the travel distance $a^x, a^y$ in the $x$ and $y$ axes, respectively.
}
\label{fig:LC_sr_weights}
\end{figure}

We investigated how the acquired readout weights vary with the spectral radius $g$, which adjusts the chaoticity of the reservoir.
The inputs for the readout are the reservoir state and the agent's state in the environment. They are received through 256 and 5 weights, respectively.
Figure \ref{fig:compare_weights} shows the readout weights after training for spectral radius $g$ of $2.2$ and $5$. The blue dots show the weights from the reservoir, and the red dots show the bypass weights.
Comparing these figures, we can see that the bypass weights are larger for the case of $g = 5$ than for the case of $g = 2.2$, and the weights from the reservoir are relatively smaller than the bypass weights.
Figure \ref{fig:LC_sr_weights} shows how the average of the absolute values of the weights from the reservoir and the bypass weights change when the spectral radius is varied.
This figure shows that the bypass weights increase as the spectral radius increases.
This result suggests that as $g$ becomes excessively large, the agent tends to ignore the reservoir states and focus more on the inputs provided directly from the environment. This indicates that the states of a strongly chaotic reservoir lose their value as an input for the agent to accomplish the task.

We have validated the goal change task with $g=5$.
Figure \ref{fig:LC_goal_change}(c) shows the results of this validation.
Comparing this with the result under the setting of $g=2.2$ shown in Fig. \ref{fig:LC_goal_change}(a), re-learning convergence was slower when $g=5$.
Figure \ref{fig:POS_goal_change}(b) shows the agent's behavior in the environment at $g=5$.
Comparing Figs. \ref{fig:POS_goal_change}(a) and (b), we can see that in the case of $g=5$, the agent continues its exploratory behavior for more steps and takes more steps to shift to an exploitation mode.
This slow re-learning seems to be because the strongly chaotic reservoir states stored in the replay buffer before the goal change negatively affected learning. To verify this, we set the replay buffer size to $64$, the same as the batch size, and trained relying only on recent experiences. Figure \ref{fig:sr_score}(d) shows the learning performance for different spectral radiuses under this setting. This figure shows that the required steps for re-learning did not increase even with large $g$, although the learning performance tends to decrease. This result suggests that excessive spectral radiuses complicate the reservoir state and reduce its correlation with the input from the environment. This makes the experience stored in the replay buffer before the goal change worthless for learning, thereby increasing the number of steps required for re-learning. Furthermore, excessively large $g$ hinders the TD3-CBRL agent's ability to properly balance exploration and exploitation.

Excessive chaoticity in a reservoir can cause its states to diverge or fluctuate erratically. Such unstable dynamics poses a concern as they degrade the agent's learning performance, especially in tasks that unfold over long time horizons. Therefore, to investigate the impact of the spectral radius $g$ on learning performance in a more extensive task field that requires long time steps to reach the goal, we experimented with the goal task whose field was expanded from the range $[0, 20]$ to $[0,100]$ in the x-y plane. The result, shown in Fig. \ref{fig:sr_score}(e), demonstrates a marked degradation in performance at larger $g$ values. Consequently, the range for successful learning was restricted to an extremely narrow range centered around $g=1.8$. This result suggests that excessive chaoticity may hinder the formation of a global exploration in an extensive state space.

The sensitivity to initial conditions of the chaotic reservoir can cause the system to react sensitively to observation noise, potentially leading to performance instability. Therefore, to evaluate the robustness of TD3-CBRL against observation noise, we investigated its performance on the goal task when Gaussian noise $\mathcal{N}(0, 10^{-2})$ was added to the observation $\bm{u}$ in the test phase. The results are shown in Fig. \ref{fig:sr_score}(f). This figure shows that, overall, no significant performance degradation was observed compared to the case without noise (Fig. \ref{fig:sr_score}(a)). This result suggests that the TD3-CBRL agent possesses a certain degree of robustness against observation noise under the experimental conditions of a simple goal task.

\subsection{Exploration with random number layer}
\label{sec:exp_random_vector}
\begin{figure}[htbp]
  \begin{minipage}[t]{0.49\linewidth}
    \centering
    \includegraphics[bb=0 0 550 225, scale=0.35]{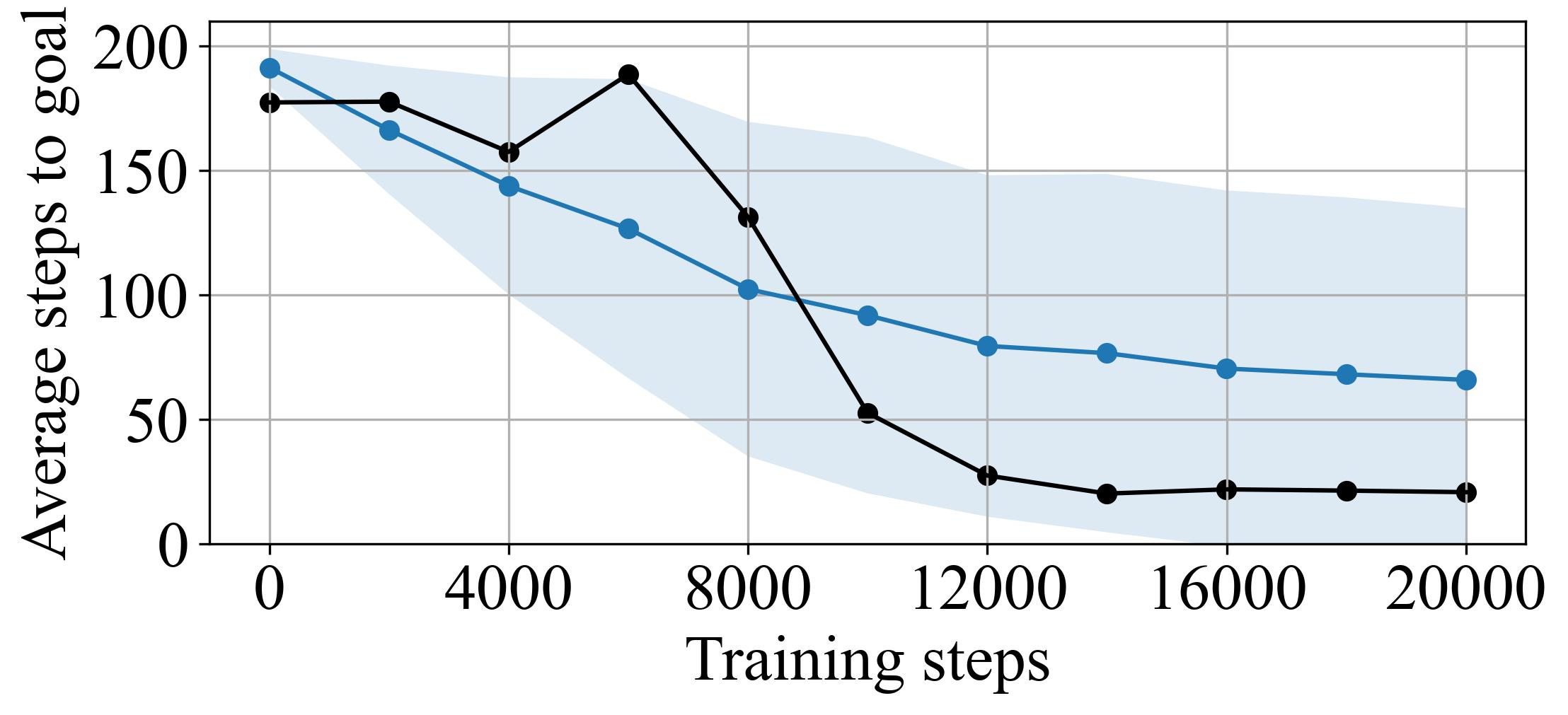}
    \subcaption{Learning curve.}
  \end{minipage}
  \begin{minipage}[t]{0.49\linewidth}
    \centering
    \includegraphics[bb=0 0 600 350, scale=0.30]{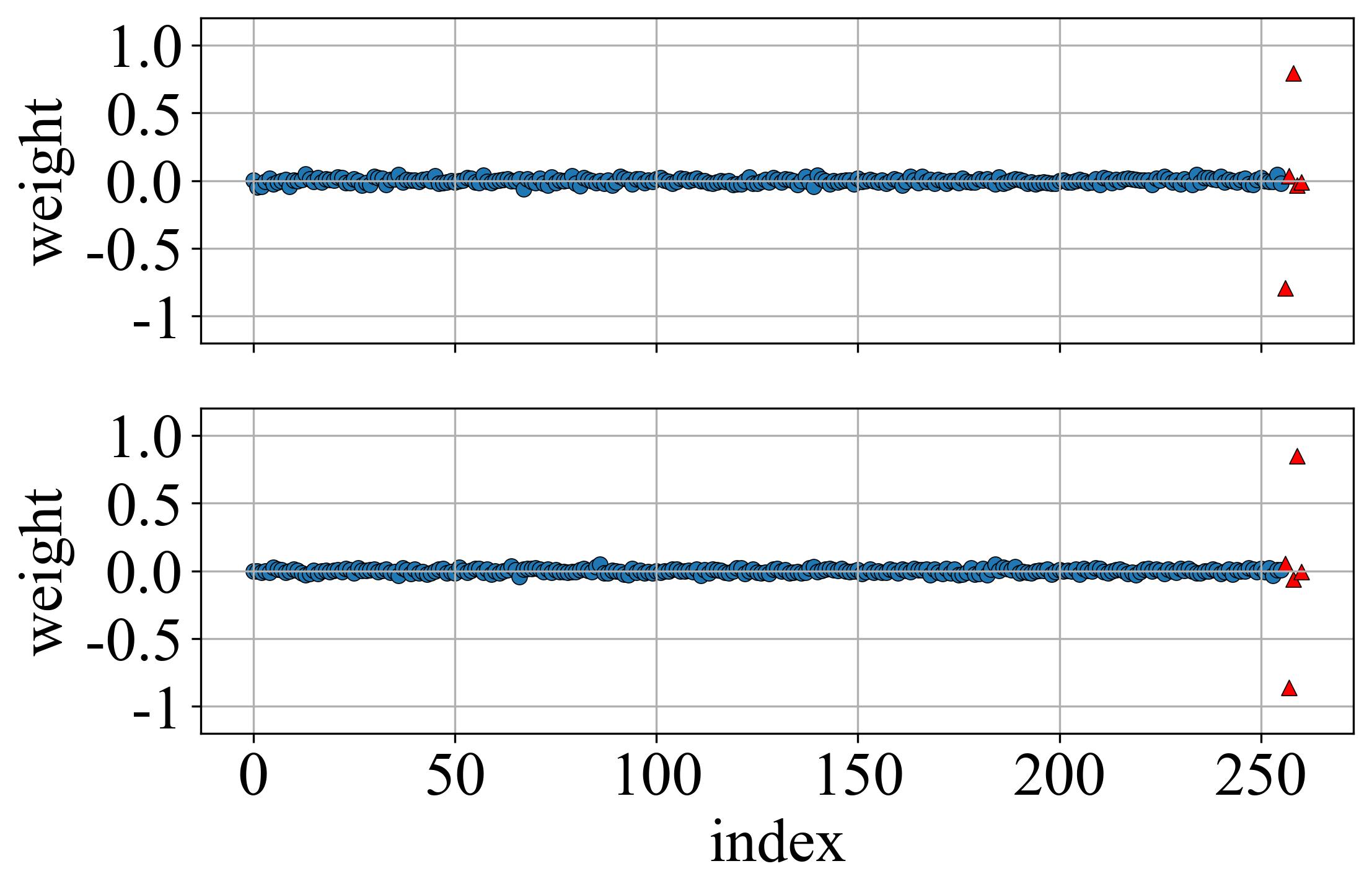}
    \subcaption{Readout weights.}
  \end{minipage}
\\ \\
  \begin{minipage}[t]{0.99\linewidth}
    \centering
    \includegraphics[bb=0 0 1500 300, scale=0.25]{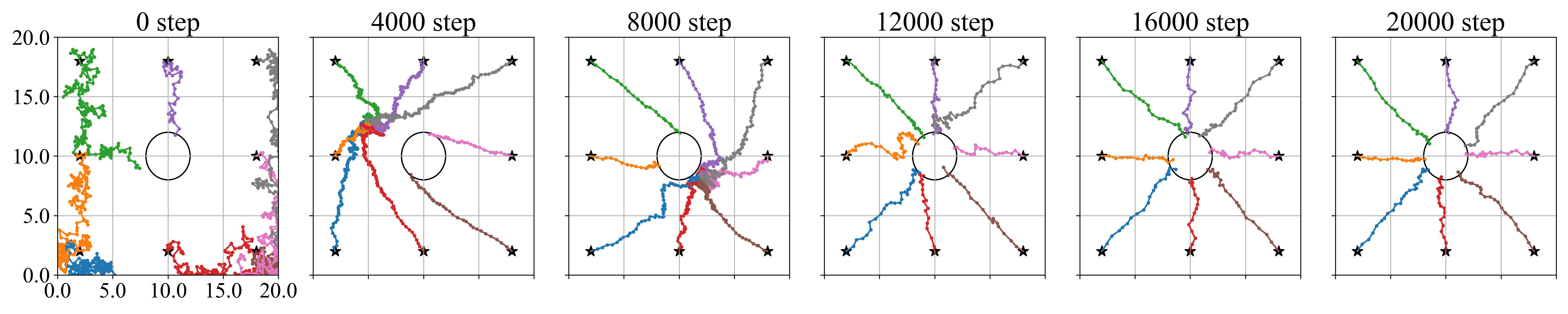}
    \subcaption{Agent trajectories.}
  \end{minipage}

  \caption{Result when a random vector is used instead of a reservoir. The definitions of line colors are the same as in Figs. \ref{fig:LC}, \ref{fig:POS}, and \ref{fig:compare_weights}.
}
\label{fig:Random_number_results}
\end{figure}

The above verifications confirmed that learning proceeds to ignore the reservoir's output and the experiences in the replay buffer stored before goal change negatively affect the re-learning ability if the reservoir is excessively chaotic.
To verify this result further, we investigated learning when a random vector of independent and identically distributed random numbers replaces the reservoir.
Specifically, we replaced the reservoir with a uniform random vector sampled from $[-1, 1]$ and conducted learning under this condition.
This setup is designed to reproduce the characteristics of a reservoir with extremely strong chaoticity.
In this setup, the random vector serves as the exploration component but is worthless as information about the agent's state.
Figure \ref{fig:Random_number_results} shows the results of this experiment. This figure shows that the agent succeeded in learning the task even when a random number vector replaces the reservoir, although the number of steps required for learning tends to increase.
The agent's trajectory in the environment is noisy due to the influence of a random vector.
The learning results for the readout weights under this situation show that the bypass weights are significantly larger than the weights from the random layer.
This result seems to be because the random layer's output is worthless as an input for accomplishing the task and suggests that a similar phenomenon occurs when the spectral radius of the reservoir is too high.

\begin{figure}[htbp]
  \begin{minipage}[t]{0.49\linewidth}
    \centering
    \includegraphics[bb=0 0 550 225, scale=0.35]{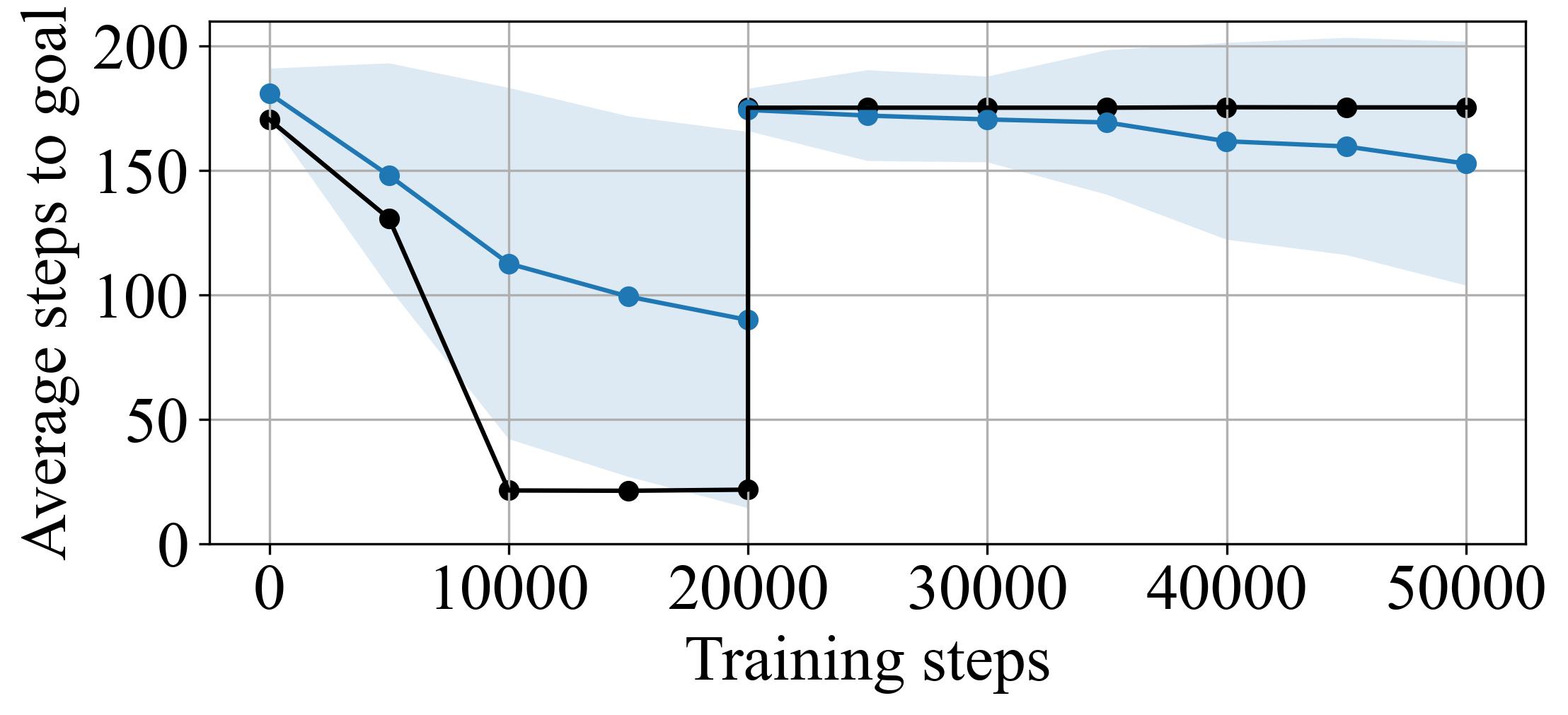}
    \subcaption{Learning curve.}
  \end{minipage}
  \begin{minipage}[t]{0.49\linewidth}
    \centering
    \includegraphics[bb=0 0 600 350, scale=0.30]{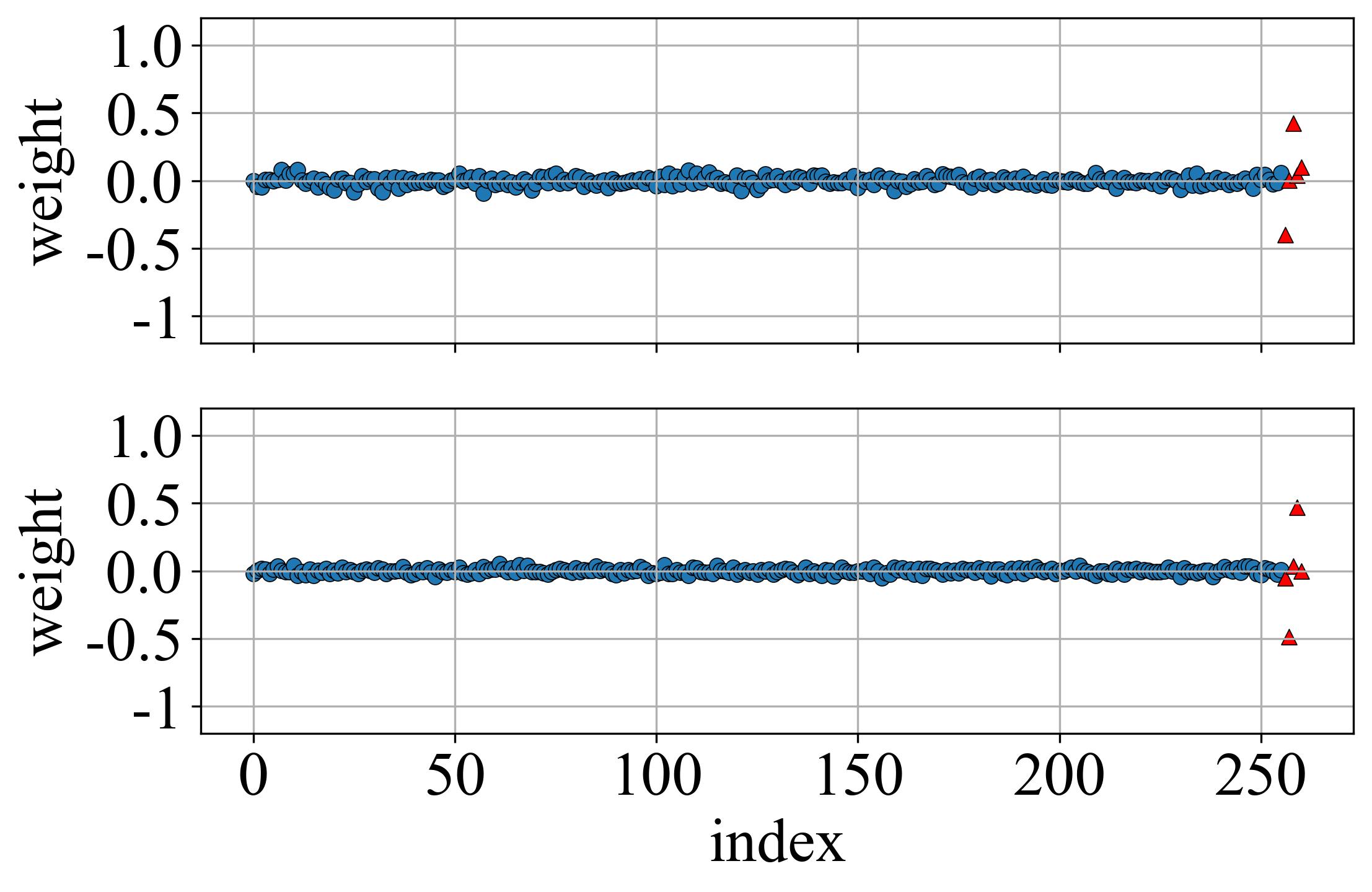}
    \subcaption{Readout weights.}
  \end{minipage}
\\ \\
  \begin{minipage}[t]{0.99\linewidth}
    \centering
    \includegraphics[bb=0 0 1500 300, scale=0.25]{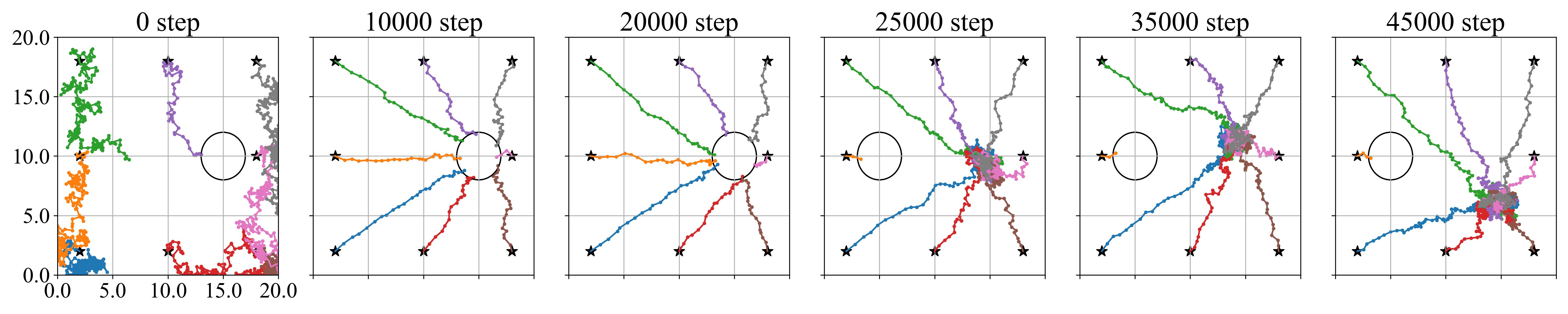}
    \subcaption{Agent trajectories.}
  \end{minipage}

  \caption{Result of the goal change task when a random vector is used instead of a reservoir. The definitions of the line colors are the same as in Figs. \ref{fig:LC}, \ref{fig:POS}, and \ref{fig:compare_weights}.
}
\label{fig:Random_number_results_goal_change}
\end{figure}

We experimented with a goal change task to verify whether the exploration with a random layer is flexible enough to adapt to environmental changes.
The task settings are the same as in Section \ref{sec:goal_change_task}.
Figure \ref{fig:Random_number_results_goal_change} shows the experimental results under these conditions.
This figure shows that the random layer model failed to learn the goal change task. This result indicates that the model with the random layer cannot flexibly switch between exploration and exploitation.

\begin{figure}[htbp]
  \begin{minipage}[t]{0.49\linewidth}
    \centering
    \includegraphics[bb=0 0 550 225, scale=0.35]{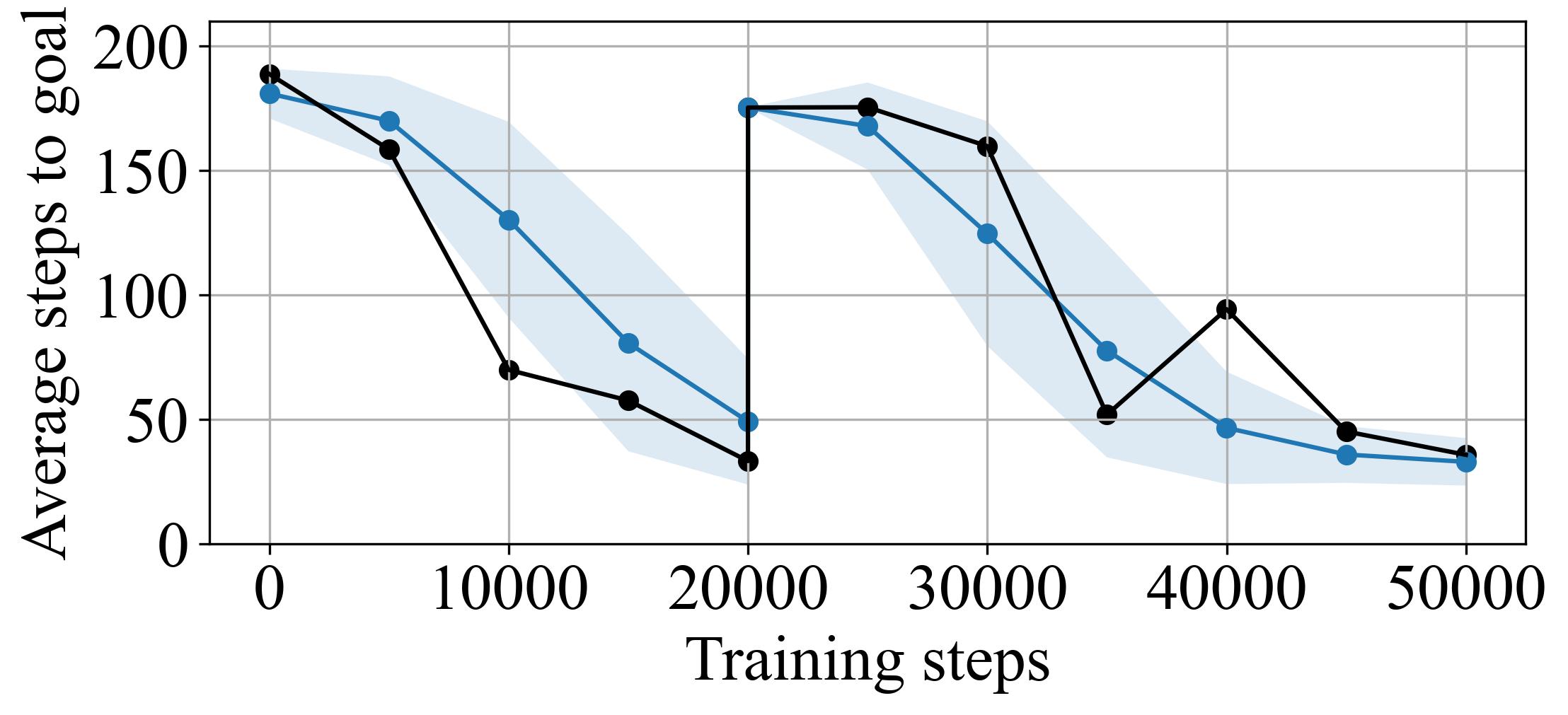}
    \subcaption{Learning curve.}
  \end{minipage}
  \begin{minipage}[t]{0.49\linewidth}
    \centering
    \includegraphics[bb=0 0 600 350, scale=0.30]{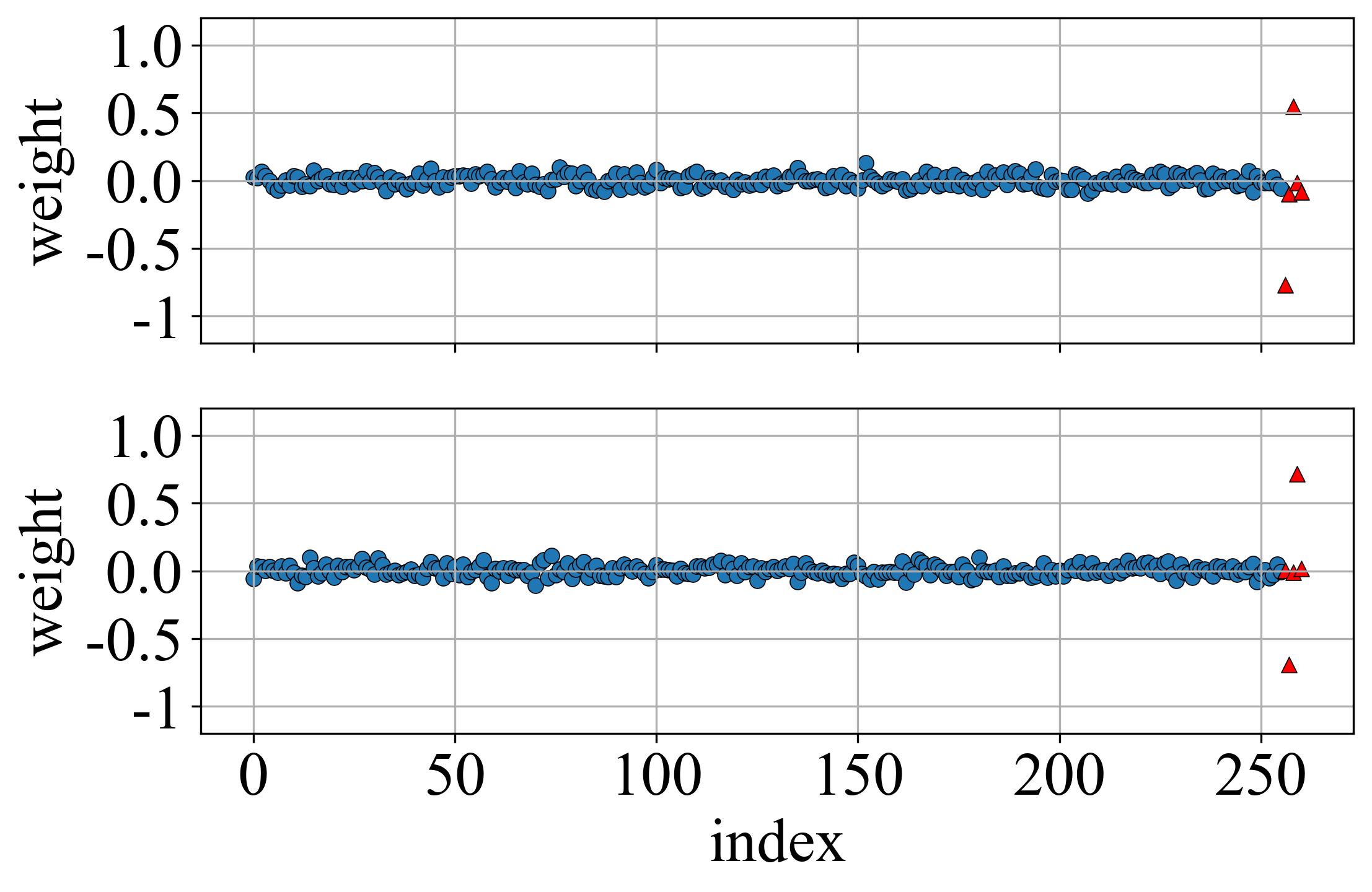}
    \subcaption{Readout weights.}
  \end{minipage}
\\ \\
  \begin{minipage}[t]{0.99\linewidth}
    \centering
    \includegraphics[bb=0 0 1500 300, scale=0.25]{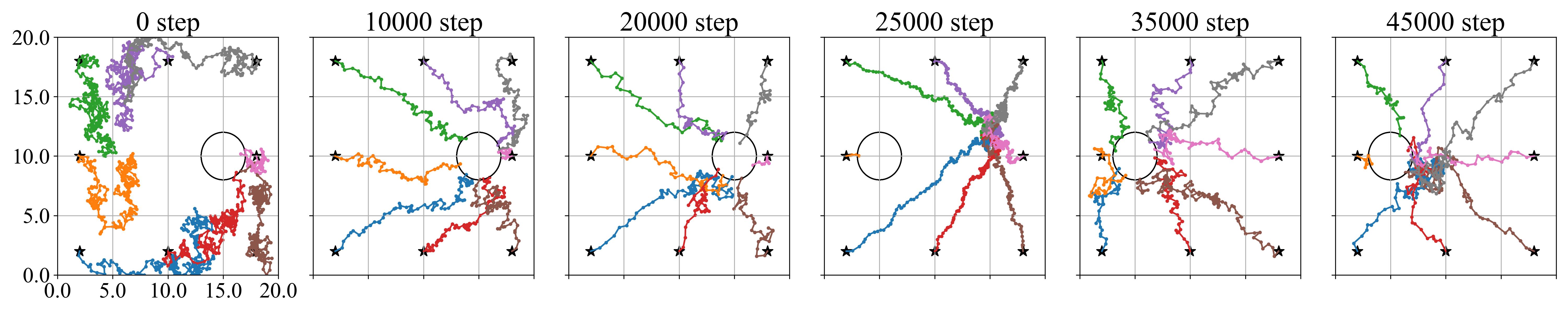}
    \subcaption{Agent trajectories.}
  \end{minipage}
  \caption{Result of the goal change task when a random vector is used instead of a reservoir and replay buffer size is set to 64. The definitions of the line colors are the same as in Figs. \ref{fig:LC}, \ref{fig:POS}, and \ref{fig:compare_weights}.
}
\label{fig:Random_number_results_goal_change_bufsize64}
\end{figure}

Figure \ref{fig:Random_number_results_goal_change_bufsize64} shows the results when the replay buffer size is set to 64. This figure indicates that while the agent's trajectory fluctuates significantly and the steps to the goal are numerous, it successfully learns to reach the new goal after the goal change. This result indicates that similar to the case with the reservoir, reducing the buffer size enables re-learning even when the model is highly irregular. These results suggest that storing meaningless vector information about the environment in the replay buffer hinders learning after environmental changes and slows down re-learning when $g$ of the reservoir is excessively large.

We observed that even when the reservoir exhibits strongly chaotic behavior, or the hidden layer is a random vector, the agent successfully re-learn if the replay buffer size is small and experiences before environmental changes are quickly removed. However, when the system is highly irregular, the trajectory toward the goal fluctuates significantly, negatively impacting performance. Additionally, experience replay is a crucial technique for improving sample efficiency, then it is undesirable to use a small buffer size. Therefore, appropriately setting the chaos level of the reservoir is critical to achieving both effective re-learning and improved learning performance in CBRL.

\subsection{reservoir dynamics}
The experimental results in Section \ref{sec:performance_chaoticity} and \ref{sec:exp_random_vector} suggest that the output of the reservoir, which is stored in the replay buffer before the goal change, affects the capability of re-learning. In other words, it is considered that some reservoir characteristics other than the role as a source of exploration components contribute effectively to re-learning.

\begin{figure}[t]
  \begin{minipage}[t]{0.49\linewidth}
    \centering
    \includegraphics[bb=0 0 600 300, scale=0.3]{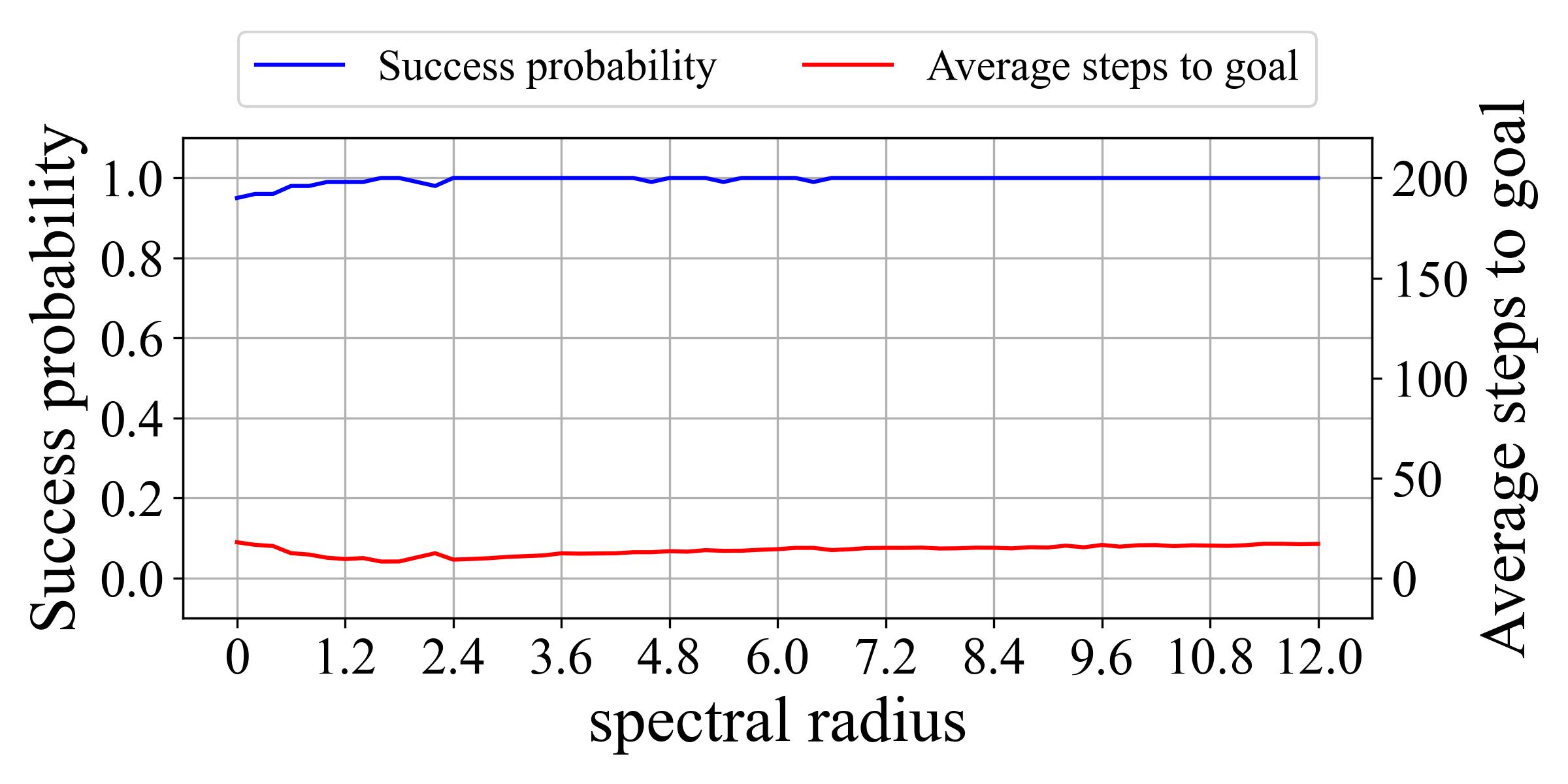}
    \subcaption{Goal task.}
  \end{minipage}
  \begin{minipage}[t]{0.49\linewidth}
    \centering
    \includegraphics[bb=0 0 600 300, scale=0.3]{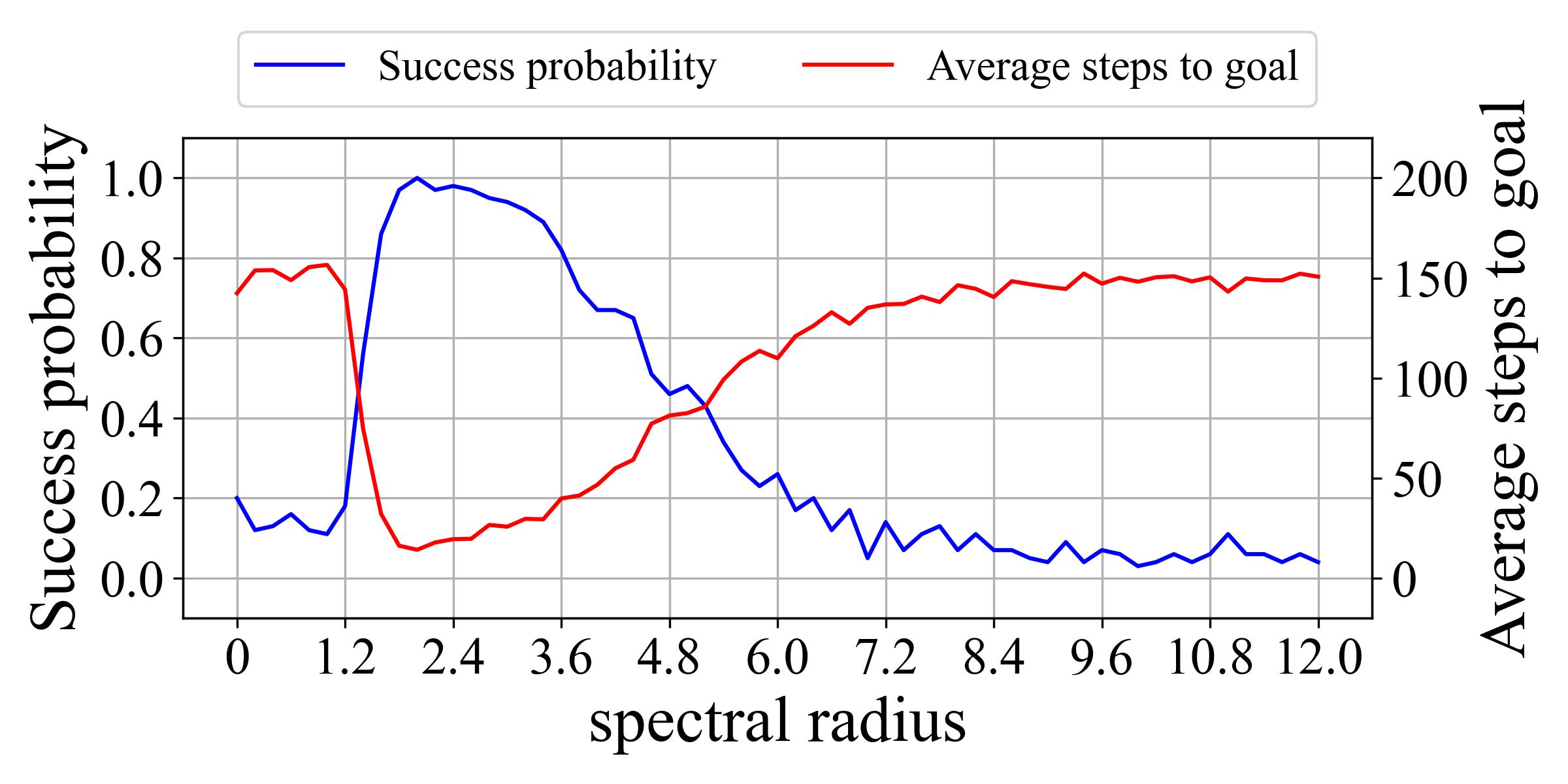}
    \subcaption{Goal change task.}
  \end{minipage}
  \caption{Learning performance of TD3-CBRL agent with the external random noise added to the action outputs, varying spectral radius. The definitions of the line colors are the same as in Fig. \ref{fig:sr_score}. (a) shows the learning results for the goal task. (b) shows the learning results of the goal change task. 
}
\label{fig:sr_score_ex_noise}
\end{figure}

To investigate the role of the reservoir other than as a source of exploration components, we trained a CBRL agent by adding external random noise to the action output and examined the relationship between the spectral radius and learning performance. Figure \ref{fig:sr_score_ex_noise}(a) shows the results of this experiment. The figure shows that even without chaotic properties in the reservoir, the agent succeeded in learning the task when external exploration noise was used.
Figure \ref{fig:sr_score_ex_noise}(b) shows the results of a similar verification for the goal change task. From the figure, we can confirm that there is an appropriate range for the spectral radius in this case. Interestingly, in cases where the task environment changes, learning performance decreases in the range where $g$ is small, even when external exploration noises are added to the actions. This result demonstrates that properly tuned chaotic reservoirs possess properties not found in random noise, which contribute to their effectiveness for re-learning.

\begin{figure}[t]
  \begin{minipage}[t]{0.99\linewidth}
    \centering
    \includegraphics[bb=0 0 1500 600, scale=0.25]{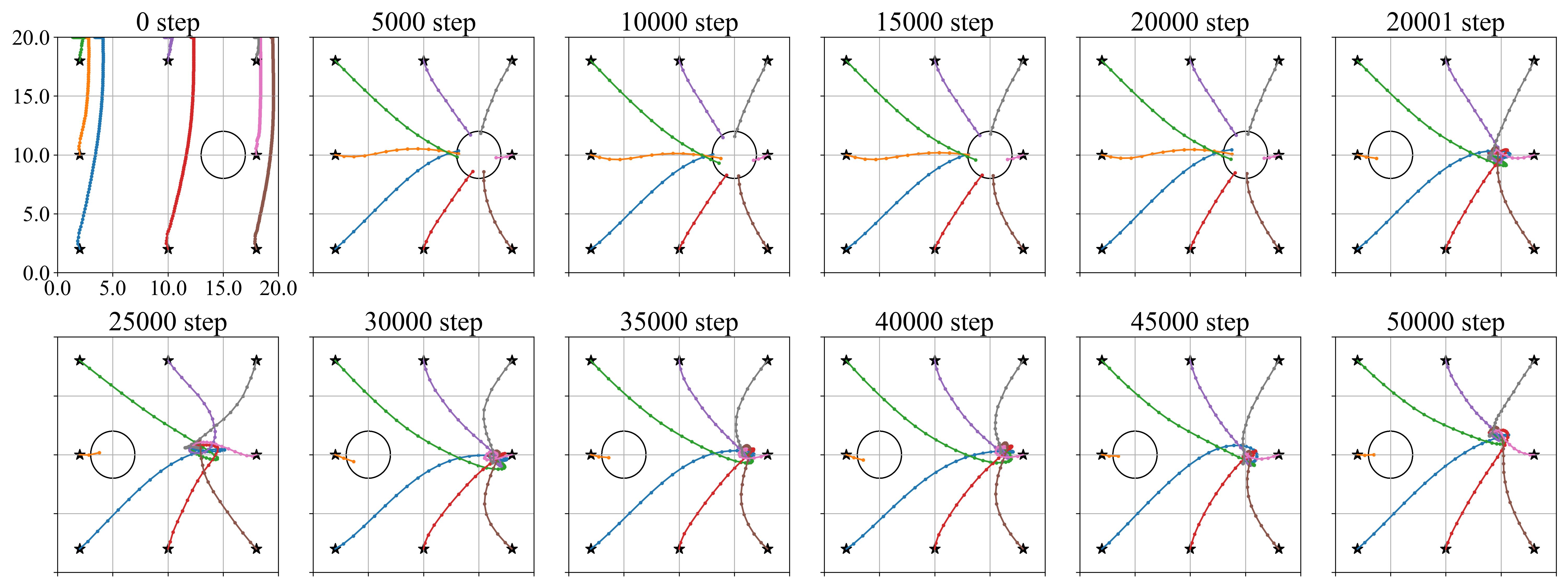}
    \subcaption{TD3-CBRL agent $(g=0.9)$ with external exploration random noise}
  \end{minipage}
\\
  \begin{minipage}[t]{0.99\linewidth}
    \centering
    \includegraphics[bb=0 0 1500 600, scale=0.25]{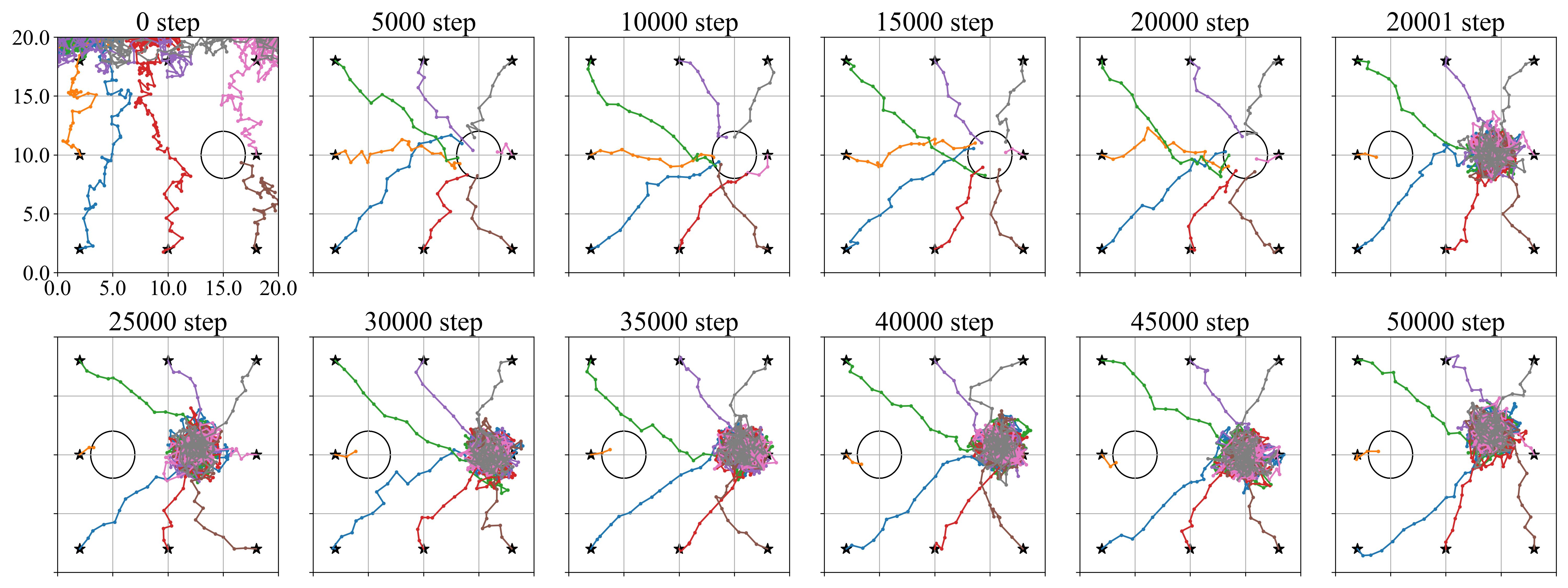}
    \subcaption{TD3-CBRL agent $(g=0.9)$ with external exploration random noise  (with the noise in the test phase)}
  \end{minipage}
  \caption{Agent trajectories during the test of the goal change task. (a) shows the result of the TD3-CBRL agent with spectral radius $g$ of 0.9. (b) shows the behavior of the same agent as in (a) when the exploration by random noises is not stopped during the test. Each graph in the subfigures shows the test results conducted every 5000 training steps. The definitions of the line colors are the same as in Fig. \ref{fig:POS}.
}
\label{fig:POS_goal_change_with_expl_noise}
\end{figure}

Figure \ref{fig:POS_goal_change_with_expl_noise}(a) shows the trajectories of an agent with $g=0.9$ during the test phase of the goal change task, while Figure \ref{fig:POS_goal_change_with_expl_noise}(b) shows the trajectories of the same agent when random noise is added to its actions during testing. From these figures, we observe that even when random noise is added to the action output, the agent continues to move towards the initial goal after the goal change, indicating that it has not learned to move towards the new goal.

\begin{figure}[t]
  \begin{minipage}[t]{0.99\linewidth}
    \centering
    \includegraphics[bb=0 0 1200 400, scale=0.25]{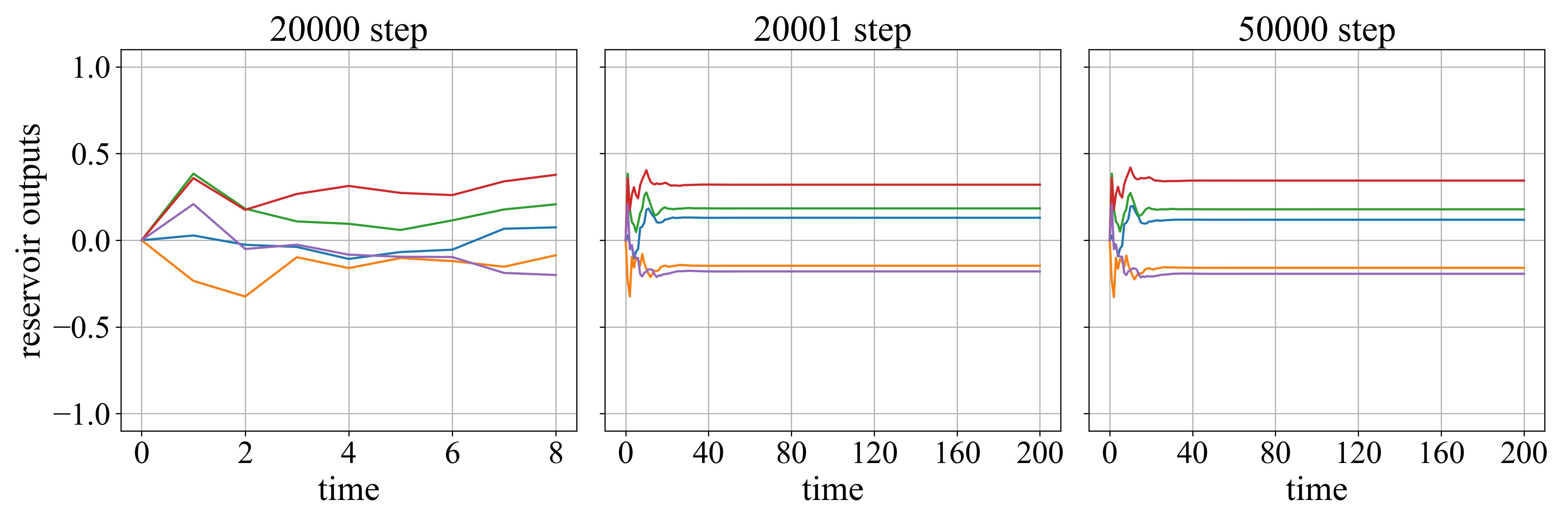}
    \subcaption{TD3-CBRL agent $(g=0.9)$ with external exploration random noise}
  \end{minipage}
  \\
  \begin{minipage}[t]{0.99\linewidth}
    \centering
    \includegraphics[bb=0 0 1200 400, scale=0.25]{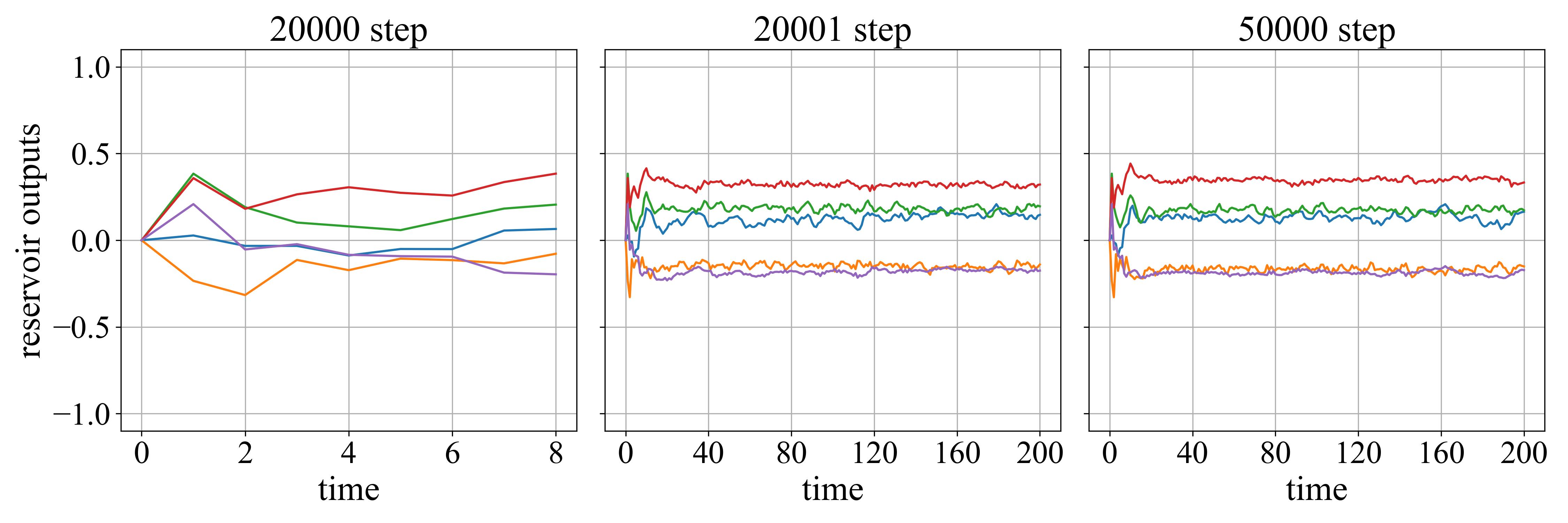}
    \subcaption{TD3-CBRL agent $(g=0.9)$ with external exploration noise even in the test phase}
  \end{minipage}
  \\
  \begin{minipage}[t]{0.99\linewidth}
    \centering
    \includegraphics[bb=0 0 1200 400, scale=0.25]{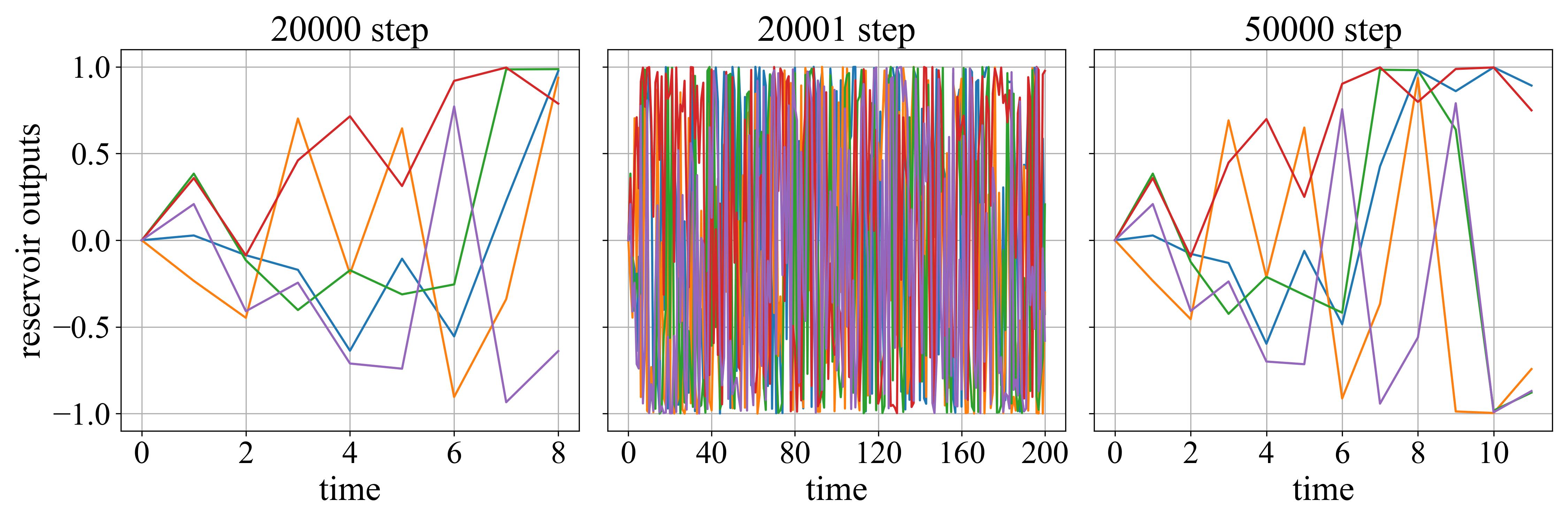}
    \subcaption{TD3-CBRL agent $(g=2.2)$ (The same agent as in Fig. \ref{fig:POS}(a))}
  \end{minipage}
  \caption{Reservoir states when the agent starts from $(10, 18)$ during testing.  (a) shows the result with $g=0.9$. (b) shows the result of the same agent as (a) with external random noise during testing. (c) shows the result with $g=2.2$ (the same condition as in Section \ref{sec:goal_change_task})
}
\label{fig:reservoir_output}
\end{figure}

Figure \ref{fig:reservoir_output} shows the outputs of 5 reservoir neurons when the agent starts from $(10, 18)$ during testing. 
From these figures, we can see that the reservoir state exhibits gradual dynamics when $g=0.9$. Notably, the output converges to a constant value after the initial transient and no longer changes. Even when random noise is added to the action output, we observe a similar trend with only slight perturbations in the output values.
On the other hand, it fluctuates wildly when $g=2.2$.

\begin{figure}[t]
  \begin{minipage}[t]{0.32\linewidth}
    \centering
    \includegraphics[bb=0 0 400 400, scale=0.42]{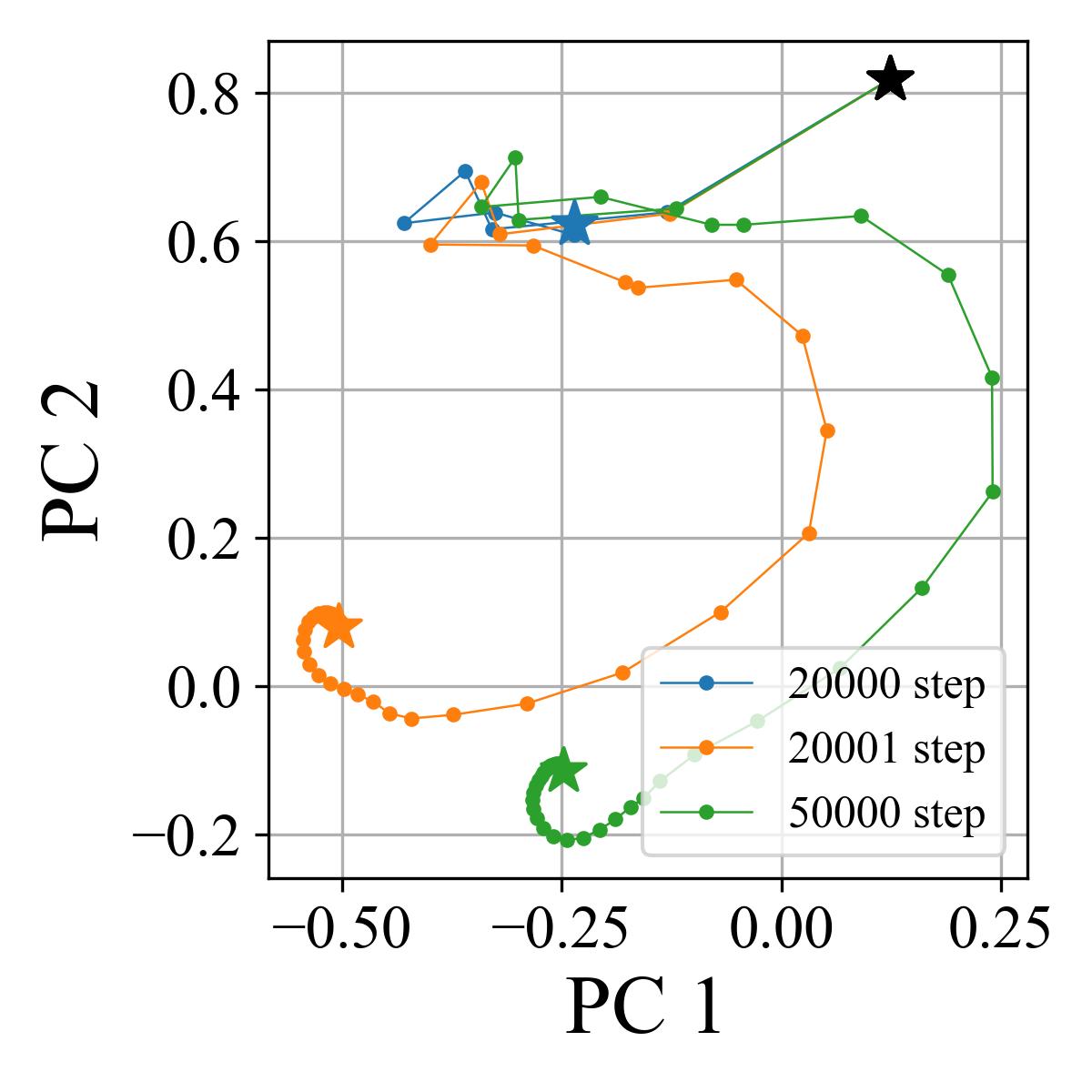}
    \subcaption{$g=0.9$}
  \end{minipage}
  \begin{minipage}[t]{0.32\linewidth}
    \centering
    \includegraphics[bb=0 0 400 400, scale=0.42]{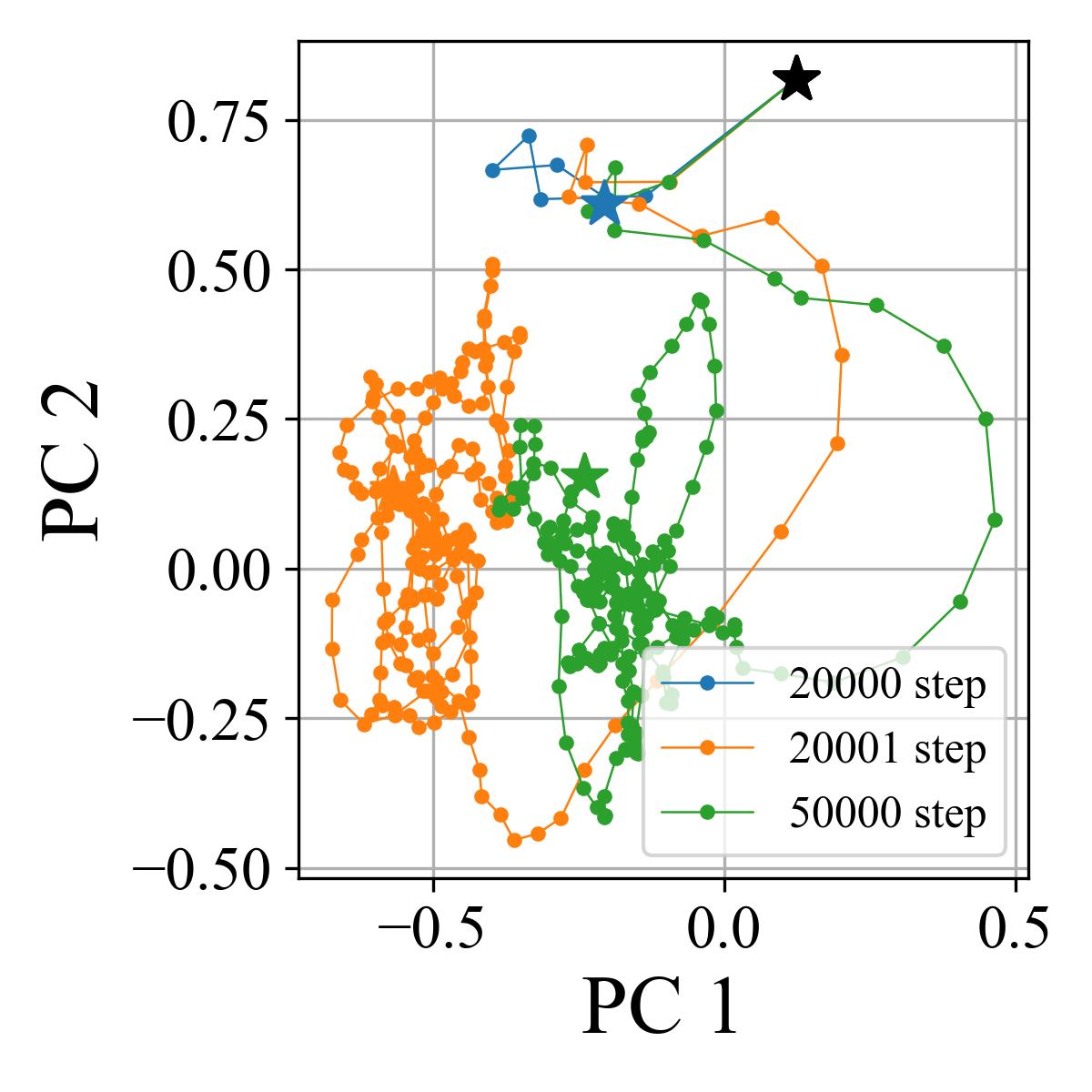}
    \subcaption{$g=0.9$ with external noise}
  \end{minipage}
  \begin{minipage}[t]{0.32\linewidth}
    \centering
    \includegraphics[bb=0 0 400 400, scale=0.42]{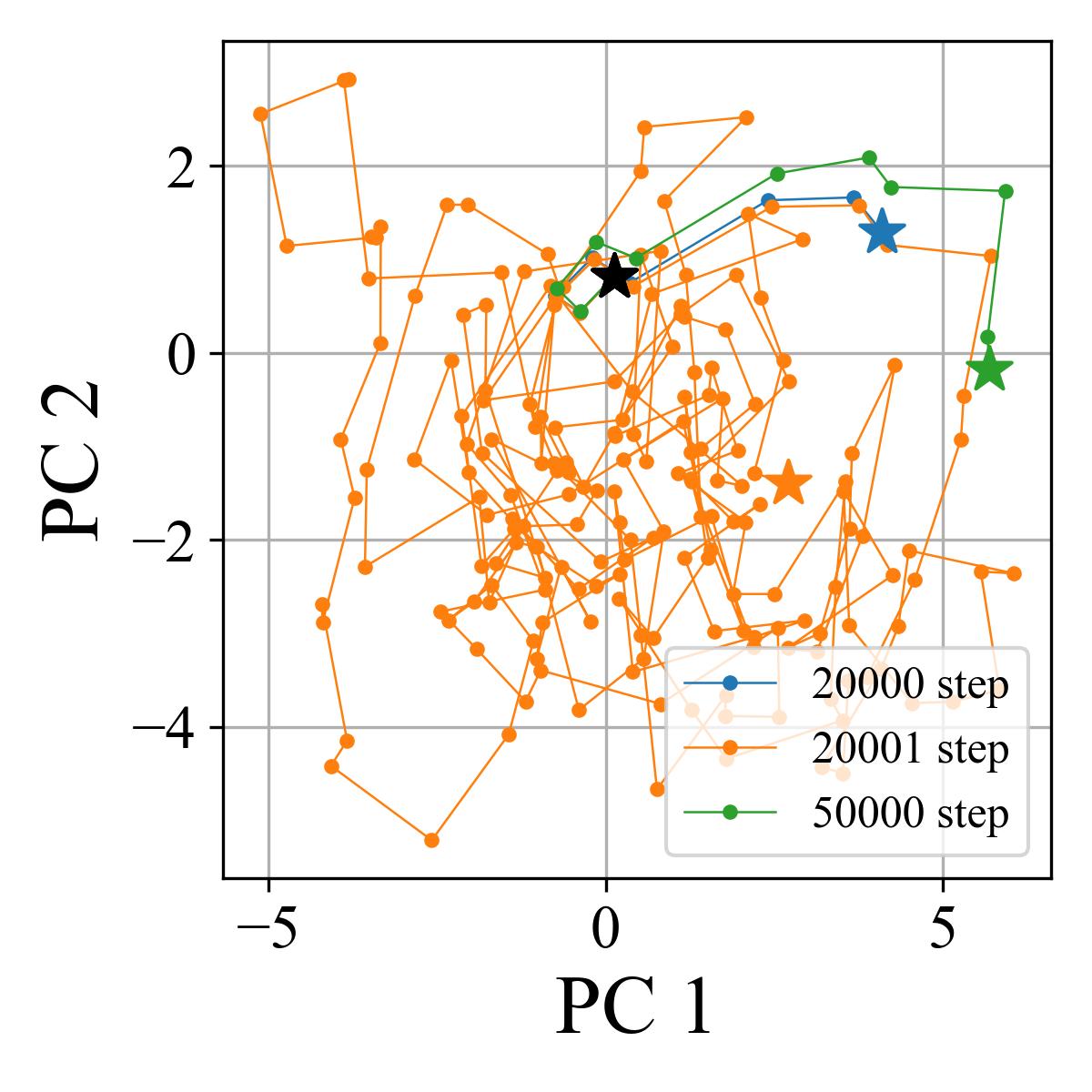}
    \subcaption{$g=2.2$}
  \end{minipage}
  \caption{PCA of reservoir states when the agent starts from $(10, 18)$ during the test. A black star indicates the initial reservoir state. Blue, orange, and green trajectories show the reservoir state transitions in the test phase at the $20000$ steps, $20001$ steps (immediately after the goal change), and $50000$ steps (final step). The colored stars indicate the reservoir state at the end of the episode when the agent reaches the goal or the 200-step elapses.
 (a) shows the result of the TD3-CBRL agent $(g=0.9)$ with external exploration random noise. (b) shows the result of TD3-CBRL agent $(g=0.9)$ with external exploration random noise even in the test phase.  (c) shows the result of TD3-CBRL agent $(g=2.2)$ (The same agent as in Fig. \ref{fig:POS}(a)). 
}
\label{fig:reservoir_output_pca}
\end{figure}

To analyze these dynamics, we recorded the reservoir states obtained at all test phases during these learning processes of the goal change task and performed principal component analysis (PCA). Figure \ref{fig:reservoir_output_pca} shows the results of the dimensionality reduction of the reservoir states. The data in this figure is based on the reservoir states when the agent starts at $(10, 18)$ and represents the test results after learning for $20000$ steps, $20001$ steps, and $50000$ steps.
From the orange trajectory, we can observe how the reservoir state evolved after reaching the initial goal. Figure \ref{fig:reservoir_output_pca}(a) shows that the agent with $g=0.9$ converges to a fixed-point after passing through the previous goal, and it can be seen that the trajectory converges to a slightly shifted point even after $50000$ steps. Figure \ref{fig:reservoir_output_pca}(b) shows that even when random noise is added to the action output, the reservoir state does not escape the constructed attractor and fluctuates near the same attractor. These results indicate that the behavior of reaching the first learned goal constitutes such an attractor, which exerts an attractive force that cannot be escaped by random noise. On the other hand, Figure \ref{fig:reservoir_output_pca}(c) shows that the reservoir's chaoticity causes it to transition to a trajectory with irregular states after reaching the endpoint of the previously learned trajectory. The result at the 50000 steps shows that it has successfully constructed a trajectory towards the new goal. This result suggests that the dynamics of spontaneous activity due to chaotic state transitions, rather than convergence due to learned behavior, play an important role in autonomously shifting to new exploration.
It is also noteworthy that the trajectories in Fig. \ref{fig:reservoir_output_pca} are close to each other in the initial steps. This suggests that the reservoir's dynamics retain memory of the sequence of states observed from the environment.

\begin{figure}[t]
    \centering
    \includegraphics[bb=0 0 400 400, scale=0.42]{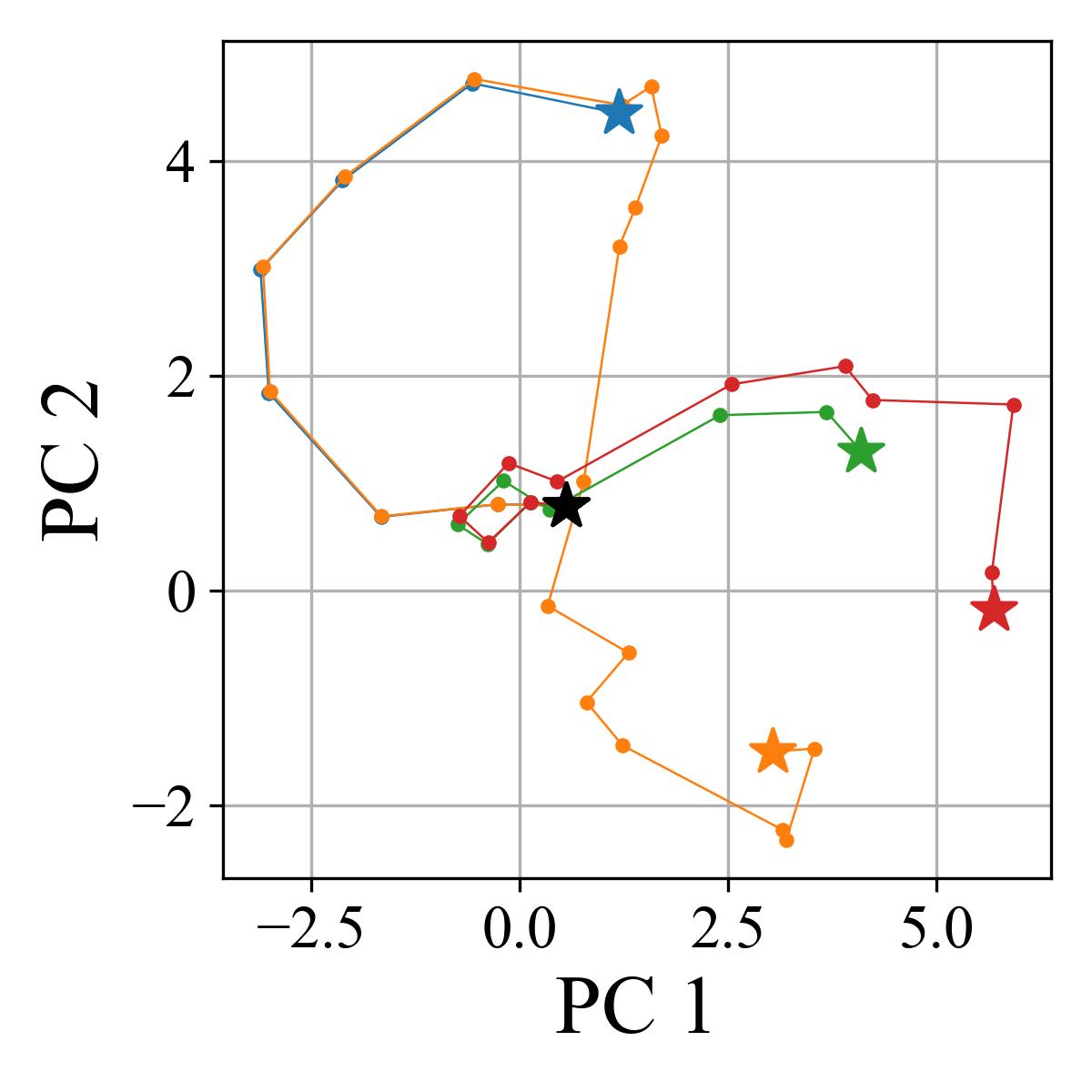}
  \caption{PCA of reservoir states of TD3-CBRL agent $(g=2.2)$ (The same agent as in Fig. \ref{fig:POS}(a)) during the test. A black star indicates the initial reservoir state. Blue and orange trajectories show the results at the $20000$ steps and $50000$ steps starting from $(10, 2)$.  Green and red trajectories show the results at the $20000$ steps and $50000$ steps starting from $(10, 18)$.
}
\label{fig:rsv_output_pca_each_start}
\end{figure}

Figure \ref{fig:rsv_output_pca_each_start} presents the PCA results of the reservoir state for a TD3-CBRL agent with a reservoir of $g=2.2$, starting from two initial positions $(10, 2)$ and $(10, 18)$, during tests at $20000$ and $50000$ steps. This figure shows that trajectories from different start points follow distinct paths, while trajectories starting from the same point are close to each other in the initial steps. This indicates that the reservoir retains the memory of the state sequence from the starting position. Furthermore, the reservoir state trajectory at the $50000$ steps transits from a point close to the point when the agent reached the initial goal before the goal change to a new trajectory that heads towards the new goal while maintaining temporal continuity. This suggests that the reservoir's short-term memory ability ensures the continuity of experiences before and after the environmental change stored in the replay buffer, which in turn positively affects the re-learning performance.
Such spatio-temporally correlated state trajectories do not appear when using random number vectors. It can be considered that the absence of the correlation causes the difference in performance between the chaotic reservoir and the random vector in the re-learning task.

\subsection{Learning performance and limitation}
Although there are some variations in the definitions of actions and states, previous CBRL research has typically employed simple goal tasks similar to the one used in this paper. This is because previous CBRL algorithms had limitations in learning more challenging tasks. To investigate the learning performance of a CBRL agent with TD3 on relatively more difficult tasks, we conducted experiments using MuJoCo continuous control tasks \cite{todorov2012mujoco} through OpenAI Gym \cite{brockman2016openaigym}, which are widely used as benchmarks in reinforcement learning.
In this experiment, the critic network and actor network (the readout of the chaotic reservoir) were implemented as MLPs with two hidden layers, each comprising 128 ReLU activation nodes. The output layer's neuron of the critic network was a linear activation node, while the neurons of the actor network were $\tanh$ nodes. The learning rate for both networks was set to $0.001$. A discount factor $\gamma$ was set to $0.99$, and the target network update time constant $\tau$ was set to $0.005$.

\begin{figure}[t]
  \begin{minipage}[t]{0.49\linewidth}
    \centering
    \includegraphics[bb=0 0 600 300, scale=0.3]{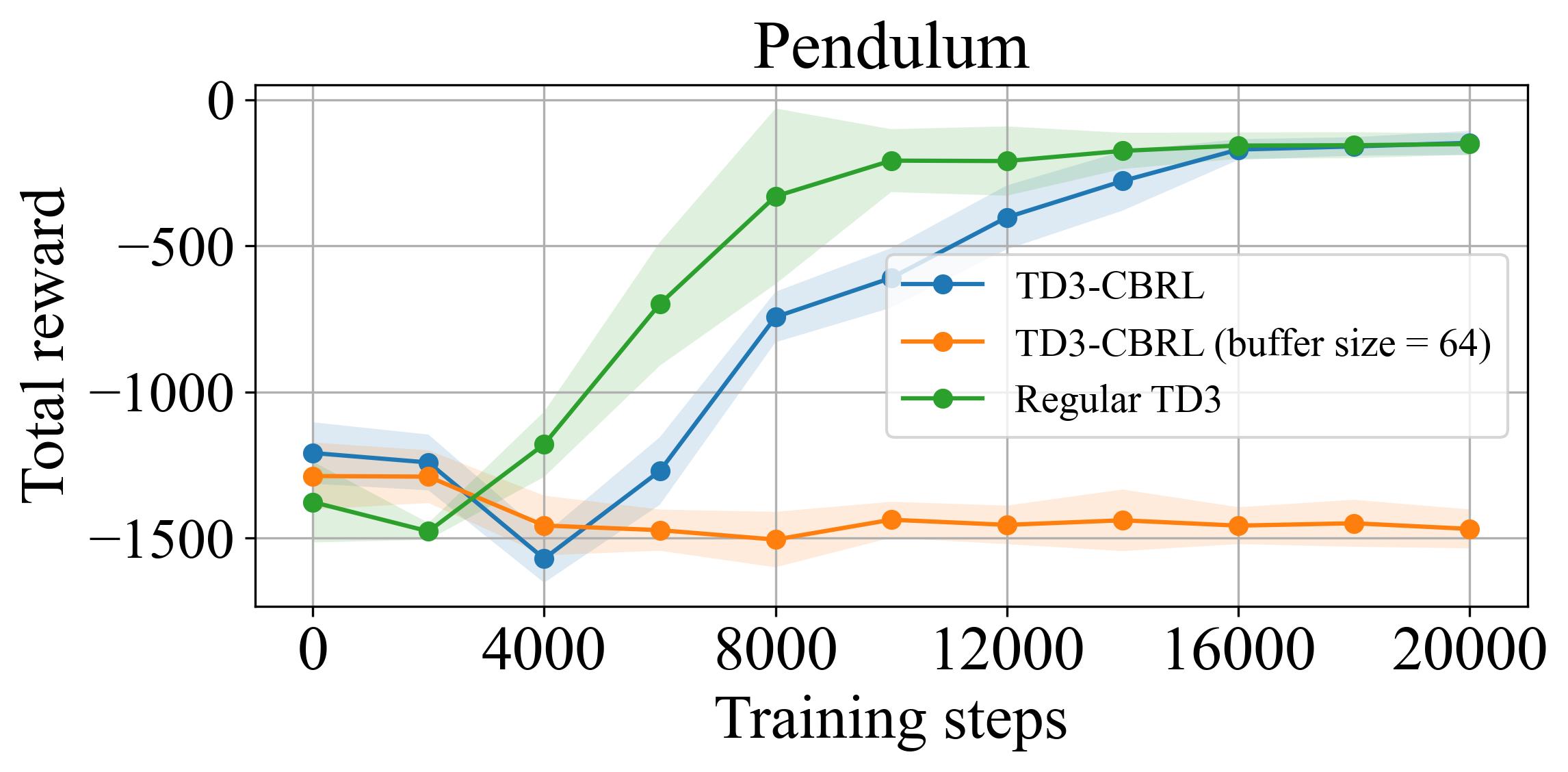}
    \subcaption{Pendulum}
  \end{minipage}
  \begin{minipage}[t]{0.49\linewidth}
    \centering
    \includegraphics[bb=0 0 600 300, scale=0.3]{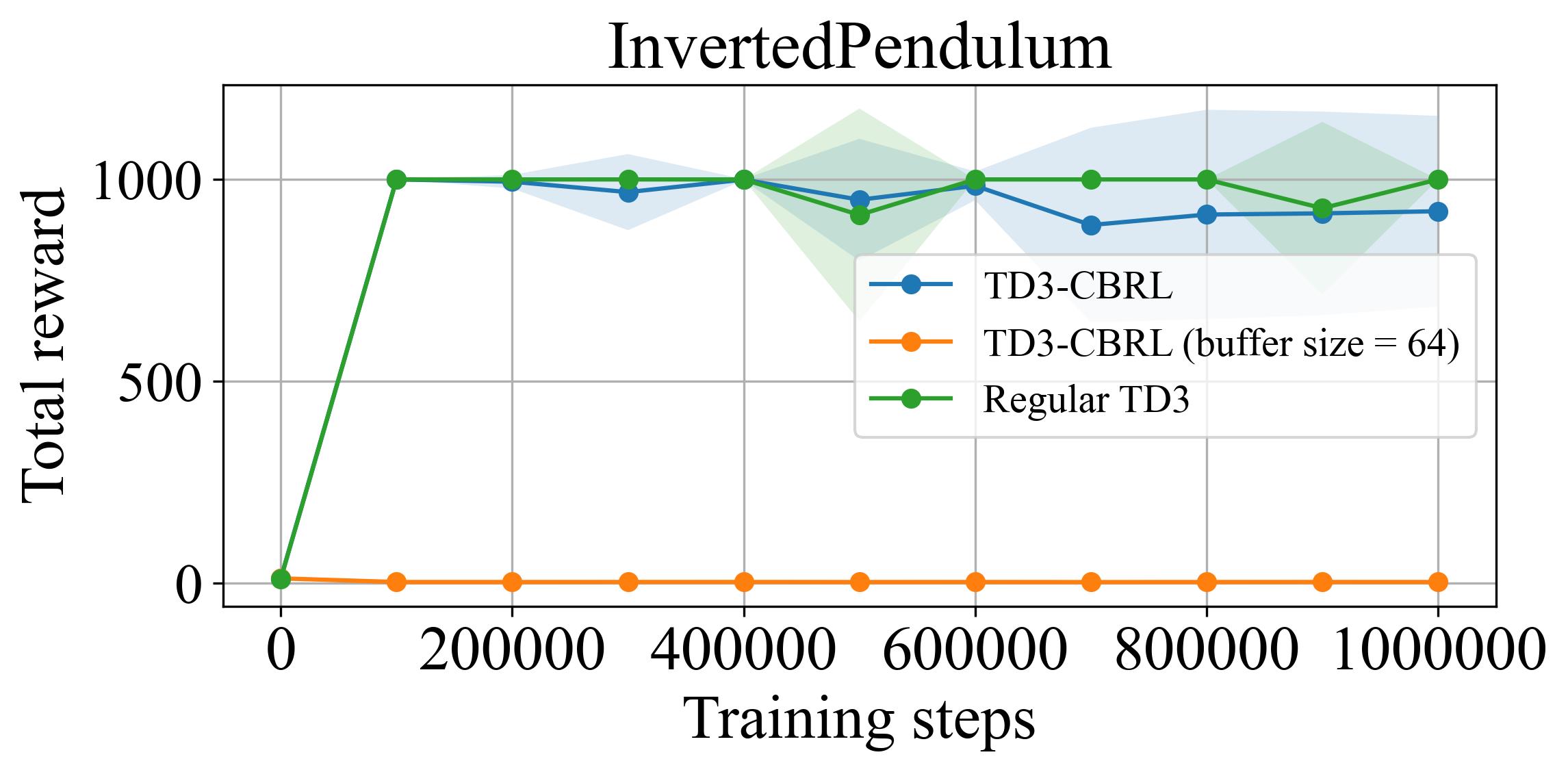}
    \subcaption{Inverted Pendulum}
  \end{minipage}
\\ \\ 
  \begin{minipage}[t]{0.49\linewidth}
    \centering
    \includegraphics[bb=0 0 600 300, scale=0.3]{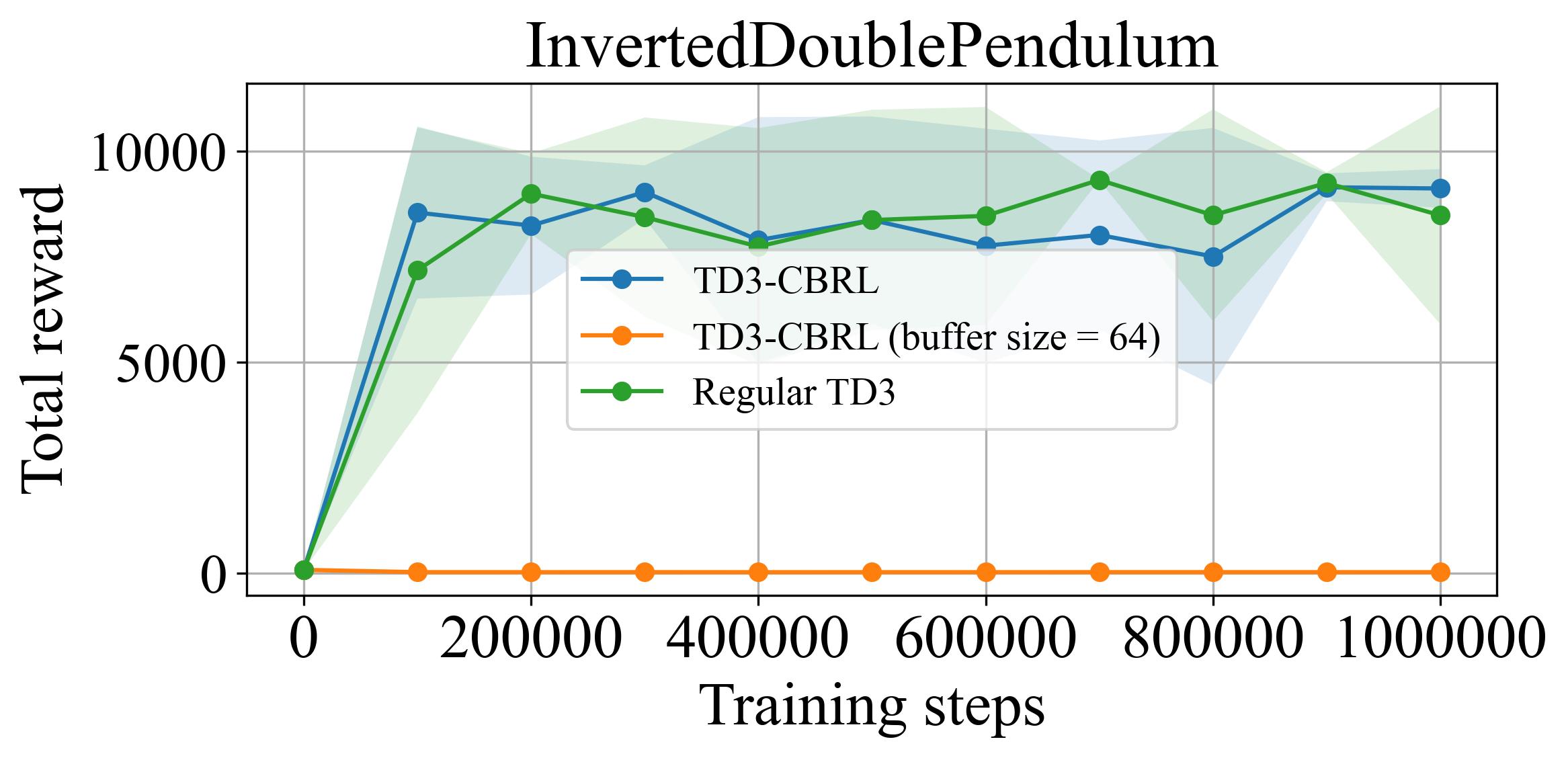}
    \subcaption{Inverted Double Pendulum}
  \end{minipage}
  \begin{minipage}[t]{0.49\linewidth}
    \centering
    \includegraphics[bb=0 0 600 300, scale=0.3]{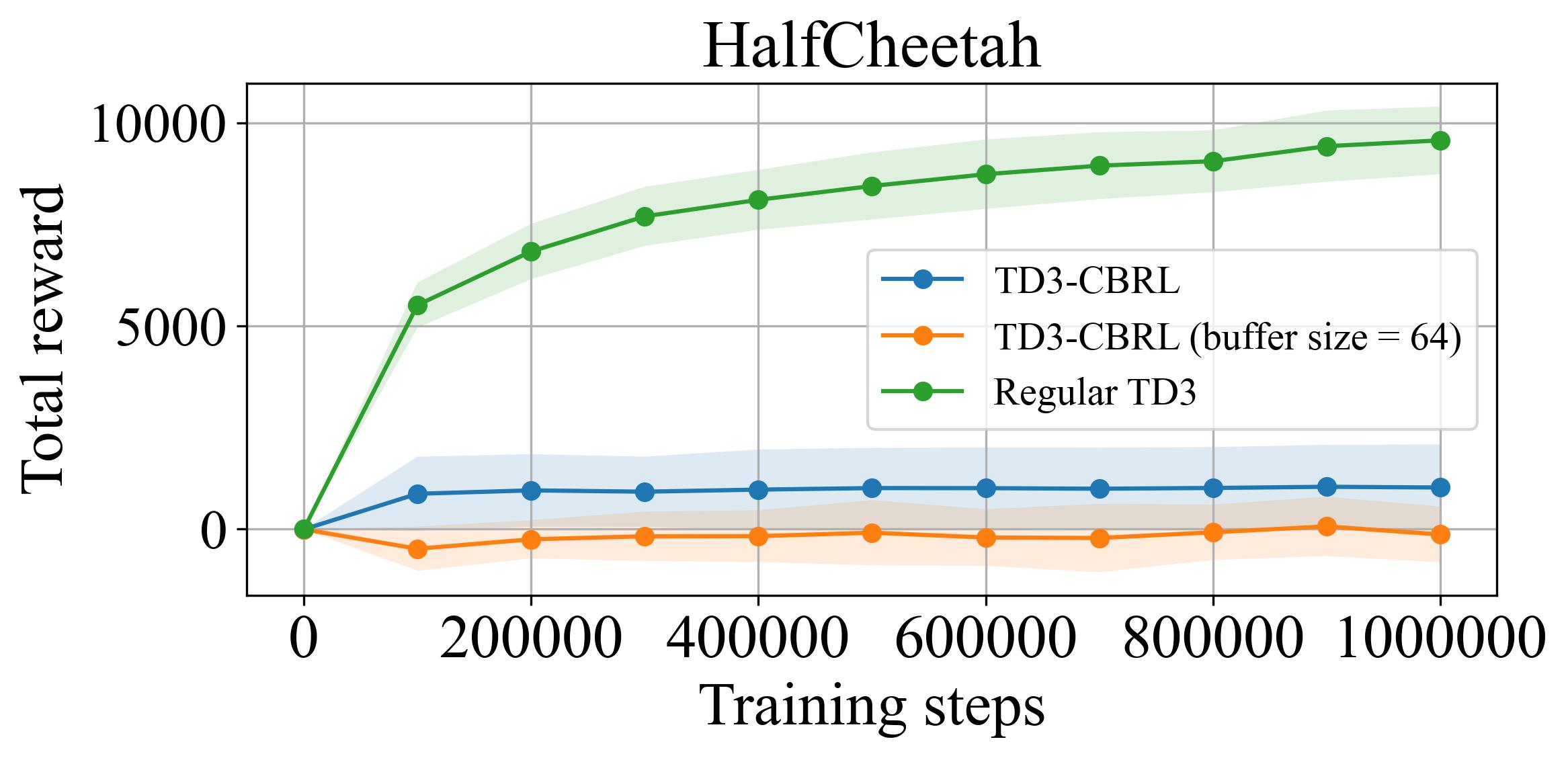}
    \subcaption{Half Cheetah}
  \end{minipage}
\\ \\ 
  \begin{minipage}[t]{0.49\linewidth}
    \centering
    \includegraphics[bb=0 0 600 300, scale=0.3]{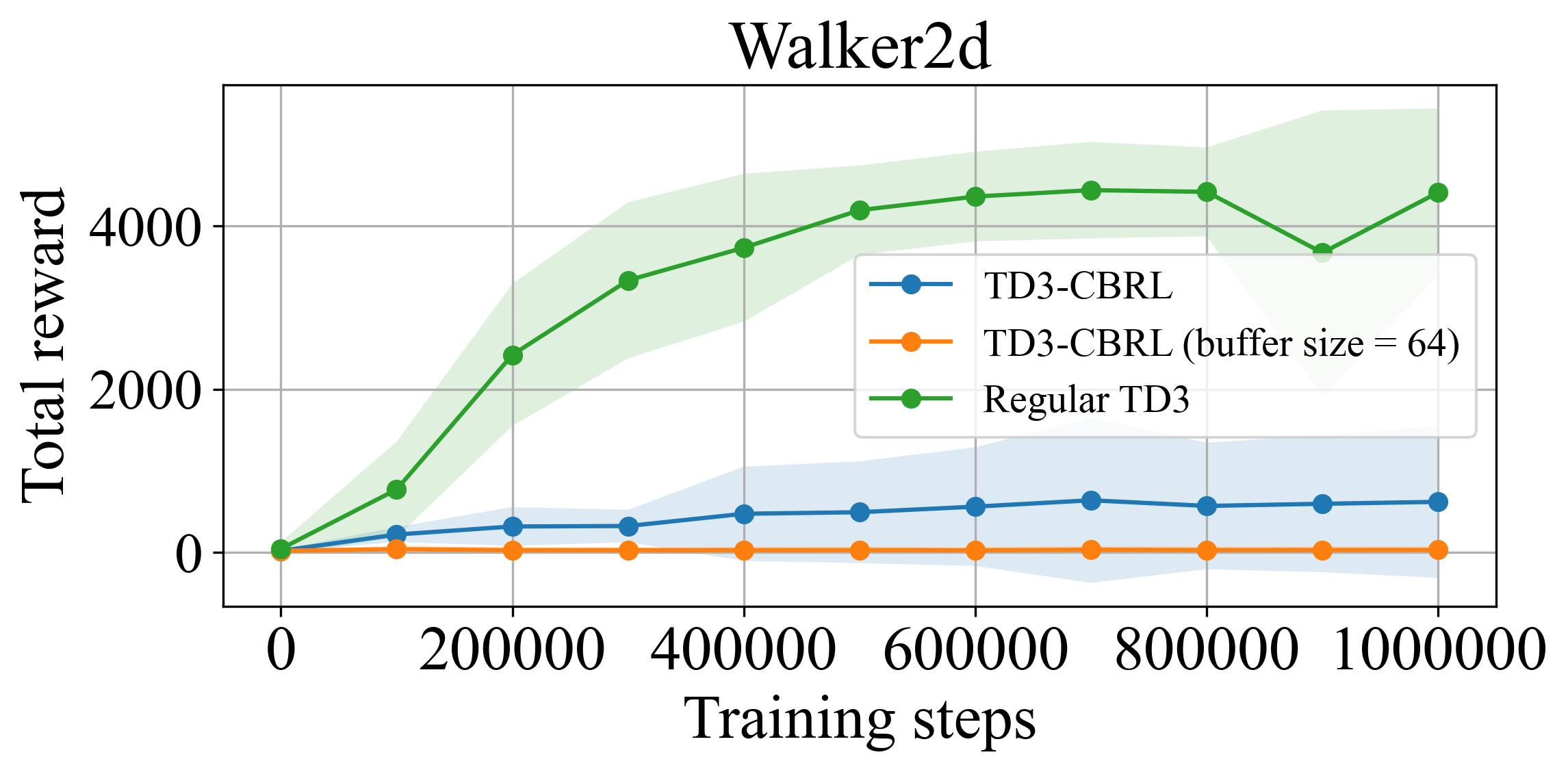}
    \subcaption{Walker2d}
  \end{minipage}
  \begin{minipage}[t]{0.49\linewidth}
    \centering
    \includegraphics[bb=0 0 600 300, scale=0.3]{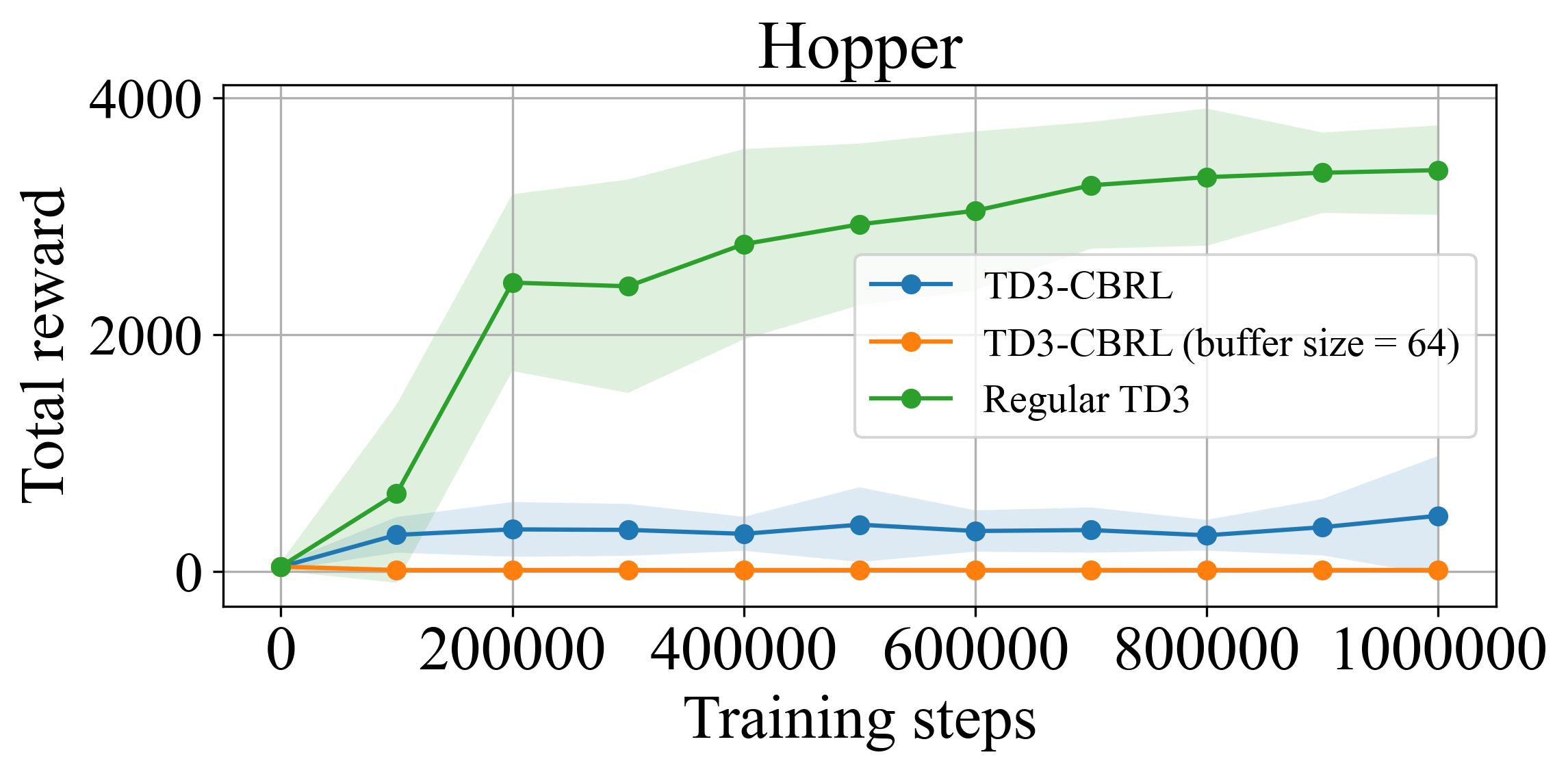}
    \subcaption{Hopper}
  \end{minipage}
  \caption{Learning performance for the MuJoCo continuous control tasks of Open AI Gym. The vertical axis shows the average total reward from the results with ten different random number seeds.
The horizontal axis shows the training steps.
The blue, orange, and green lines show the average total rewards of 
TD3-CBRL, TD3-CBRL with a replay buffer size of 64, and regular TD3, respectively.
Each shaded area shows its standard deviations.
The results of regular TD3 were obtained from \url{https://github.com/sfujim/TD3}.}
\label{fig:mujoco_results}
\end{figure}

The learning results are shown in Figure \ref{fig:mujoco_results}.
The figure shows that the agent exhibits comparable learning performance to the regular TD3 in learning tasks such as Pendulum, Inverted Pendulum, and Inverted Double Pendulum. 
On the other hand, the agent failed to learn more complex tasks, such as HalfCheetah, Walker2d, and Hopper. 
These results demonstrate that TD3-CBRL has successfully learned tasks that are more challenging than those in previous studies while simultaneously highlighting the current limitations of CBRL agents.
Moreover, the figure reveals that setting the replay buffer size to 64 results in learning failure for all tasks. This emphasizes the importance of incorporating experience replay techniques for improving learning performance in CBRL.

\subsection{Partially observable Markov decision process (POMDP) task}
Environments whose dynamics does not follow the Markov property and where the learning agent can observe only partial and incomplete information regarding the current state are formally described as Partially Observable Markov Decision Processes (POMDPs). Algorithms using RNNs have been shown to be one of the effective approaches for reinforcement learning in such environments \cite{ni2021recurrent, meng2021memory}.
Since the CBRL agent uses an ESN as a source of chaotic dynamics, it possesses the potential to handle time-series processing. This section investigates whether the CBRL agent can use the short-term memory of the ESN to learn a POMDP task.
The experiments in the previous sections have focused on the characteristics of the chaotic ESN as a source of exploration, using environments satisfying the Markov property.
Here, we conduct learning on the ``flickering goal task" to investigate whether the TD3-CBRL agent can learn POMDP tasks.  The flickering problem is usually used to introduce partial observability to an environment in reinforcement learning tasks\cite{hausknecht2015deep, meng2021memory}. Although the basic environmental rules in this task are the same as in the goal task, the agent can only observe the state of the environment with a probability of $p_{\rm{obs}}=0.5$ in this task, except in the first step of an episode. In other words, the input $\bm{u}_t$ is replaced by a zero vector with probability $(1 - p_{\rm{obs}})$.
To enable the critic network $Q$ to estimate state-action values from the observation sequence, the output of the ESN was concatenated with the inputs $\bm{u}_t$ to critic network $Q$. Furthermore, we adjusted the spectral radius to $g = 1.2$ to ensure that the ESN fully exhibited not only the exploration capability but also the time-series processing capability and increased the number of learning steps to 50,000 to accommodate the increased task difficulty.

\begin{figure}[t]
  \begin{minipage}[t]{0.49\linewidth}
    \centering
    \includegraphics[bb=0 0 450 225, scale=0.35]{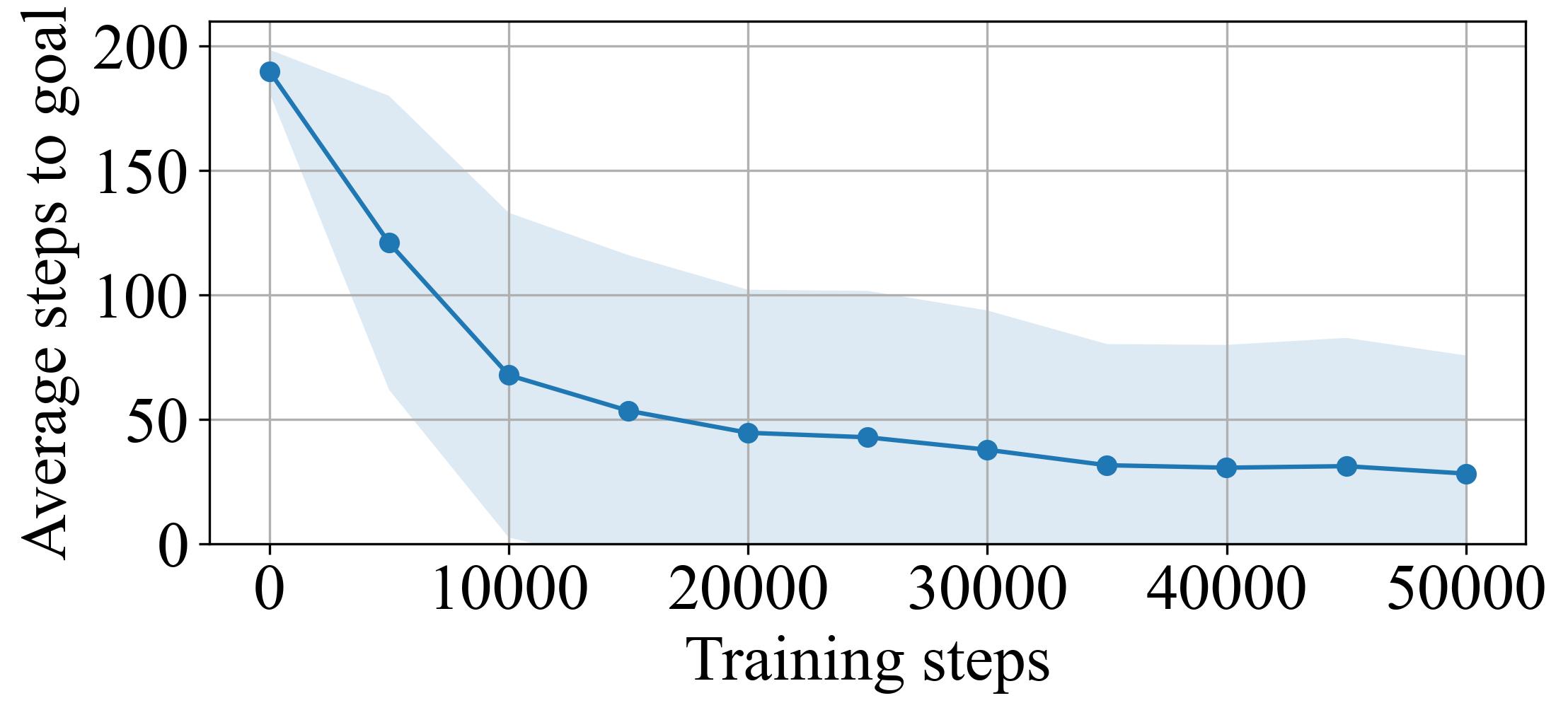}
    \subcaption{TD3-CBRL ($g=1.2$ and ESN is connected to $Q$)}
  \end{minipage}
  \begin{minipage}[t]{0.49\linewidth}
    \centering
    \includegraphics[bb=0 0 450 225, scale=0.35]{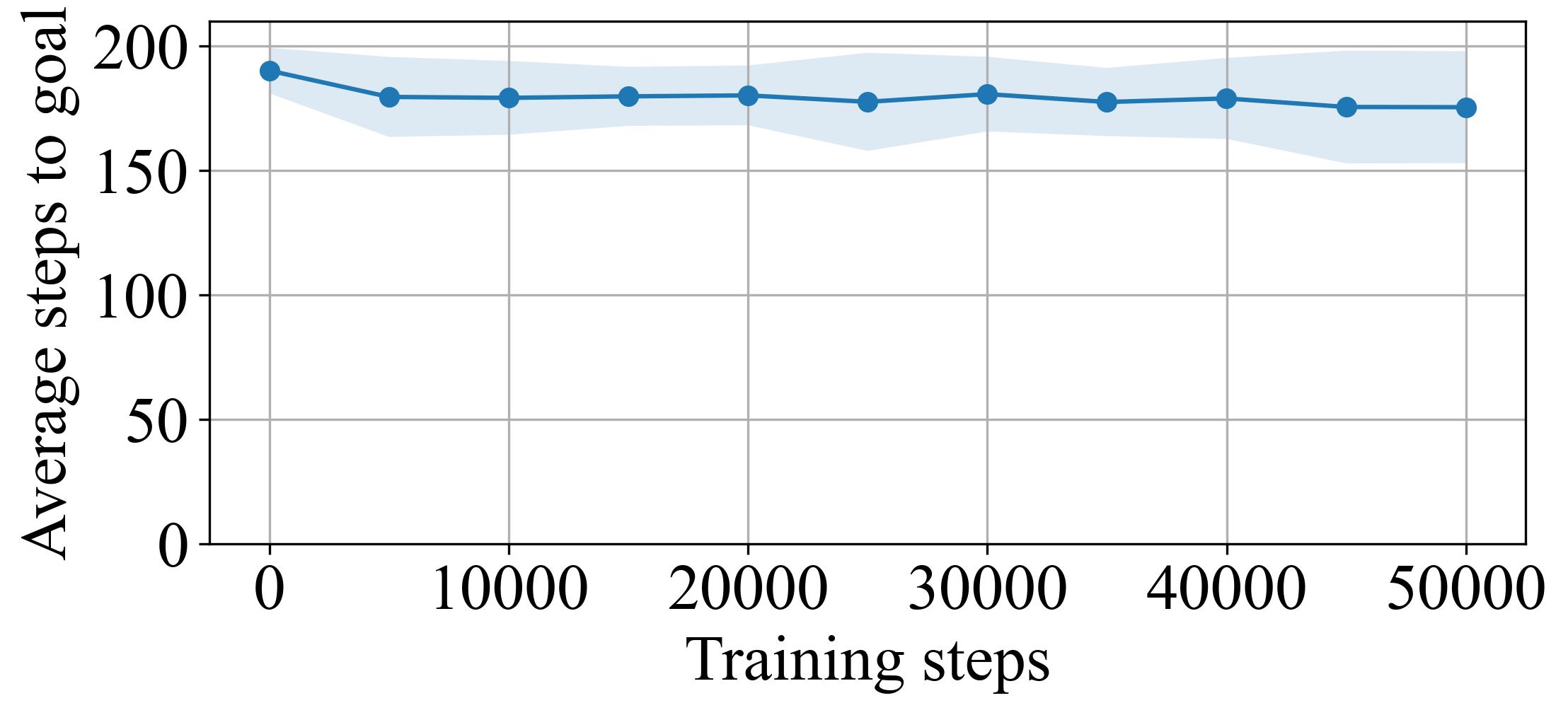}
    \subcaption{TD-CBRL ($g=1.2$)}
  \end{minipage}
\\ \\ \\
  \begin{minipage}[t]{0.99\linewidth}
    \centering
    \includegraphics[bb=0 0 450 225, scale=0.35]{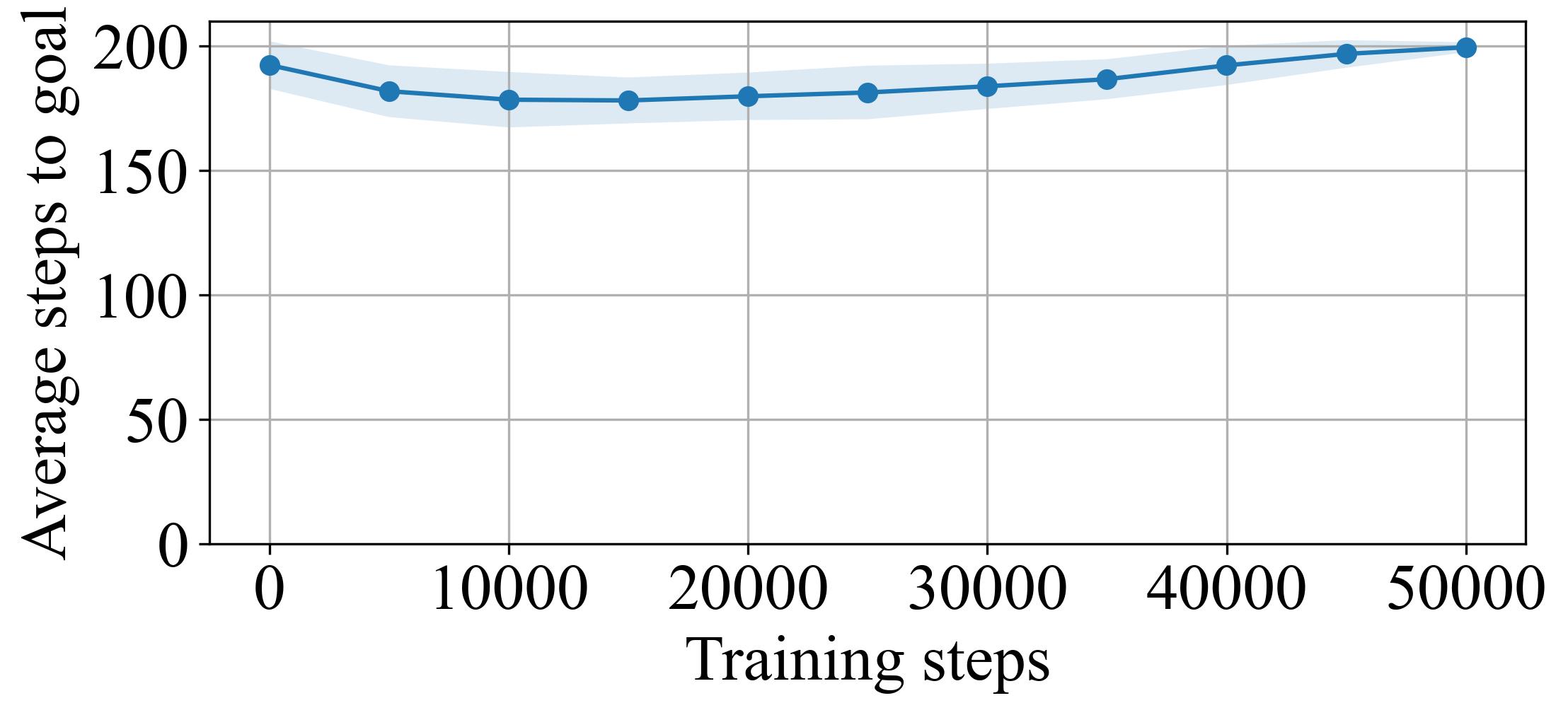}
    \subcaption{Regular TD3}
  \end{minipage}
  \caption{Learning curves of the flickering goal task. The definition of the line color is the same as in Fig. \ref{fig:LC}.
(a) shows the learning results with TD3-CBRL ($g=1.2$) in which the ESN was connected to the critic network $Q$.
(b) shows the learning results with TD3-CBRL ($g=1.2$).
(c) shows the learning results with regular TD3.}

\label{fig:LC_POMDP}
\end{figure}

Figure \ref{fig:LC_POMDP}(a) shows the learning curve for this experiment. This figure indicates that the number of steps to reach the goal decreases as learning progresses and that the TD3-CBRL agent is successfully learning the POMDP task. 
On the other hand, Fig. \ref{fig:LC_POMDP}(b) shows the results when the ESN was not connected to the critic network Q. In this case, the agent fails to learn. This result demonstrates that time-series processing by the ESN is essential for predicting state-action values in the flickering goal task. Furthermore, as a comparison, Fig. \ref{fig:LC_POMDP}(c) shows the learning results using a standard TD3 algorithm with added external exploration noise. This figure shows that regular TD3 agent fails to learn the flickering goal task, indicating that models without time-series processing capabilities cannot learn this task. These results suggest that TD3-CBRL agents have the potential to learn in situations where the current state provides insufficient information for decision-making.

\begin{figure}[t]
    \centering
    \includegraphics[bb=0 0 600 300, scale=0.4]{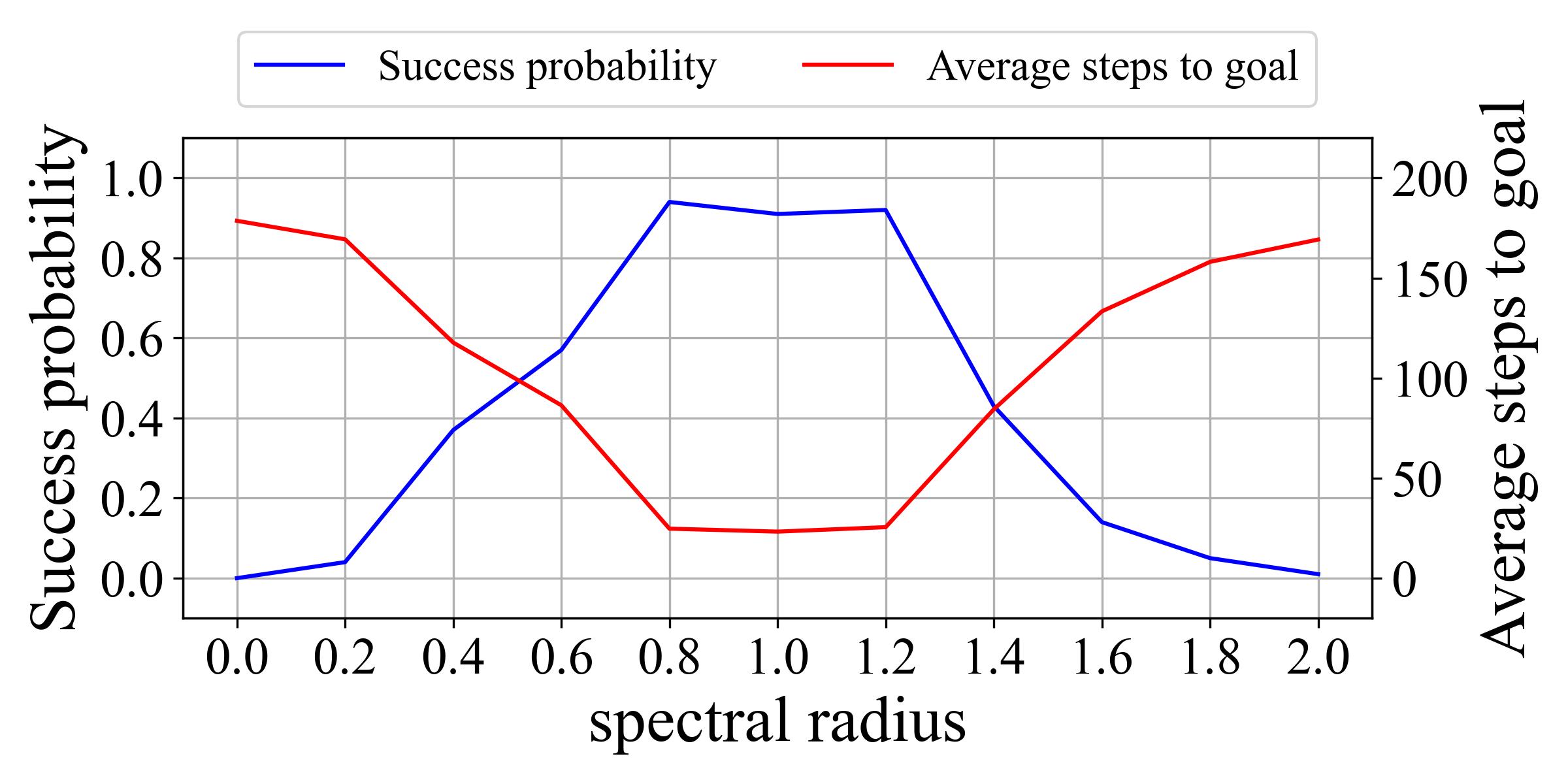}
  \caption{Learning performance of the flickering goal task with varying the spectral radius of the ESN in the CBRL-TD3 agent. The definitions of the line colors are the same as in Fig. \ref{fig:sr_score}. }
\label{fig:sr_score_pomdp}
\end{figure}

In time-series processing with ESNs, satisfying the Echo State Property (ESP) is a necessary condition for stable and reliable operation \cite{lukovsevivcius2012practical}. The ESP depends on the spectral radius $g$ of the recurrent weight matrix of the ESN. We investigated the influence of $g$ varying from 0 to 2.0 in increments of 0.2. The results of this experiment are presented in Figure \ref{fig:sr_score_pomdp}. The figure reveals a distinct peak in learning success probability centered around $g = 1.0$. This result suggests that too small $g$ probably diminishes the ESN's capacity to retain sufficient information from the observation history, limiting its memory of past input. In contrast, it suggests that too large $g$ induces highly sensitive temporal dynamics within the ESN, hindering the memory capability of past information necessary to learn the POMDP tasks.

These results suggest that while the TD3-CBRL agent with an ESN in its architecture can exhibit time-series processing capabilities, the trade-off between the short-term memory retention and the emergence of chaotic dynamics constrains the optimal range of $g$ to a relatively narrow one in learning POMDP tasks.

\section{Discussion}
This study introduced TD3 as a learning algorithm for chaos-based reinforcement learning (CBRL) and revealed several characteristics. 
It was confirmed that the TD3-CBRL agent can learn the goal task. This suggests that TD3 is a good candidate as a learning algorithm for CBRL, which had not been well-established in previous studies.
Although a regular TD3 agent succeeded in learning the goal task, the agent failed to learn it without external exploration noise, despite the simplicity of the task. On the other hand, the TD3-CBRL agent, whose actor model has chaotic dynamics, can learn without exploration with random noises. This comparison indicates that the internal chaotic dynamics of the reservoir contributes to the agent's exploratory behavior.

As learning progressed, the agent was able to autonomously suppress its exploratory behavior and transition to an exploitation mode. Furthermore, we confirmed that the agent resumed chaotic exploration and successfully re-learned when faced with an environmental change, such as changing the goal location during the learning process. However, the regular TD3, where the model does not have a reservoir and exploration is driven by external random numbers, failed this re-learning task. This suggests that exploration based on the reservoir's chaotic dynamics has properties that external random noise added to the action output does not possess.

The varying experiments revealed a suitable range of chaos strength in the agent's model for the TD3-CBRL agent to switch between exploration and exploitation modes autonomously. Specifically, it was found that when the spectral radius is too large, or the reservoir states are almost completely random, the number of steps for re-learning becomes excessively large. This increase in required steps is mitigated by reducing the buffer size, although the learning performance tends to decrease. This suggests that the reservoir states stored in the replay buffer before the environmental change are affecting the re-learning process.
In regular TD3, experiences stored before the rule change may be inappropriate for re-learning because they do not reflect the environment after the rule change. However, in CBRL, where the reservoir states are stored in the experience buffer, the reservoir states with an appropriately tuned $g$ can retain short-term memory of the state sequence from the start position. This enables ensuring the temporal continuity of experience before and after environmental changes. On the other hand, when the reservoir's chaoticity is too strong or its output is almost completely random, it only provides irregular noise. It seems to be the reason why TD3-CBRL agents with excessive reservoir chaoticity fail to adapt to environmental changes and show reduced re-learning performance.

It has also been confirmed that the dynamics of the reservoir should not be convergent. In experiments where external noise was added to the TD3-CBRL for exploration, it was observed that even with the promotion of exploration by external noise, the agent failed to re-learn when $g$ was small. This result suggests that it is not enough for the reservoir to simply act as a state sequence memory layer. We observed the reservoir states in the re-learning task by applying PCA that reduces its dimensionality. The CBRL agent with convergent dynamics ($g=0.9$) failed to re-learn because its reservoir state could not escape from the fixed-point attractor. On the other hand, the agent with $g=2.2$ acquired a new behavior to reach the new goal after the reservoir exhibited a new irregular trajectory. This result suggests that chaotic reservoirs are essential for re-learning, as they generate rich dynamics that create a variety of background activities within the system, incorporating the influence of the input and enabling the formation of new attractors leading to the desired goal.

These results are consistent with the criticality hypothesis of the brain in neuroscience \cite{Beggs, cocchi2017criticality}. The result observed in this study, that learning of the CBRL agents is optimized within a specific range of the spectral radius $g$, supports the functional advantages of this criticality in the context of reinforcement learning.
While further investigation on task dependency is necessary,  the chaos-based exploration exhibited more effective exploration capabilities than the external random number-based exploration in response to environmental changes in the goal task of this study.
The agent's internal chaotic dynamics can be regarded as a computational model of the brain's spontaneous activity, which generates diverse patterns of activity.
This activity seems to provide the fluctuations necessary for exploration, maintaining the temporal correlations of inputs in reservoir states, and being a resource that drives new exploratory behavior when encountering unknown experiences.

\section{Conclusion and Future Work}
CBRL is studied with the expectation of understanding a possible role of chaotic fluctuations on biological brains, realizing intelligent systems based on transient dynamics, and developing reinforcement learning algorithms that can optimize exploratory behavior through learning. 
This study extended the foundation of CBRL's learning algorithm by incorporating TD3, and we confirmed the existence of an optimal range of chaos intensity for the best balance between exploration and exploitation. Furthermore, we demonstrated that our approach enables the learning of challenging tasks that were previously unmanageable, contributing to advancements in exploratory learning models utilizing chaotic fluctuations. However, further verification and overcoming challenges are necessary for the practical application of CBRL.

Various studies have shown that neural populations operate in a critical state \cite{Beggs, cocchi2017criticality, beggs2003neuronal, shi2022criticality}.
Reservoir network performance has also been shown to be optimized at the edges of chaos \cite{bertschinger2004real, Asada}.
In addition, a study in which ESNs performed chaos-based exploration learning also showed results suggesting that exploration and exploitation are balanced at the edges of chaos \cite{matsuki2020adaptive}.
It is a very important and promising future research direction to investigate how to quantify the chaoticity of CBRL agents. 
In this study, the spectral radius was not optimal around $g=1$, where the reservoir dynamics is typically on the edge of chaos. One hypothesis for explaining this result is that the spectral radius of the subsystem, the reservoir, needed to be larger to place the entire system's dynamics, including the environment, at the edge of chaos. It is important to verify this by focusing on the entire system's behavior, including the interaction between the CBRL agent and the environment.
Furthermore, clarifying the differences between chaos-driven and random-noise exploration is a crucial challenge for evaluating the effectiveness of chaos-based exploration methods. To this end, investigating the distribution and entropy of behaviors and states and comparing them with the results obtained by conventional exploration using random noise is expected to provide valuable insights.

Experimental results of MuJoCo continuous control tasks
indicate that TD3-CBRL can solve tasks that were difficult for the previous CBRL approaches. However, it was also found that TD3-CBRL is limited in its ability to learn tasks with high-dimensional state and action spaces.
It is an important future problem to gain deeper insights by using various tasks. For example, tasks that require handling intermediate outputs could be considered. The optimal action in this study's goal tasks is maximizing the output to reach the goal, which did not necessitate intermediate outputs, thus failing to fully utilize the advantages of TD3 capable of handling continuous action outputs. As tasks that require intermediate outputs, we may consider tasks that treat the agent's acceleration as the action or tasks that impose penalties on large movements.
Furthermore, it is necessary to conduct further investigations across a broader range of environments, rather than using a simple goal task, to evaluate how learning is affected by various conditions, such as tasks with a long time horizon, sparse rewards, multi-step planning, or the presence of observation noise.
By improving the model to learn more complex tasks and conducting experiments with a wider range of tasks, we can examine the learning system based on exploration by internal chaotic fluctuations and interactions with the environment from a new perspective.

Introducing the new reservoir structure proposed to extend its performance is useful to improve the learning performance of CBRL.
Reservoirs do not perform well with high-dimensional input such as images.
Several studies have proposed methods that use untrained convolutional neural networks for feature extraction \cite{tong2018reservoir, chang2020reinforcement}.
Introducing these methods can enable CBRL agents to learn tasks with high-dimensional input, such as raw images.
Structural improvement that constructs reservoirs in multiple layers \cite{gallicchio2017deep, gallicchio2017echo, ma2017deep} and methods that reduce the model size by using multi-step reservoir states as input to readouts \cite{sakemi2020model} have been proposed.
It is worth verifying the use of these new reservoir techniques to improve the performance of CBRL agents.

Improvements in learning algorithms are also worth considering. Sensitivity adjustment learning (SAL), which adjusts the chaoticity of neurons based on "sensitivity," has been proposed \cite{Shibata2021SAL}.
This method can modify the recurrent weights while keeping the chaoticity of the recurrent neural networks.
Using SAL to maintain chaoticity and learning with Backpropagation Through Time may allow CBRL agents to learn more difficult tasks.
Self-modulated reservoir computing (SM-RC) that extends the reservoir network's ability by dynamically changing the characteristics of the reservoir and attention to the input through a self-modulated function has been proposed \cite{sakemi2024learning}.
Since SM-RC can adjust its spectral radius, it is also expected that CBRL agents with SM-RC can learn to change their chaoticity and switch between exploration and exploitation states more dynamically.
Furthermore, it is worth exploring approaches that combine the exploration driven by chaotic fluctuations with other algorithms aiming at efficient exploration, such as intrinsically motivated reinforcement learning and meta-learning, to guide the exploration in a more efficient direction.

Improving adaptability to non-stationary environments is a critical challenge in reinforcement learning, and a variety of methods have been proposed to address this issue. For instance, Steinparz et al. used intrinsic rewards to encourage exploration in response to environmental changes \cite{steinparz2022reactive}. Canonaco et al. proposed NSD-RL that actively detects environmental changes to re-adapt \cite{canonaco2020model}, and Zhu et al. proposed a method that autonomously rebalances exploration and exploitation based on the model uncertainty of the Q-network \cite{zhu2022adaptive}. 
While a detailed comparative analysis between these existing methods and CBRL remains a subject for future work, they are not mutually exclusive; combining them could yield synergistic effects. The inherent sensitivity of the chaotic dynamics utilized in CBRL may amplify the novelty of unknown states or facilitate the detection of subtle state changes arising from environmental shifts. Verifying the integration of CBRL with other algorithms designed for non-stationary environments is worthwhile future work for more efficient re-exploration and re-learning.

\section*{Acknowledgments}
The author would like to thank Prof. Katsunari Shibata for the useful discussions about this research.
This work was supported by
Moonshot R\&D Grant Number JPMJMS2021,
Institute of AI and Beyond of UTokyo,
the International Research Center for Neurointelligence (WPI-IRCN) at The University of Tokyo Institutes for Advanced Study (UTIAS),
Cross-ministerial Strategic Innovation Promotion Program (SIP), the 3rd period of SIP, Grant Numbers JPJ012207, 
JST PRESTO Grant Number JPMJPR22C5,
JSPS KAKENHI Grant Numbers JP22K17969, JP22KK0159.

\appendix
\section{Varying learning rate}
\begin{figure}[H]
  \begin{minipage}[t]{0.99\linewidth}
    \centering
    \includegraphics[bb=0 0 600 300, scale=0.45]{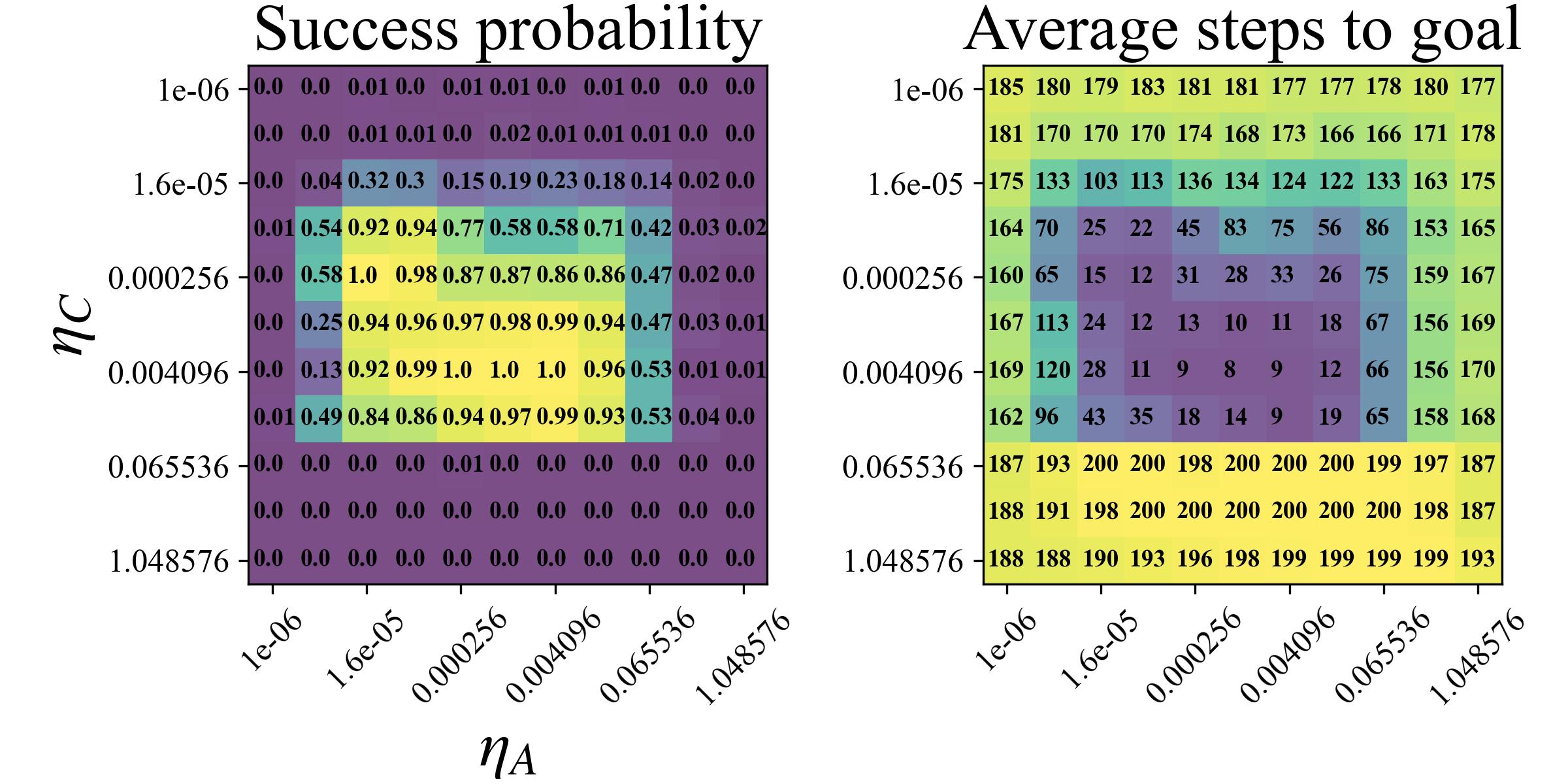}
    \subcaption{TD3-CBRL agents.}
  \end{minipage}
  \\ \\
  \begin{minipage}[t]{0.99\linewidth}
    \centering
    \includegraphics[bb=0 0 600 300, scale=0.45]{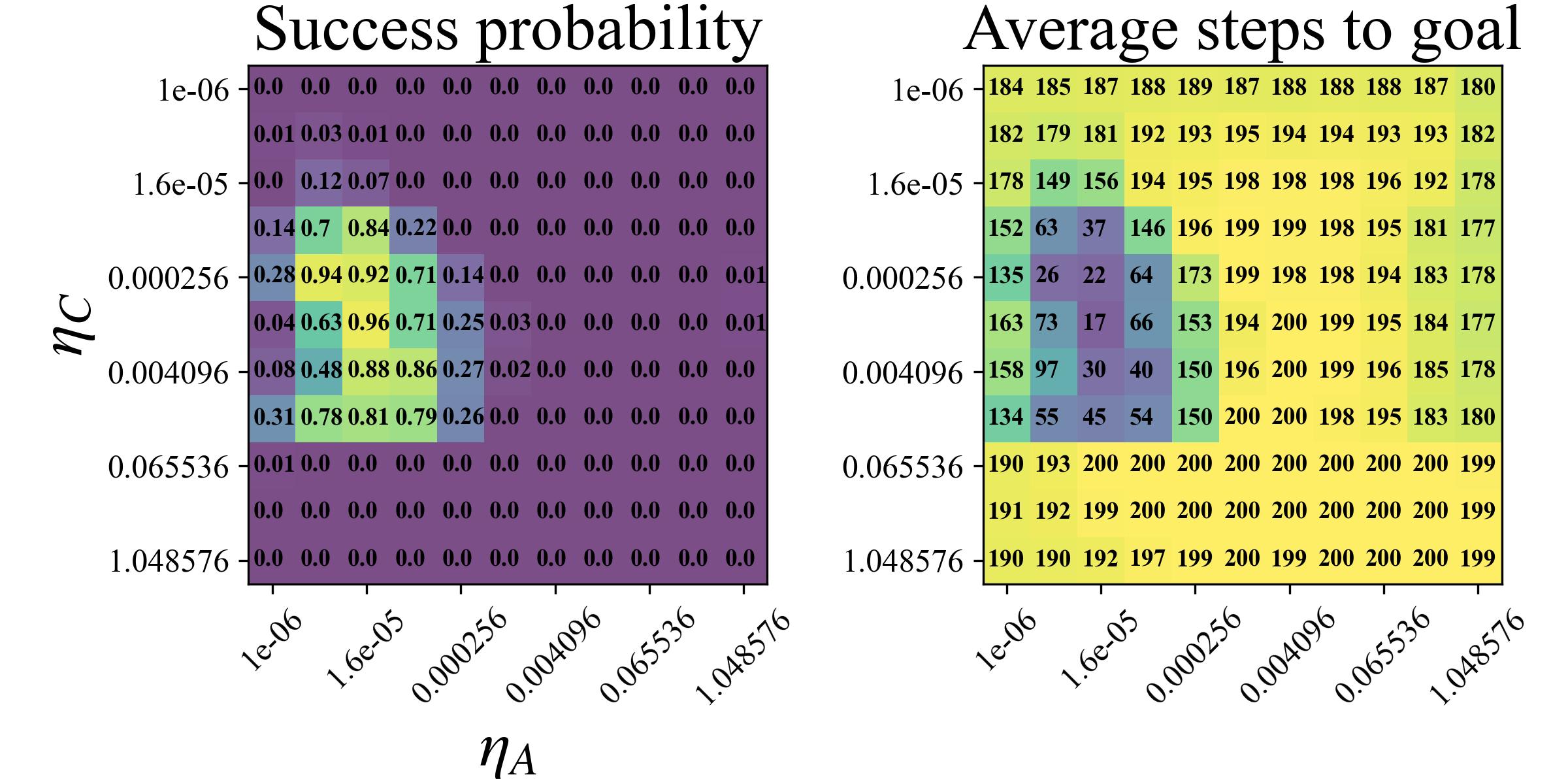}
    \subcaption{Regular TD3 agents.}
  \end{minipage}
  \caption{Learning performance with varying learning rate. The vertical axis indicates the learning rate of the critic network $\eta_C$ and the horizontal axis indicates the learning rate of the actor network $\eta_A$. The numbers in each cell show the probability of success and the average steps to the goal calculated with 100 different random seeds.
}
\label{fig:sr_score_lr}
\end{figure}

It is generally known that the performance of reinforcement learning is highly sensitive to hyperparameters. Here, we investigated the impact of the learning rate $\eta$ on performance and determined the optimal learning rate for our experiments. We varied the learning rates of the actor network $\eta_A$ and critic network $\eta_C$ of both the TD3-CBRL agent and the Regular TD3 agent with exploration noise by a factor of $10^{-6} \times 4^n$, where $n$ ranges from 0 to 10. Then, We measured the successful learning probability and the average number of steps taken. To ensure statistical validity, we conducted experiments with 100 different random seeds. The results are shown in Figure \ref{fig:sr_score_lr}. Based on these results, we set both  $\eta_A$ and  $\eta_C$ to $5 \times 10^{-4}$ for TD3-CBRL and readjusted $\eta_A$ to $1.6 \times 10^{-5}$ for Regular TD3 in our experiments.

\section{Varying the exploration noise}
\begin{figure}[t]
  \begin{minipage}[t]{0.49\linewidth}
    \centering
    \includegraphics[bb=0 0 600 300, scale=0.3]{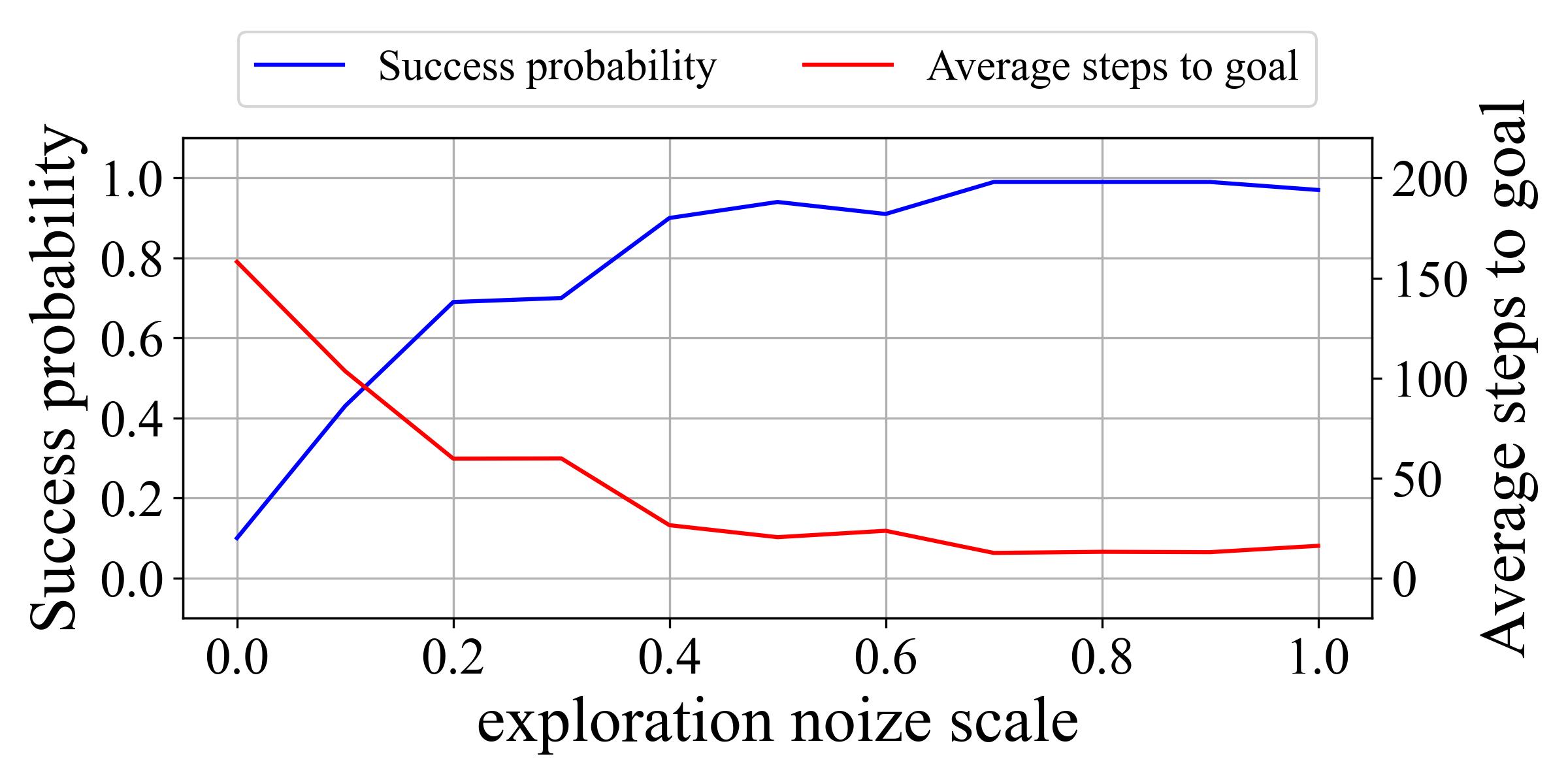}
    \subcaption{Goal task.}
  \end{minipage}
  \begin{minipage}[t]{0.49\linewidth}
    \centering
    \includegraphics[bb=0 0 600 300, scale=0.3]{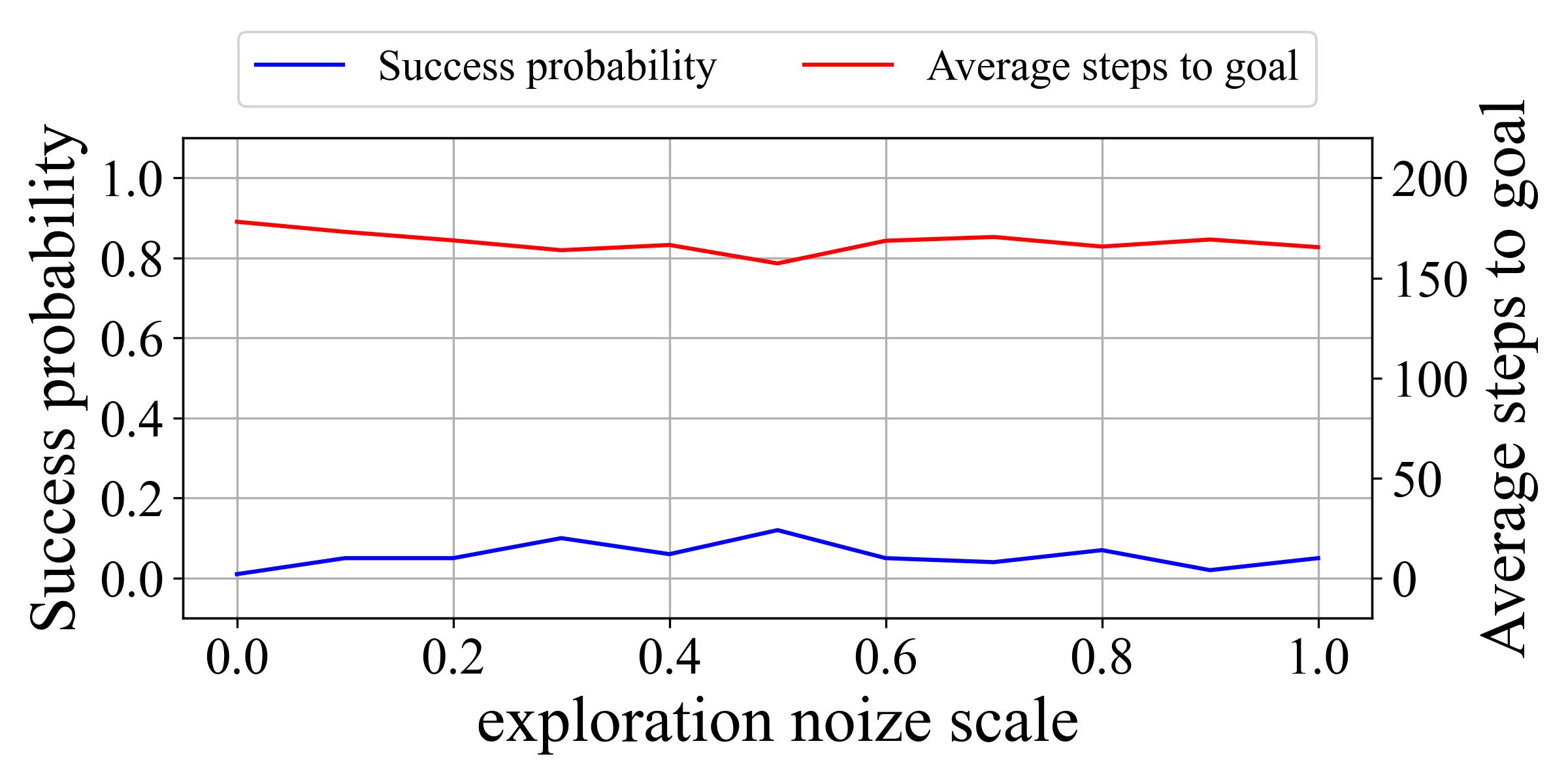}
    \subcaption{Goal change task.}
  \end{minipage}
  \caption{Learning performance of regular TD3 agents with varying the scale of external exploration noise. The definitions of the line colors are the same as in Fig. \ref{fig:sr_score}. }
\label{fig:sr_score_ex_noise_scale}
\end{figure}

The exploration noise scale is a crucial hyperparameter in regular TD3. To investigate the impact of the noise scale on learning performance in our experiments, we varied the standard deviation of the exploration noise, $\epsilon^a$, from 0 to 1 with increments of 0.1. We then conducted experiments with 100 different random seeds and examined the successful learning probability and average steps to reach the goal. The results are shown in Figure \ref{fig:sr_score_ex_noise_scale}(a).
This figure indicates that a noise scale of approximately 0.5 or greater is desirable. The fact that learning performance does not degrade even when the scale is increased to the maximum action value of the task is likely due to the simplicity of the goal task used in this study, which involves reaching a goal on a plane.
Next, Figure \ref{fig:sr_score_ex_noise_scale}(b) presents the results of investigating the relationship between exploration noise and learning a goal-changing task. This figure reveals that regular TD3 fails to re-learn regardless of the noise scale value. This result implies that, at least for environmental changes such as goal position shifts, noise variations do not affect the success or failure of learning for the regular TD3. In this study, we conducted experiments with the standard deviation of 0.5 for the exploration noise $\epsilon^a$.

In DDPG, sampling exploration noise from an Ornstein Uhlenbeck (OU) process \cite{uhlenbeck1930theory} is proposed to introduce temporal correlation in exploration \cite{ddpg}. We investigated the learning performance of Regular TD3 with exploration driven by such temporally correlated noise in our exploration. 
The discretized form of the OU process is given as follows:
\begin{eqnarray}
 X_{t+\Delta t} = X_t + \theta(\mu - X_t)\Delta t + \sigma \sqrt{\Delta t}  \varepsilon_t,
\label{eq:OUprocess}
\end{eqnarray}
where $X_t$ is the value of the OU process at time $t$, $\mu=0$ is the mean value of $X_t$ in long term, $\theta$ is the speed of reversion to the mean value, $\sigma$ is the constant value that determines the volatility, $\Delta t$ is the discrete time step, $\varepsilon_t$ is a random value drawn from $\mathcal{N}(0,1)$.

\begin{figure}[t]
  \begin{minipage}[t]{0.99\linewidth}
    \centering
    \includegraphics[bb=0 0 600 300, scale=0.45]{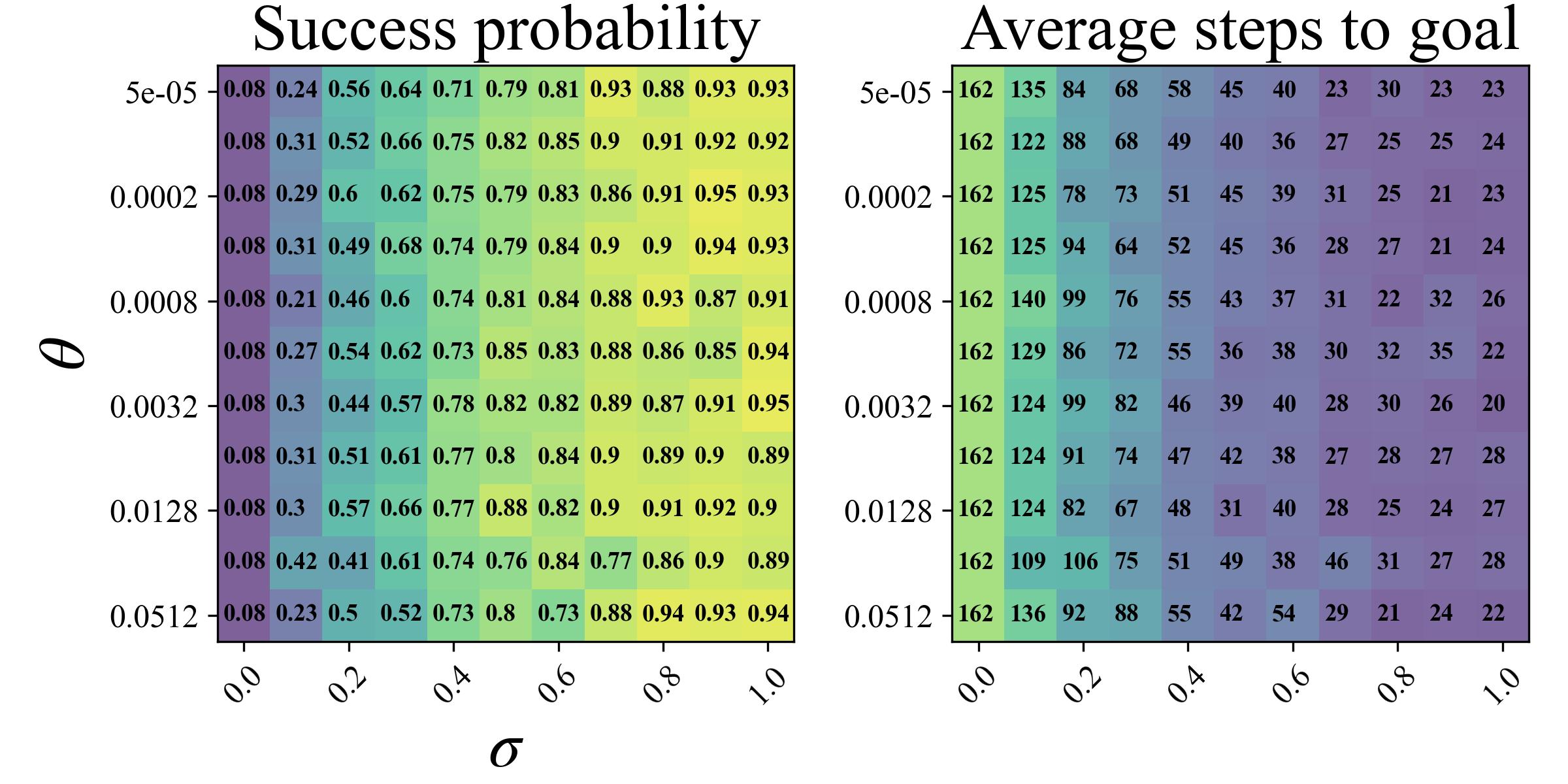}
  \end{minipage}
  \\ 
  \begin{minipage}[t]{0.99\linewidth}
    \centering
    \includegraphics[bb=0 0 600 300, scale=0.45]{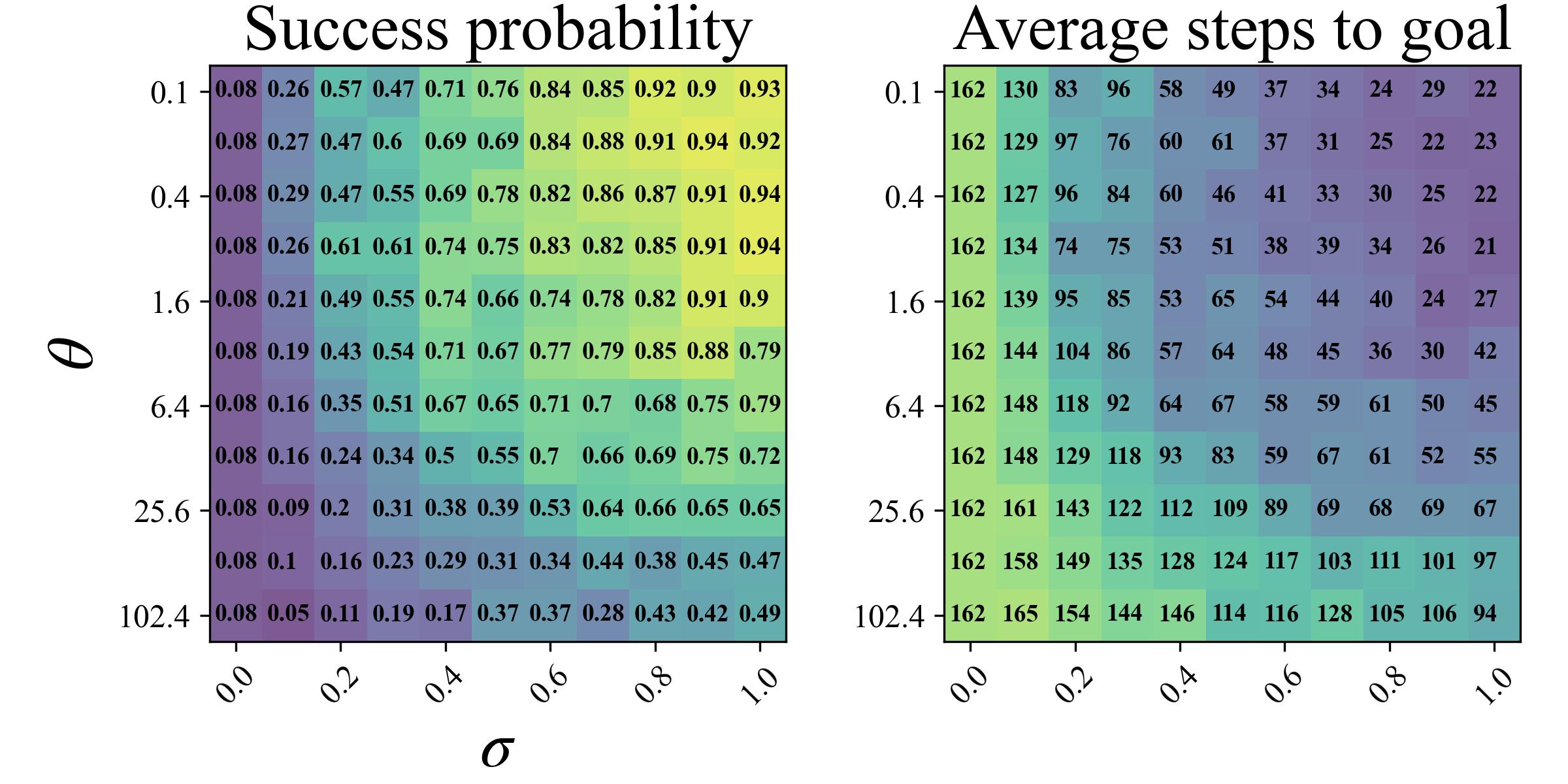}
  \end{minipage}
  \caption{Learning performance of regular TD3 agents using the OU process exploration with varying $\theta$ and $\sigma$. The vertical axis indicates $\theta$ and the horizontal axis indicates $\sigma$. The data representation is the same as in Fig. \ref{fig:sr_score_lr}.
}
\label{fig:sr_score_ou_theta}
\end{figure}

\begin{figure}[t]
  \begin{minipage}[t]{0.99\linewidth}
    \centering
    \includegraphics[bb=0 0 600 300, scale=0.45]{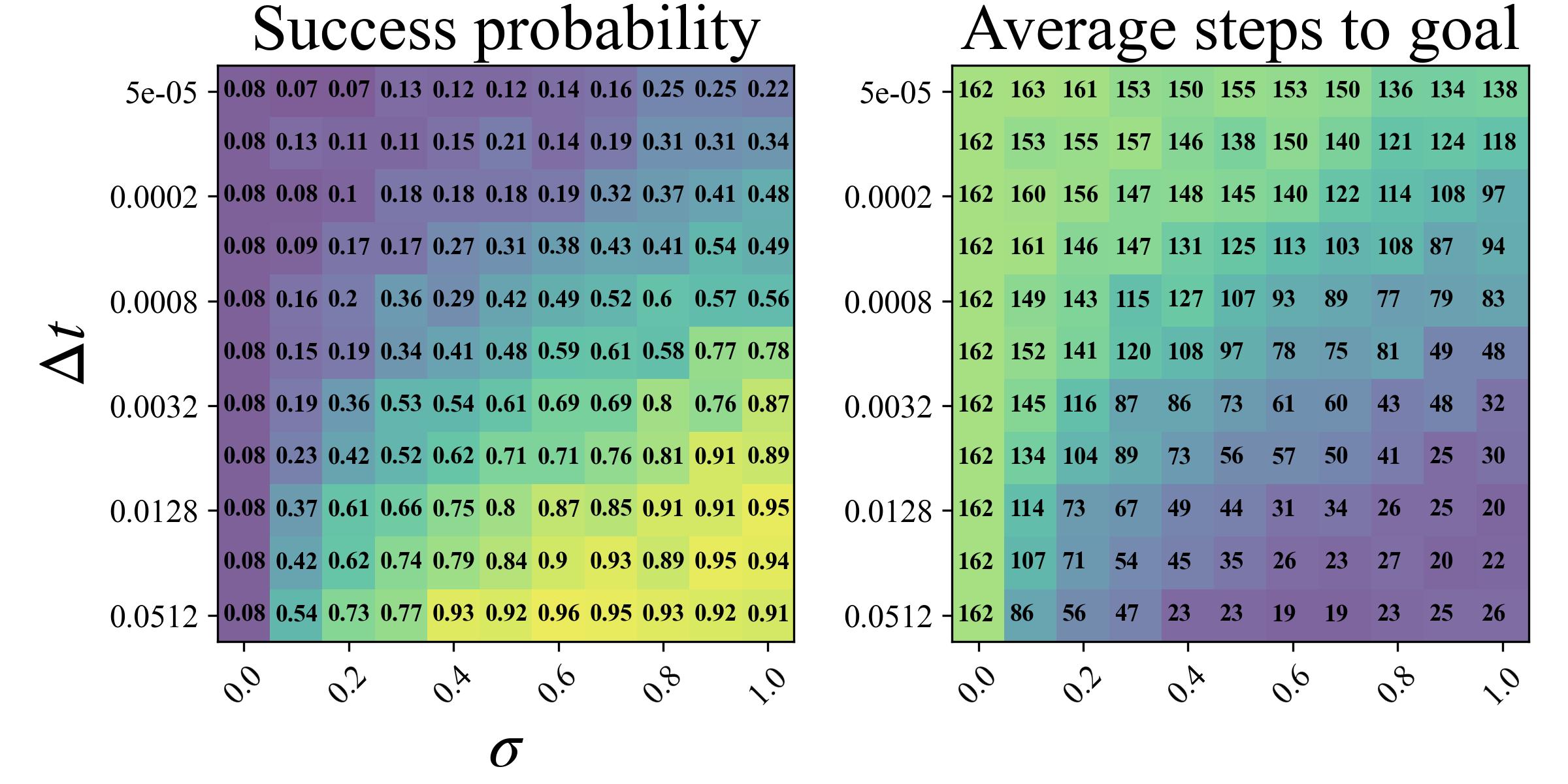}
  \end{minipage}
  \\ 
  \begin{minipage}[t]{0.99\linewidth}
    \centering
    \includegraphics[bb=0 0 600 300, scale=0.45]{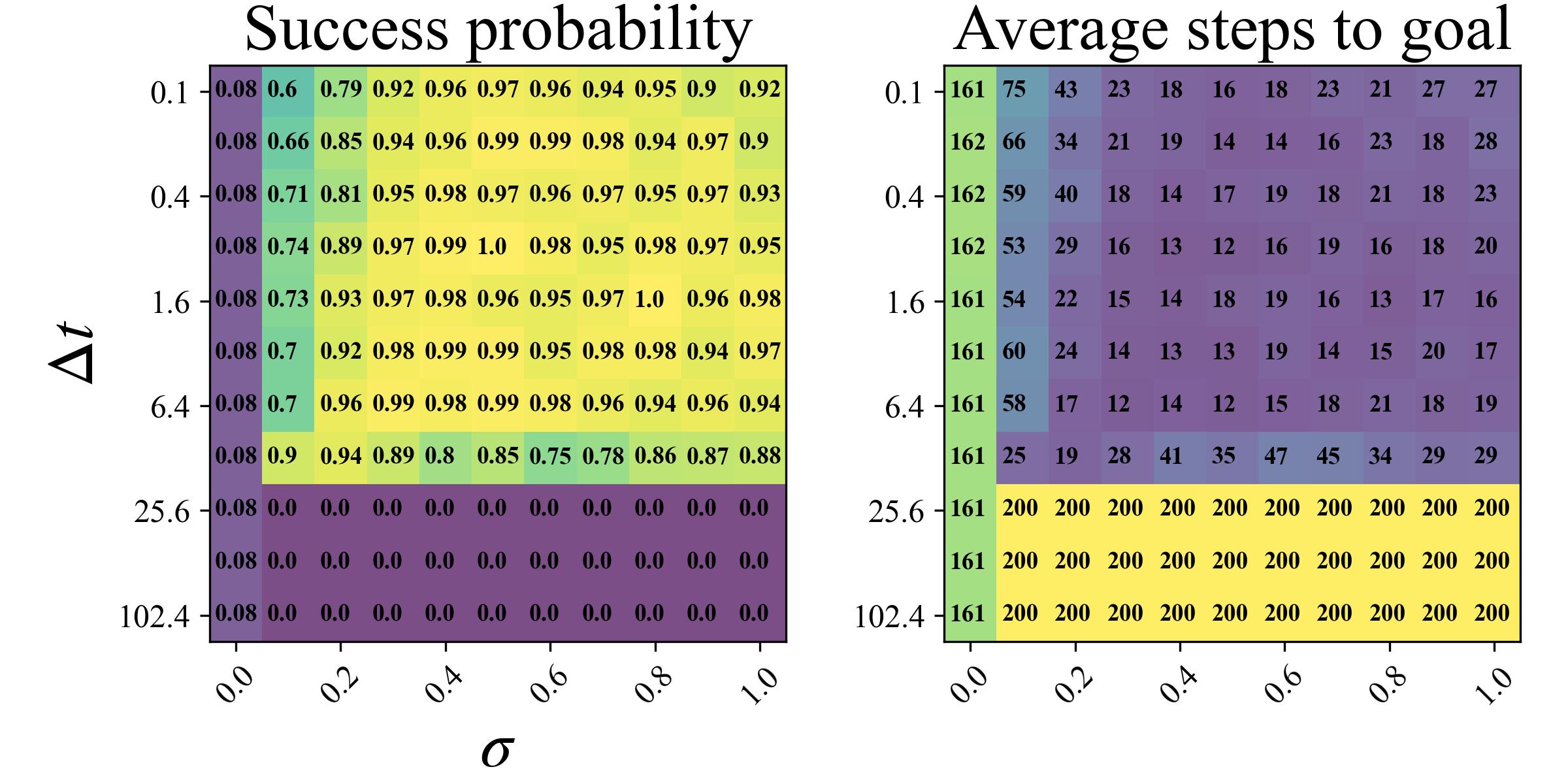}
  \end{minipage}
  \caption{Learning performance of regular TD3 agents using the OU process exploration with varying $\Delta t$ and $\sigma$. The vertical axis indicates $\Delta t$ and the horizontal axis indicates $\sigma$. The data representation is the same as in Fig. \ref{fig:sr_score_lr}.
}
\label{fig:sr_score_ou_dt}
\end{figure}

\begin{figure}[t]
  \begin{minipage}[t]{0.99\linewidth}
    \centering
    \includegraphics[bb=0 0 600 300, scale=0.45]{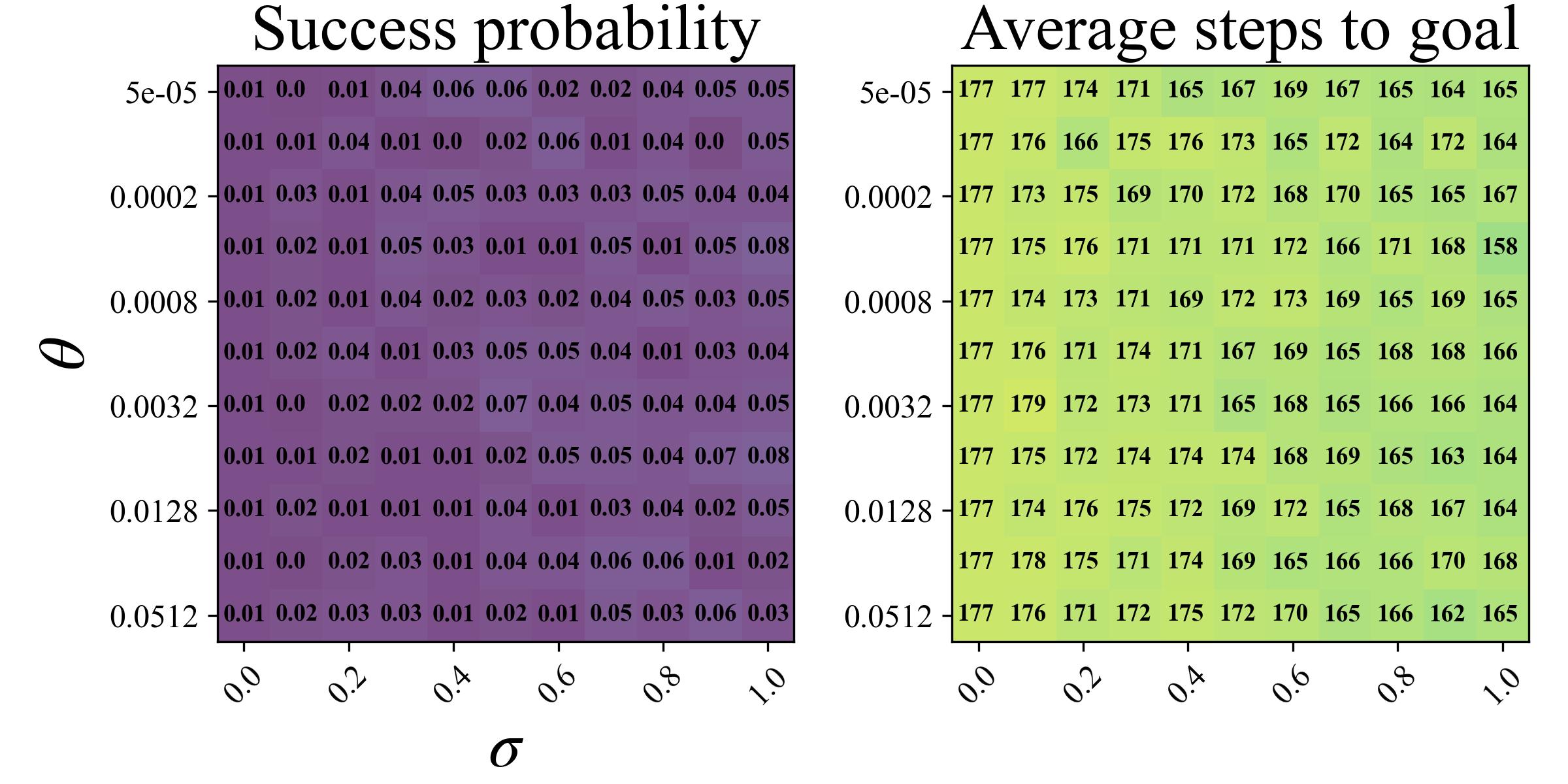}
  \end{minipage}
  \\ 
  \begin{minipage}[t]{0.99\linewidth}
    \centering
    \includegraphics[bb=0 0 600 300, scale=0.45]{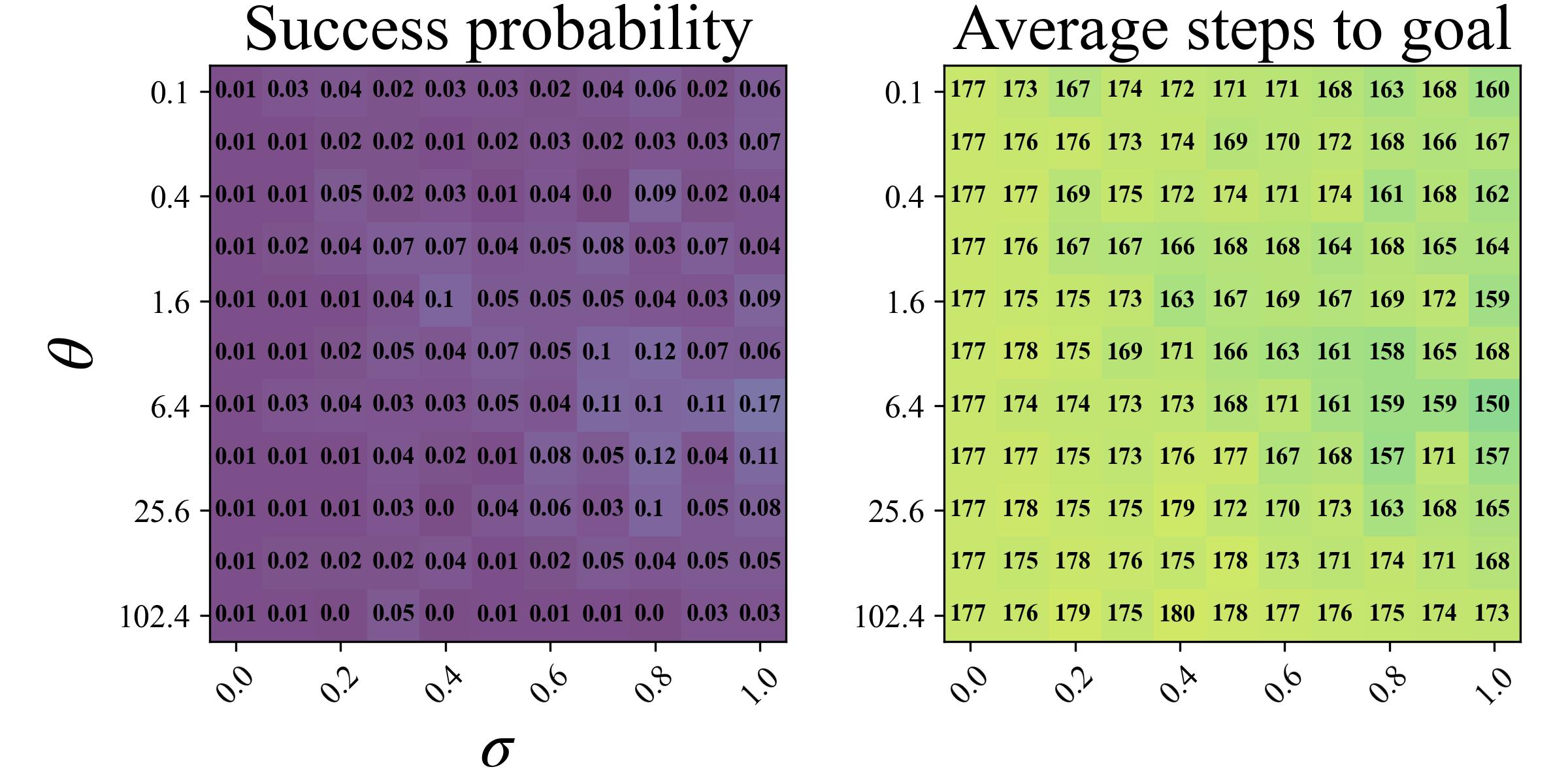}
  \end{minipage}
  \caption{Learning performance of regular TD3 agents using the OU process exploration in the goal change task with varying $\theta$ and $\sigma$. The vertical axis indicates $\theta$ and the horizontal axis indicates $\sigma$. The data representation is the same as in Fig. \ref{fig:sr_score_lr}.
}
\label{fig:sr_score_ou_theta_goal_change}
\end{figure}

\begin{figure}[t]
  \begin{minipage}[t]{0.99\linewidth}
    \centering
    \includegraphics[bb=0 0 600 300, scale=0.45]{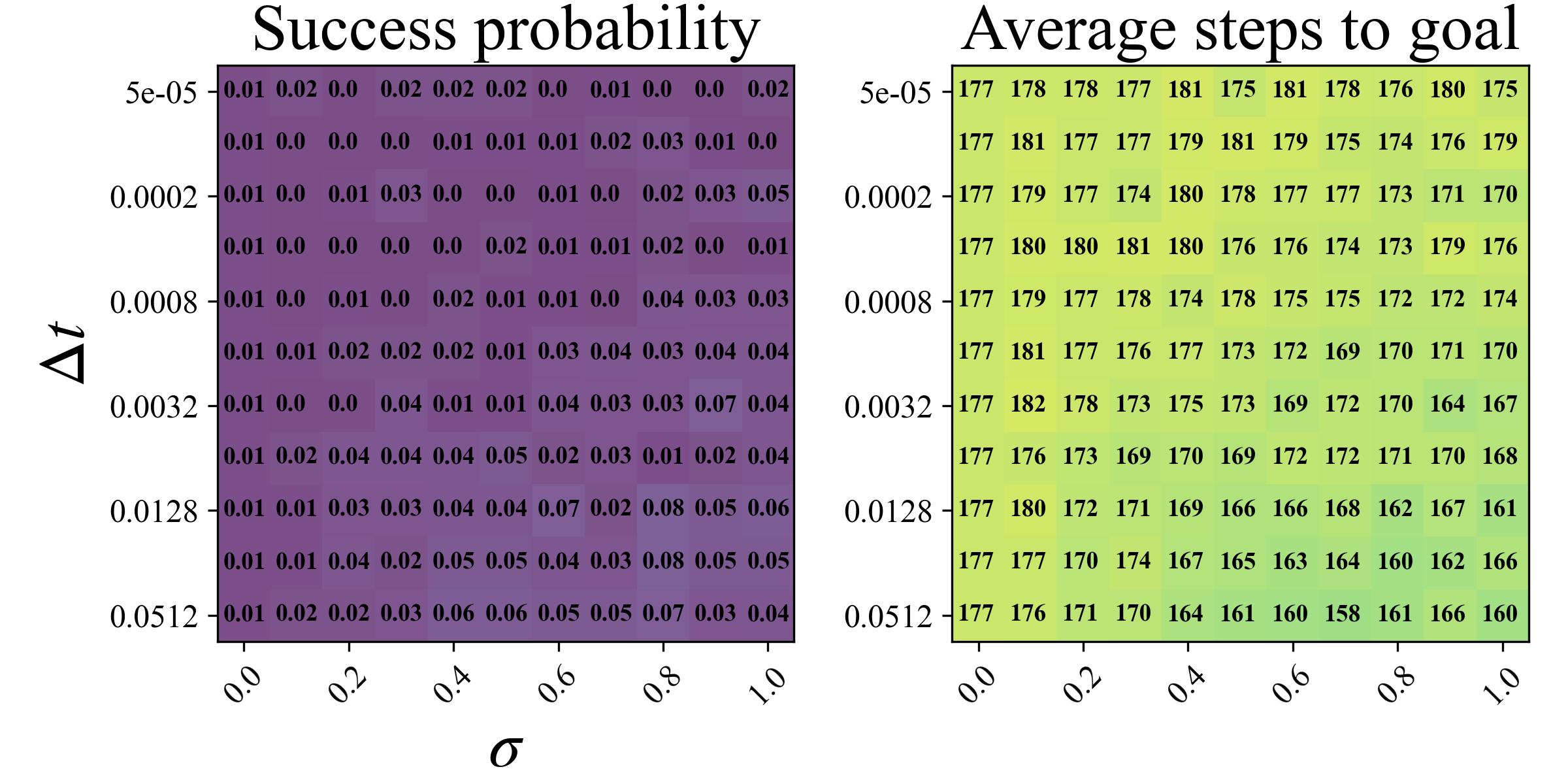}
  \end{minipage}
  \\ 
  \begin{minipage}[t]{0.99\linewidth}
    \centering
    \includegraphics[bb=0 0 600 300, scale=0.45]{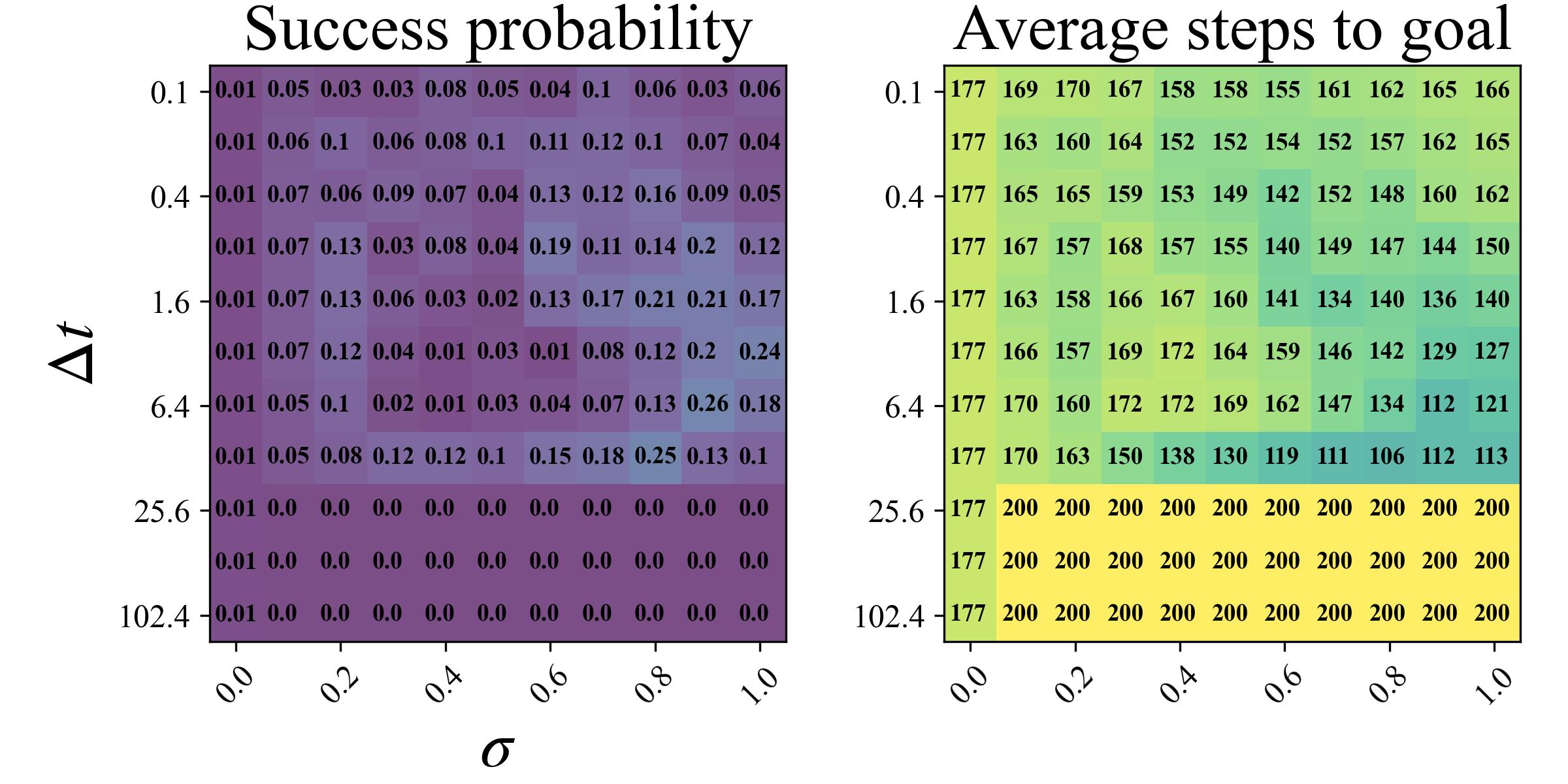}
  \end{minipage}
  \caption{Learning performance of CBRL agents in the goal change task with varying $\Delta t$ and $\sigma$ of the OU process. The vertical axis indicates $\Delta t$ and the horizontal axis indicates $\sigma$. The data representation is the same as in Fig. \ref{fig:sr_score_lr}.
}
\label{fig:sr_score_ou_dt_goal_change}
\end{figure}

We evaluated the learning performance of the agent on the goal task and the goal change task with varying the parameters $\sigma$, $\theta$, and $\Delta t$. Figure \ref{fig:sr_score_ou_theta} shows the successful learning probability and the average steps to reach the goal in the goal task when $\Delta t$ was set to $0.01$ and $\sigma$ was varied from $0$ to $1.0$ in increments of $0.2$, and $\theta$ was exponentially varied from $0.00005$ to $0.0512$ and from $0.1$ to $102.4$.
This figure shows a trend where the agent succeeds when $\theta \leq 6.4$ and $\sigma$ is large. The performance degradation at larger values of $\theta$ seems to be caused by a weakening of the noise variance due to a strong reversion force. Furthermore, the trend for learning to be more successful with larger values of $\sigma$ is likely due to the simplicity of the goal task, similar to the case with Gaussian noise.
Figure \ref{fig:sr_score_ou_dt} shows the results when $\theta$ was set to $0.15$ and $\Delta t$ was exponentially varied from $0.00005$ to $0.0512$ and from $0.1$ to $102.4$. This figure indicates that increasing the time step and the noise scale stabilizes learning within the range of $\Delta t \leq 12.8$. On the other hand, the agent failed to learn due to the divergence of the OU process values at $\Delta t \geq 25.6$.
Figures \ref{fig:sr_score_ou_theta_goal_change} and \ref{fig:sr_score_ou_dt_goal_change} show the results of similar validation experiments conducted on the goal change task. These figures demonstrate that even when varying $\theta$ and $\Delta t$, the agent failed to re-learn in response to environmental changes.
The result that the goal change task could not be learned even when using exploration with a temporally correlated noise process like OU noise suggests that other factors in addition to the temporal correlation of exploration are also important for TD3-CBRL to exhibit re-learning capability.

\section{Varying the random vector scale}
\begin{figure}[t]
  \begin{minipage}[t]{0.49\linewidth}
    \centering
    \includegraphics[bb=0 0 600 300, scale=0.3]{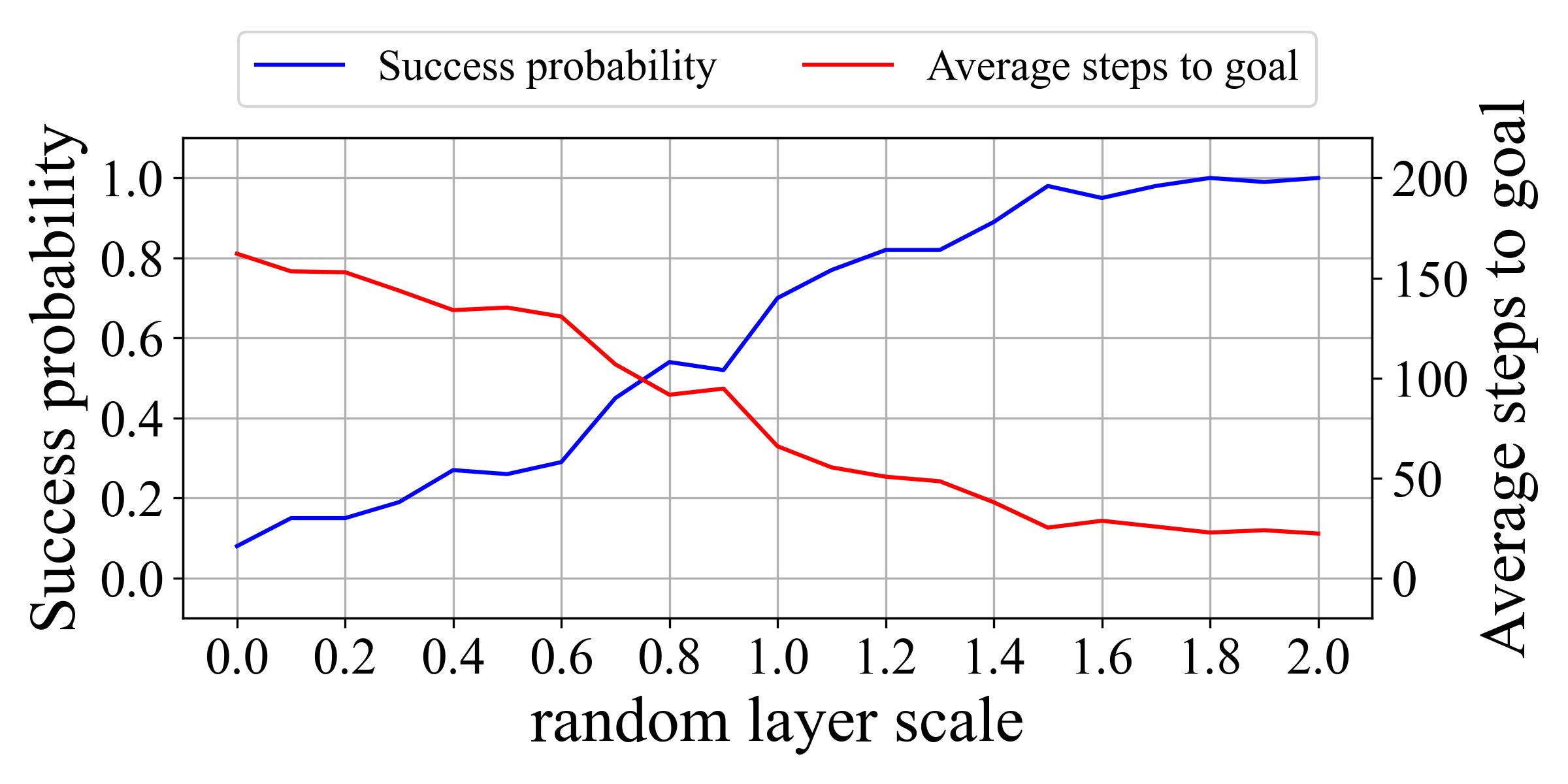}
    \subcaption{Goal task.}
  \end{minipage}
  \begin{minipage}[t]{0.49\linewidth}
    \centering
    \includegraphics[bb=0 0 600 300, scale=0.3]{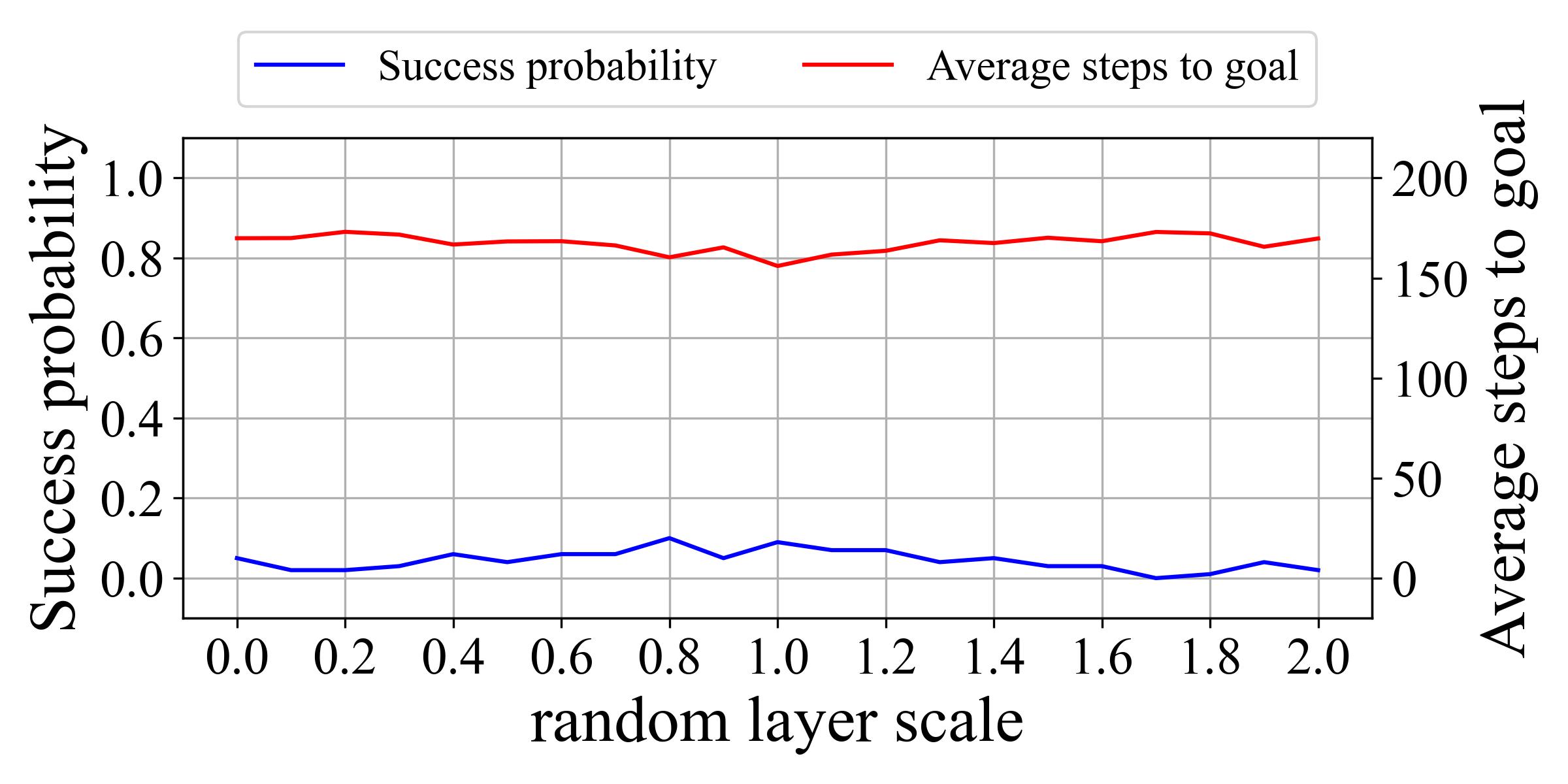}
    \subcaption{Goal change task.}
  \end{minipage}
  \caption{Learning performance of the agents using a random number vector with varying the scale of uniform distribution for the random vector layer. The definitions of the line colors are the same as in Fig. \ref{fig:sr_score}. (a) shows the learning results for the goal task. (b) shows the learning results of the goal change task. }
\label{fig:sr_score_random_rsv_scale}
\end{figure}

To investigate the relationship between the scale of uniform random vectors sampled from $[-s, s]$ and learning performance when using random vectors instead of reservoirs, we varied the random scale $s$ from 0 to 2 with increments of 0.1. For each scale, we ran the experiment with 100 different random seeds and measured the successful learning probability and average number of steps. Figure \ref{fig:sr_score_random_rsv_scale} shows the results of this experiment. Figure \ref{fig:sr_score_random_rsv_scale}(a) shows that a larger random scale leads to a higher successful learning probability. Figure \ref{fig:sr_score_random_rsv_scale}(b) reveals that in the goal-changing task, the agent fails to learn regardless of the noise scale.
In this study, we used uniformly distributed random numbers within the range of $s=1$ to match the reservoir state, which is restricted by the tanh function.

\section{Varying the reservoir size and connectivity}
\begin{figure}[t]
  \begin{minipage}[t]{0.49\linewidth}
    \centering
    \includegraphics[bb=0 0 600 300, scale=0.3]{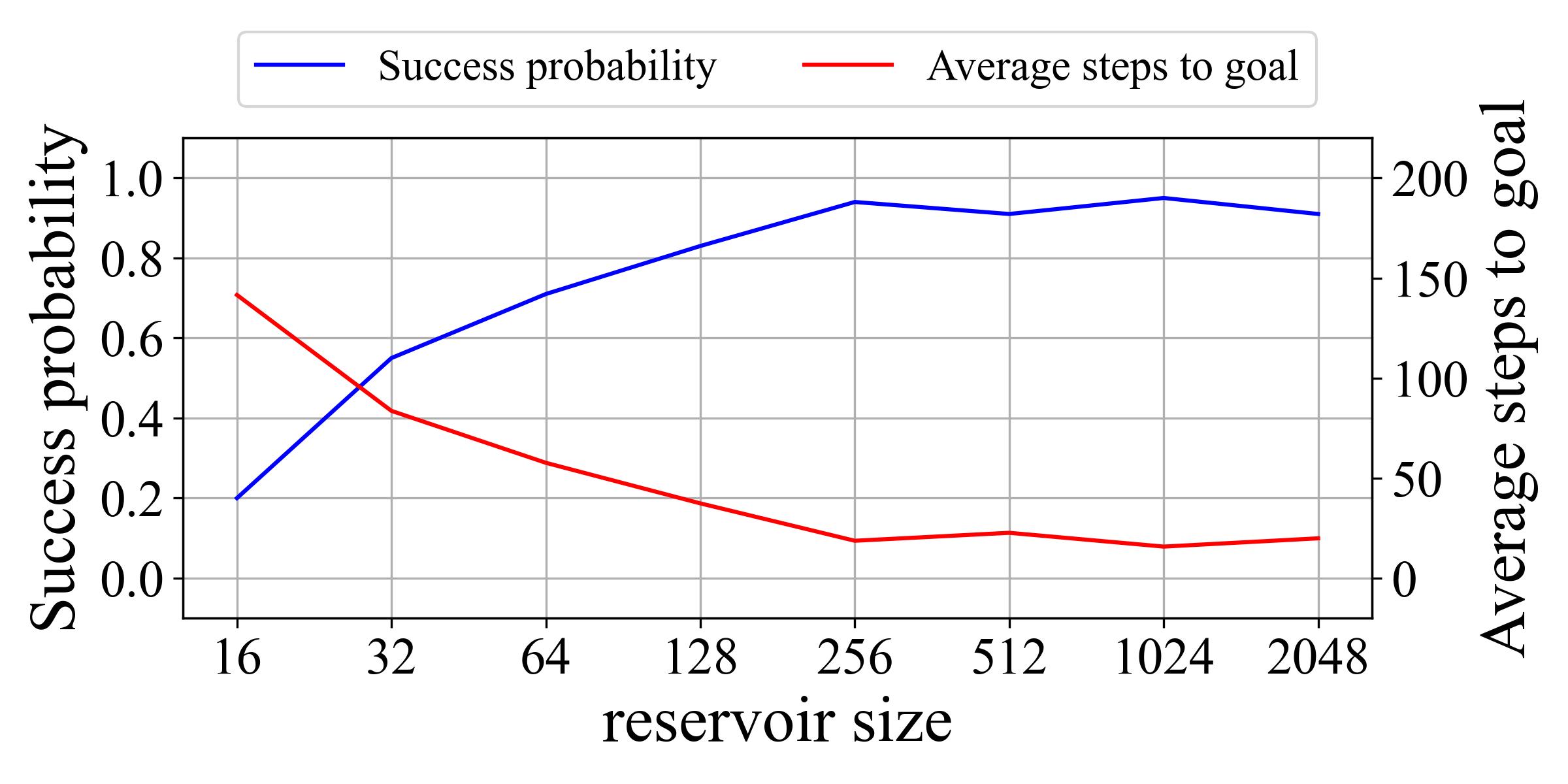}
    \subcaption{Goal task.}
  \end{minipage}
  \begin{minipage}[t]{0.49\linewidth}
    \centering
    \includegraphics[bb=0 0 600 300, scale=0.3]{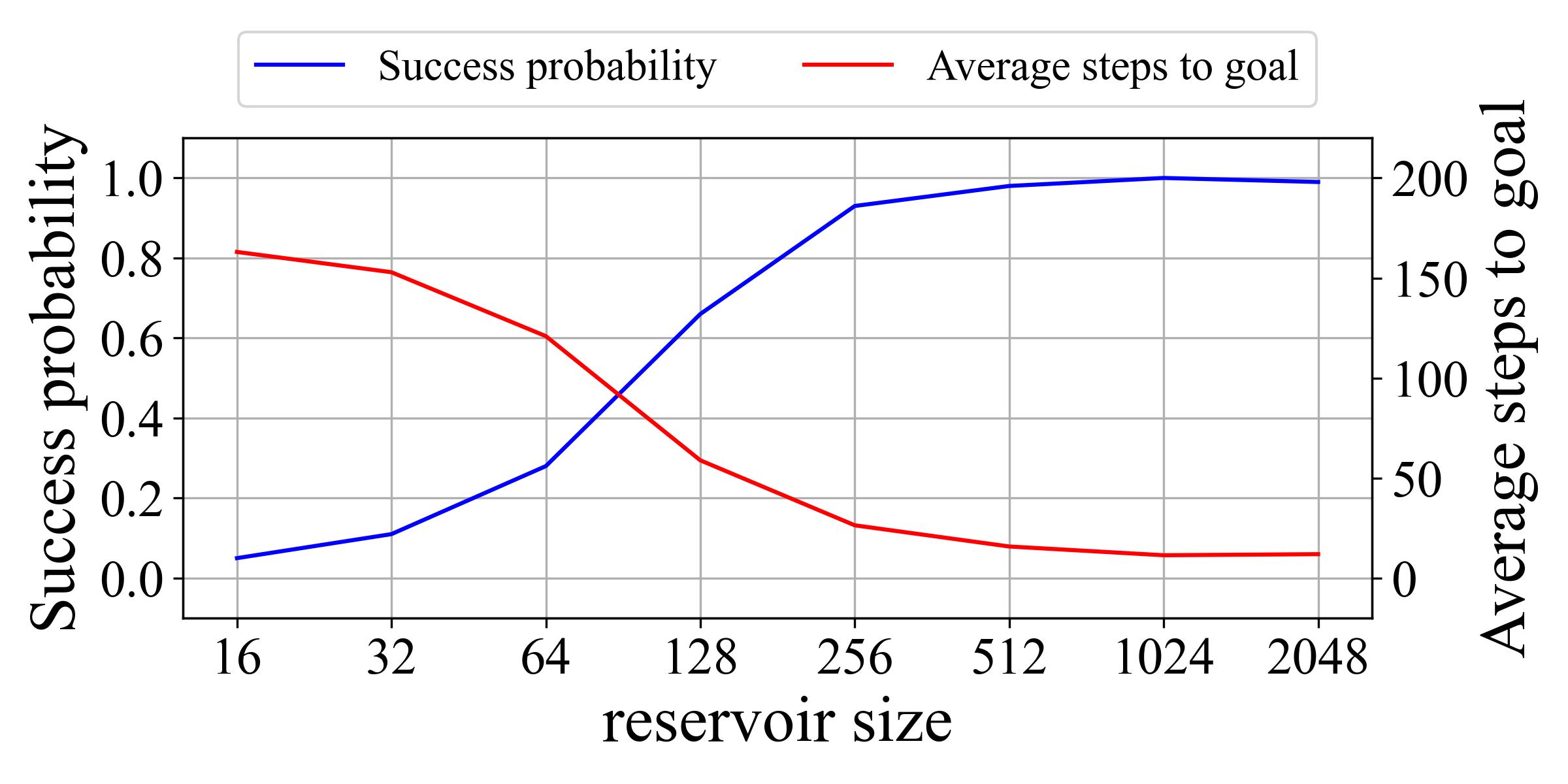}
    \subcaption{Goal change task.}
  \end{minipage}
  \caption{Learning performance of CBRL agents ($g=2.2$) with varying the reservoir size. The definitions of the line colors are the same as in Fig. \ref{fig:sr_score}. (a) shows the learning results for the goal task. (b) shows the learning results of the goal change task. }
\label{fig:sr_rsv_size}
\end{figure}

In addition to the spectral radius $g$, reservoir networks are defined by other crucial parameters. We therefore investigated the dependency of the CBRL agent's learning performance on these parameters, while holding the spectral radius at $g=2.2$. Here, we investigated the successful learning probability and the average number of steps to reach the goal with different reservoir sizes $16 \times 2^n$ where $n$ varies from $0$ to $10$. The experimental results are shown in Figure \ref{fig:sr_rsv_size}. Figure \ref{fig:sr_rsv_size}(a) indicates that a reservoir size of at least 64 is desirable for successful learning. Furthermore, Figure \ref{fig:sr_rsv_size}(b) shows that a reservoir size of at least $256$ is necessary to achieve sufficient performance in the goal change task. It is known that a randomly connected network needs a sufficient number of dimensions to exhibit chaotic behavior. Considering the trade-off between computational cost and performance, we conducted experiments with a reservoir size of $256$ in this study.

\begin{figure}[t]
  \begin{minipage}[t]{0.49\linewidth}
    \centering
    \includegraphics[bb=0 0 600 300, scale=0.3]{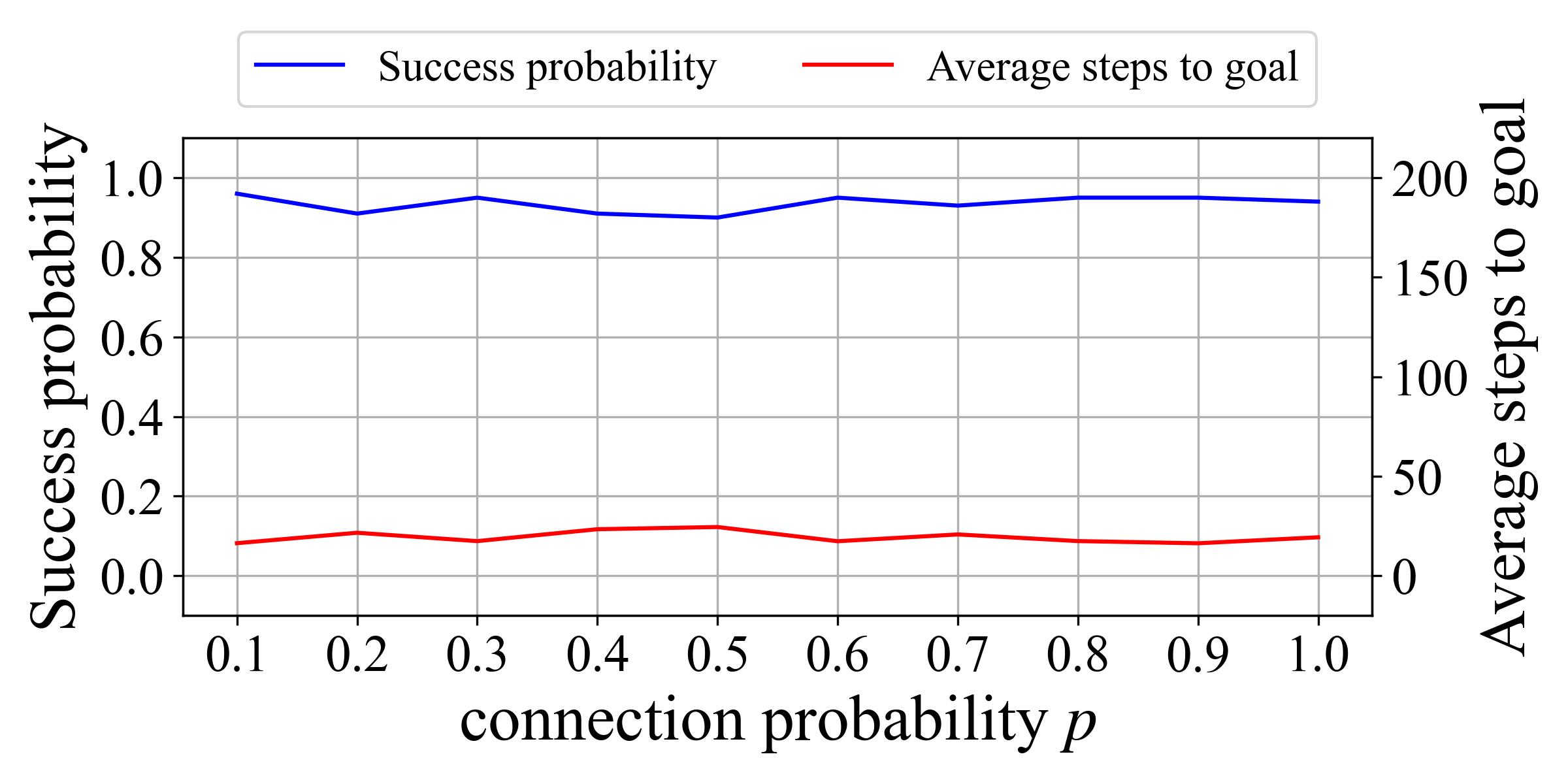}
    \subcaption{Goal task.}
  \end{minipage}
  \begin{minipage}[t]{0.49\linewidth}
    \centering
    \includegraphics[bb=0 0 600 300, scale=0.3]{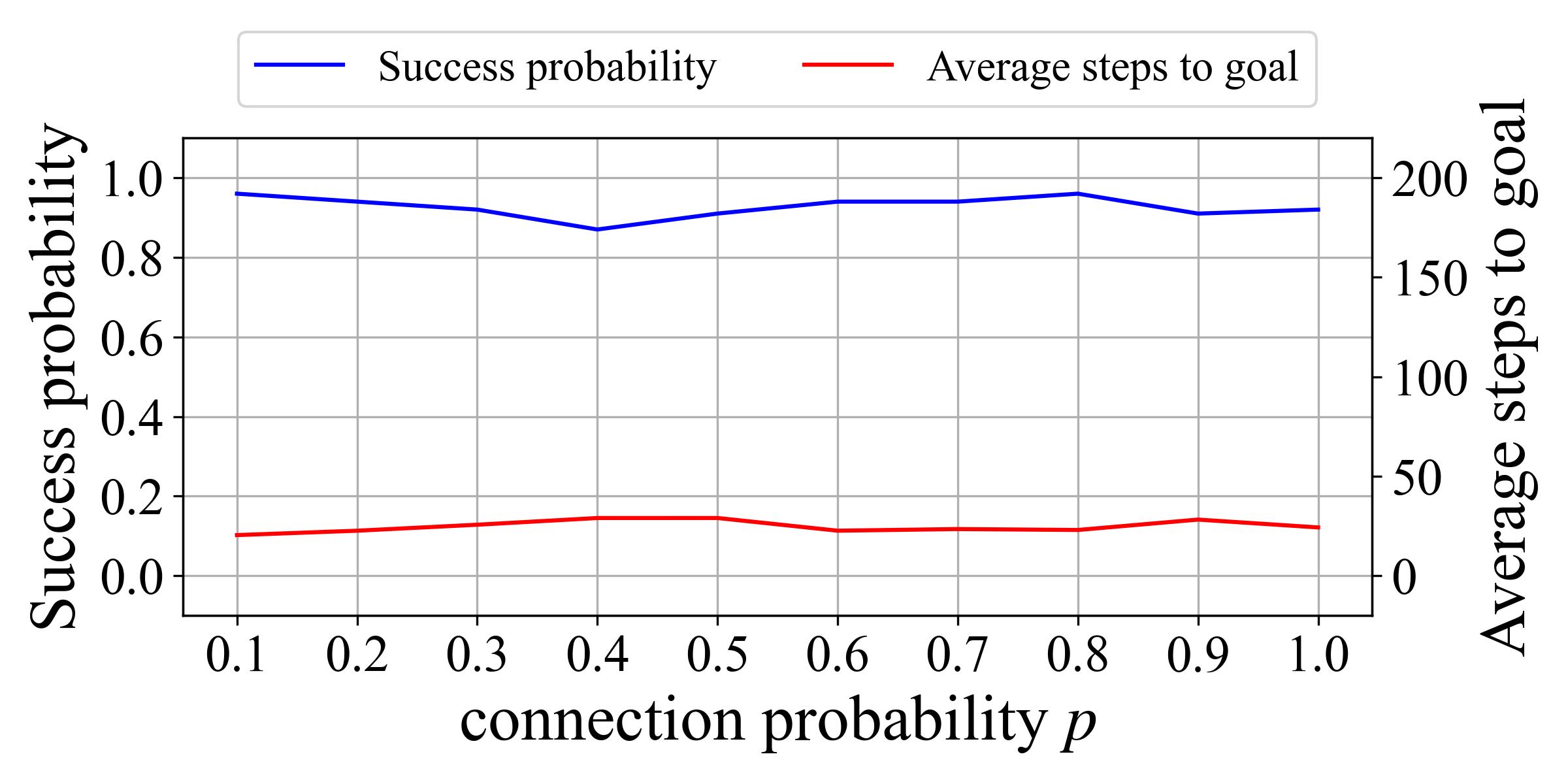}
    \subcaption{Goal change task.}
  \end{minipage}
  \caption{Learning performance of CBRL agents ($g=2.2$) with varying the reservoir connection probability $p$. The definitions of the line colors are the same as in Fig. \ref{fig:sr_score}. (a) shows the learning results for the goal task. (b) shows the learning results of the goal change task. }
\label{fig:sr_rsv_connection}
\end{figure}

Sparse connectivity among reservoir neurons is considered a crucial factor in determining the performance of reservoir computing. Therefore, we investigated how learning performance is affected by varying the connection probability $p$. The results are presented in Figure \ref{fig:sr_rsv_connection}. This figure indicates that under our experimental conditions, the connection probability parameter has no significant impact on performance. This can be considered to be attributed to the fact that, in the simple goal task,  the agent essentially uses the reservoir solely as a source of chaos, rather than for its computational capacity in non-linear time series processing.
\bibliography{main.bib}

\end{document}